\documentclass[10pt,twocolumn,letterpaper]{article}

\usepackage[pagenumbers]{cvpr} 

\usepackage{algorithm}
\usepackage{algorithmic}

\newcommand{\method}{\emph{The Matrix}\xspace}
\newcommand{\controller}{Interactive Module\xspace}
\newcommand{\streamer}{Shift-Window Denoising Process Model\xspace}
\newcommand{\streamerabbr}{Swin-DPM\xspace}
\newcommand{\lcm}{Stream Consistency Model\xspace}
\newcommand{\lcmabbr}{SCM\xspace}
\newcommand{\dataengine}{\emph{GameData} Platform\xspace}
\newcommand{\dataset}{Synthesized Observations of Unreal Rendered Contextual Environments\xspace}
\newcommand{\datasetabbr}{\emph{Source}\xspace}

\newcommand{\secDiT}{\emph{Appendix Section A.1}\xspace}
\newcommand{\secTrain}{\emph{Appendix Section A.2}\xspace}
\newcommand{\secEngine}{\emph{Appendix Section B.1}\xspace}
\newcommand{\secData}{\emph{Appendix Section B.2}\xspace}
\newcommand{\secRobot}{\emph{Appendix Section B.3}\xspace}
\newcommand{\secExampleControl}{\emph{Appendix Section C.1\xspace}}

\newcommand{\supVideo}{\emph{Supplementary Videos}\xspace}
\newcommand{\tocite}[1]{\textcolor{red}{[TO CITE]}}

\newcommand{\tightpara}{\noindent\textbf}

\newcommand{\click}{\textcolor{purple}{\textbf{Click to play with Adobe Acrobat Reader!}}\xspace}

\usepackage[bigfiles]{pdfbase}
\ExplSyntaxOn
\NewDocumentCommand\embedvideos{smm}{
  \group_begin:
  \leavevmode
  \tl_if_exist:cTF{file_\file_mdfive_hash:n{#3}}{
    \tl_set_eq:Nc\video{file_\file_mdfive_hash:n{#3}}
  }{
    \IfFileExists{#3}{}{\GenericError{}{File~`#3'~not~found}{}{}}
    \pbs_pdfobj:nnn{}{fstream}{{}{#3}}
    \pbs_pdfobj:nnn{}{dict}{
      /Type/Filespec/F~(#3)/UF~(#3)
      /EF~<</F~\pbs_pdflastobj:>>
    }
    \tl_set:Nx\video{\pbs_pdflastobj:}
    \tl_gset_eq:cN{file_\file_mdfive_hash:n{#3}}\video
  }
  \pbs_pdfobj:nnn{}{dict}{
    /Type/RichMediaInstance/Subtype/Video
    /Asset~\video
    /Params~<</FlashVars (
      source=#3&
      skin=SkinOverAllNoFullNoCaption.swf&
      skinAutoHide=true&
      skinBackgroundColor=0x5F5F5F&
      skinBackgroundAlpha=0
      autoRewind=true
    )>>
  }
  \pbs_pdfobj:nnn{}{dict}{
    /Type/RichMediaConfiguration/Subtype/Video
    /Instances~[\pbs_pdflastobj:]
  }
  \pbs_pdfobj:nnn{}{dict}{
    /Type/RichMediaContent
    /Assets~<<
      /Names~[(#3)~\video]
    >>
    /Configurations~[\pbs_pdflastobj:]
  }
  \tl_set:Nx\rmcontent{\pbs_pdflastobj:}
  \pbs_pdfobj:nnn{}{dict}{
    /Activation~<<
      /Condition/\IfBooleanTF{#1}{PV}{XA}
      /Presentation~<</Style/Embedded>>
    >>
    /Deactivation~<</Condition/PI>>
  }
  \hbox_set:Nn\l_tmpa_box{#2}
  \tl_set:Nx\l_box_wd_tl{\dim_use:N\box_wd:N\l_tmpa_box}
  \tl_set:Nx\l_box_ht_tl{\dim_use:N\box_ht:N\l_tmpa_box}
  \tl_set:Nx\l_box_dp_tl{\dim_use:N\box_dp:N\l_tmpa_box}
  \pbs_pdfxform:nnnnn{1}{1}{}{}{\l_tmpa_box}
  \pbs_pdfannot:nnnn{\l_box_wd_tl}{\l_box_ht_tl}{\l_box_dp_tl}{
    /Subtype/RichMedia
    /BS~<</W~0/S/S>>
    /Contents~(embedded~video~file:#3)
    /NM~(rma:#3)
    /AP~<</N~\pbs_pdflastxform:>>
    /RichMediaSettings~\pbs_pdflastobj:
    /RichMediaContent~\rmcontent
  }
  \phantom{#2}
  \group_end:
}
\ExplSyntaxOff

\newcommand{\embedvideo}[3]{\embedvideos{\includegraphics[width=#3]{#2}}{#1}}

\definecolor{cvprblue}{rgb}{0.21,0.49,0.74}
\usepackage[pagebackref,breaklinks,colorlinks,allcolors=cvprblue]{hyperref}

\usepackage{epigraph}
\usepackage{multirow}
\usepackage{media9}
\usepackage{subcaption}

\def\paperID{11580} 
\def\confName{CVPR}
\def\confYear{2025}

\title{The Matrix: Infinite-Horizon World Generation with Real-Time Moving Control}

\author{
Ruili Feng$^{1*\ddag}$, Han Zhang$^{1*}$, Zhantao Yang$^{1*}$, Jie Xiao$^{1*}$, Zhilei Shu$^{1*}$,\\
Zhiheng Liu$^1$, Andy Zheng$^{3}$, Yukun Huang$^{2}$, Yu Liu$^{1\dag}$, Hongyang Zhang$^{3,4\ddag}$\\
\\
$^1$Tongyi Lab, $^2$University of Hong Kong, $^3$University of Waterloo, $^4$Vector Insititute\\
\small{*Equal Contribution, $^\dag$Engineer Advisor, $^\ddag$Project Leader}\\
\url{https://thematrix1999.github.io/}\\
}

\begin{document}
\maketitle

\begin{abstract}
We present The Matrix, the first foundational \textbf{realistic world simulator} capable of generating \textbf{infinitely long} 720p \textbf{high-fidelity real-scene} video streams with \textbf{real-time}, responsive control in both first- and third-person perspectives, enabling immersive exploration of richly dynamic environments. Trained on limited supervised data from AAA games like Forza Horizon 5 and Cyberpunk 2077, complemented by large-scale unsupervised footage from real-world settings like Tokyo streets, The Matrix allows users to traverse diverse terrains—deserts, grasslands, water bodies, and urban landscapes—in continuous, uncut hour-long sequences. With speeds of up to 16 FPS, the system supports real-time interactivity and demonstrates \textbf{zero-shot generalization}, translating virtual game environments to real-world contexts where collecting continuous movement data is often infeasible. For example, The Matrix can simulate a BMW X3 driving through an office setting—an environment present in neither gaming data nor real-world sources. This approach showcases the potential of AAA game data to advance robust world models, bridging the gap between simulations and real-world applications in scenarios with limited data. All the codes, data, and model checkpoints in this paper will be open sourced.
\end{abstract}

\epigraph{``This is the world that you know; the world as it was at the end of the 20th century. It exists now only as part of a neural-interactive simulation that we call the Matrix.''}{Morpheus to Neo}

\begin{figure}
    \centering
    \embedvideo{figs/teaser/switch3.mp4}{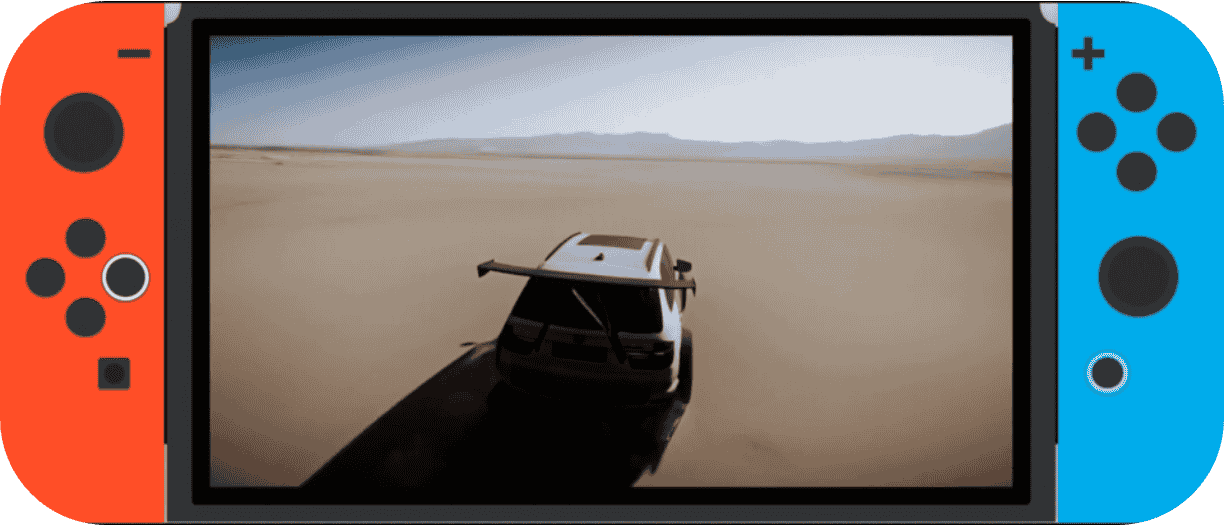}{1\linewidth}\vspace{-2pt}
\embedvideo{figs/teaser/comboVedio3_21M.mp4}{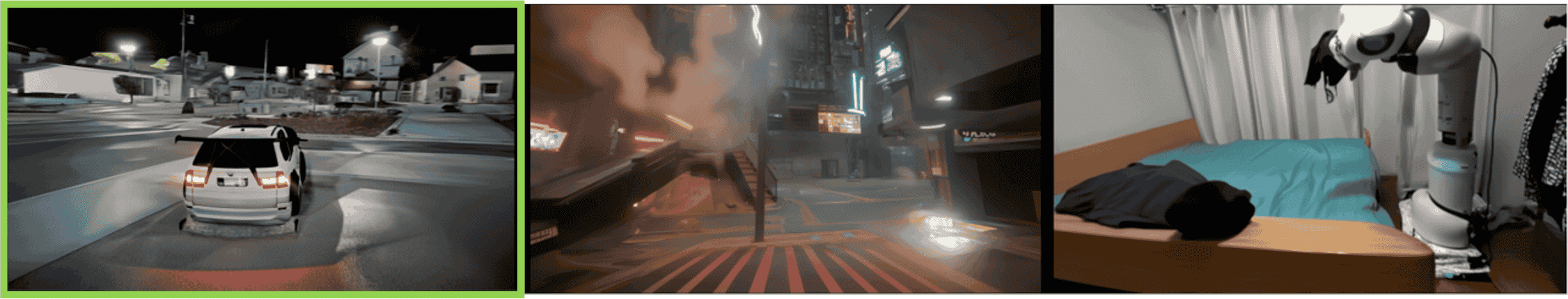}{1\linewidth}
    \captionof{figure}{\method is a foundational \textbf{realistic world simulator} capable of generating \textbf{infinitely long} 720p \textbf{high-fidelity real-scene} video streams with \textbf{real-time}, precise moving control. \click The upper 1-minute demo may need flushing time.}
    \label{fig:enter-label}
\end{figure}

\section{Introduction}
\label{sec:intro}

Neural-interactive simulation, a concept popularized by \emph{The Matrix} (1999), envisions a world fully constructed by AI to replicate 20th-century human society. This paper takes an initial step toward realizing this vision by developing a world model that enables neural networks to `dream' visually authentic environments. The result is an infinite-horizon, high-resolution (720p) simulation that supports real-time (8 - 16 FPS) interactive exploration across diverse landscapes, including deserts, grasslands, water terrains, and urban settings. Responding to real-time control signals, the world model predicts future frames in these environments in a streaming and auto-regressive fashion.


World models offer a promising solution to the overwhelming costs of AAA game development, which can easily run into tens or even hundreds of millions of dollars. Traditional game creation depends on engines such as Unity 3D, Unreal Engine, and Blender, each requiring substantial expertise, intensive asset preparation, and meticulous hyperparameter tuning. Furthermore, games built with these engines are often limited in reusability, as each new title demands a comprehensive redesign. In contrast, data-driven world models tackle these issues by minimizing the need for manual configuration, simplifying development workflows, and boosting scalability across projects.

\begin{table*}[!ht]
    \centering
    \caption{Comparison of recent generative models for game simulation. \method distinguishes itself as a foundation model capable of generating infinitely long videos with AAA game quality, high resolution, frame-level real-time control, and robust domain generalization. Here, * indicates concurrent work with \method, and supervised/unsupervised refers to the video data with/without true control signal.}
    \resizebox{\linewidth}{!}{
    \begin{tabular}{l||c|c|c|c|c|c|c} 
        \hline
        \textbf{Feature} & \textbf{Genie} & \textbf{DIAMOND} & \textbf{MarioVGG*} & \textbf{GameNGen*} & \textbf{Oasis*} & \textbf{GameGen-X*} & \textbf{The Matrix} \\ \hline\hline
        Video Length & 2s & \cellcolor{gray!20} Infinite & 6 Frames & \cellcolor{gray!20} Infinite & \cellcolor{gray!20} Infinite & 4s–16s & \cellcolor{gray!20} Infinite \\ \hline
        \multirow{2}{*}{Training Corpus} & 2D Games & Atari & \multirow{2}{*}{Mario} & \multirow{2}{*}{DOOM} & \multirow{2}{*}{Minecraft} & \multirow{2}{*}{AAA Games} & \cellcolor{gray!20} AAA Games (supervised, small) \\
        & (unsupervised) & CS:GO & & & & & \cellcolor{gray!20} Internet Videos (unsupervised, large)\\ \hline
        Resolution & 360p & $280\times 150$ & $64\times48$ & 240p & \cellcolor{gray!20} 720p & \cellcolor{gray!20} 720p & \cellcolor{gray!20} 720p \\ \hline
        Control & \cellcolor{gray!20} Frame-Level  & \cellcolor{gray!20} Frame-Level  & Video-Level  & \cellcolor{gray!20} Frame-Level  & \cellcolor{gray!20} Frame-Level  & Video-Level  & \cellcolor{gray!20} Frame-Level  \\ \hline
        Real-Time & No & \cellcolor{gray!20} Yes & No & \cellcolor{gray!20} Yes & \cellcolor{gray!20} Yes & No & \cellcolor{gray!20} Yes \\ \hline
        Control Generalization & \cellcolor{gray!20} Yes & No & No & No & No & No & \cellcolor{gray!20} Yes \\ \hline
    \end{tabular}
    }
    \label{tab:comparison with related work}
\end{table*}

Despite extensive research in world models~\cite{schmidhuber2015learning}, key challenges remain. First, prior studies have predominantly focused on non-AAA video games, such as Atari~\cite{schrittwieser2020mastering,hafner2020mastering,alonso2024diffusion}, Mario~\cite{Virtuals2024mariovgg}, Minecraft~\cite{oasis2024,hafner2023mastering}, Counter-Strike: Global Offensive (CS:GO)~\cite{alonso2024diffusion}, and DOOM~\cite{valevski2024diffusion}, which fall short in replicating real-world fidelity. Second, current video generation techniques, like Sora~\cite{openai2024sora}, are constrained to short sequences of about 1 minute, forcing existing world models to assemble independently generated clips with noticeable transitions. Finally, achieving real-time generation remains a major hurdle. For example, state-of-the-art 2D platformer game generator Genie~\cite{bruce2024genie} runs as slow as 1 FPS. This paper addresses these limitations by introducing the first scalable, high-fidelity (1280$\times$720 pixels) world model in real time that enhances simulation realism and bridges the gap between virtual environments and reality. Notably, our world model is the first with strong domain generalization and real-time control. For example, our foundation model allows us to control BMW X3 driving through an indoor setting or in the sea—an environment
present in neither gaming data nor real-world sources.

\subsection{Our Contributions}

Our contributions are as follows:

\begin{itemize}
    \item
    We introduce \emph{The Matrix}, the first foundational simulator for realistic worlds, capable of generating infinitely long, high-fidelity 720p real-scene video streams with real-time, interactive controls and strong domain generalization. The model is light and consists of 2.7B parameters.
    \item
    At the core of \emph{The Matrix} is a novel diffusion technique, the Shift-Window Denoising Process Model (Swin-DPM), enabling pre-trained DiT models~\cite{peebles2023scalable} to extrapolate seamlessly for smooth, continuous, and infinitely extendable video creation. This technique holds potential for broader applications in long-form video generation.
    \item
    Additionally, we introduce \emph{GameData}, a platform that autonomously captures paired in-game states—extracted from CPU memory—alongside corresponding video frames, significantly reducing labeling costs and complexity. This platform produces \datasetabbr, a new training dataset for world models with action-frame paired data.
\end{itemize}

\subsection{Technical Advantages of \textbf{\emph{The Matrix}}}
Tab. \ref{tab:comparison with related work} highlights a comparison between \method and other game generation models across six key features. Our work advances the state-of-the-art of world models in the following aspects:

\begin{itemize}
    \item \textbf{Infinite Video Generation:} \method generates consistent, infinitely long video sequences using a streaming, auto-regressive approach.
    \item \textbf{High-Quality Rendering:} \method delivers AAA-level, realistic rendering at a resolution of $1280\times 720$.
    \item \textbf{Real-Time, Frame-Level Control:} \method operates with speeds of 8 - 16 FPS, providing real-time, frame-level control for interactive applications.
    \item \textbf{Domain Generalization:} Trained with small amounts of supervised AAA game data and large amounts of unsupervised internet videos, \method achieves strong domain generalization to real-world settings.
\end{itemize}

\section{Related Work}
\label{sec:related work}

\tightpara{World Model for Agent Learning.} Developing world models for training agents has been a long-standing research focus, aimed at enhancing policy learning within simulated environments rather than solely achieving high-fidelity reconstructions of observations. This research involves two primary stages: 1) modeling the training environment by reconstructing observations, rewards, and continuation signals, often through a recurrent state-space model; and 2) utilizing this model to predict future states, enabling reinforcement learning to optimize robust policy functions. Studies indicate that this method provides sample efficiency gain of over 1000\% compared to directly learning policies from real environments, shows resilience across diverse domains, and can outperform fine-tuned expert agents on a range of benchmarks and data budgets~\cite{hafner2023mastering}. Key contributions in this area include Recurrent World Models~\cite{ha2018recurrent}, Dreamer (v1~\cite{hafner2019dream}, v2~\cite{hafner2020mastering}, and v3~\cite{hafner2023mastering}), TD-MPC (v1~\cite{hansen2022temporal} and v2~\cite{hansen2023td}), DayDreamer~\cite{wu2023daydreamer}, SafeDreamer~\cite{huang2023safe}, and MuDreamer~\cite{burchi2024mudreamer}. Notably, MuZero~\cite{schrittwieser2020mastering} runs the self-play of Monte Carlo tree search to build world models for Atari, Go, chess and shogi, without external data.

\medskip
\tightpara{World Simulation.} Distinct from world models designed for agent learning, another research direction emphasizes world simulation, focusing on human interaction with neural networks through high-quality rendering, robust control, and strong domain generalization to real-world scenarios. This research explores two types of control: video-level and frame-level. In video-level control, a control signal is given at the start, and the model generates a responsive video sequence; notable examples include UniSim~\cite{yang2023learning}, Pandora~\cite{xiang2024pandora}, GameGen-X~\cite{che2024gamegen}, MicroVGG~\cite{Virtuals2024mariovgg}, and GAIA-1~\cite{hu2023gaia}. To approximate continuous control, this approach often stitches together independently generated clips, which may result in visible transitions. In contrast, frame-level control provides fine-grained adjustments every few frames, enabling more precise, responsive interactions similar to gameplay, as seen in examples like Genie~\cite{bruce2024genie}, DIAMOND~\cite{alonso2024diffusion}, GameNGen~\cite{valevski2024diffusion}, and Oasis~\cite{oasis2024}. Prior work in world simulation has typically focused on one of three aspects—video length, high resolution, or domain generalization—without addressing all three simultaneously. Table \ref{tab:comparison with related work} presents a comparison between \method and prior works. \method uniquely stands out as a foundation model capable of generating infinitely long, AAA-quality videos with high resolution, frame-level real-time control, and strong generalization to real-world contexts.
\section{Methods}
\begin{figure}
    \centering
    \includegraphics[width=1\linewidth]{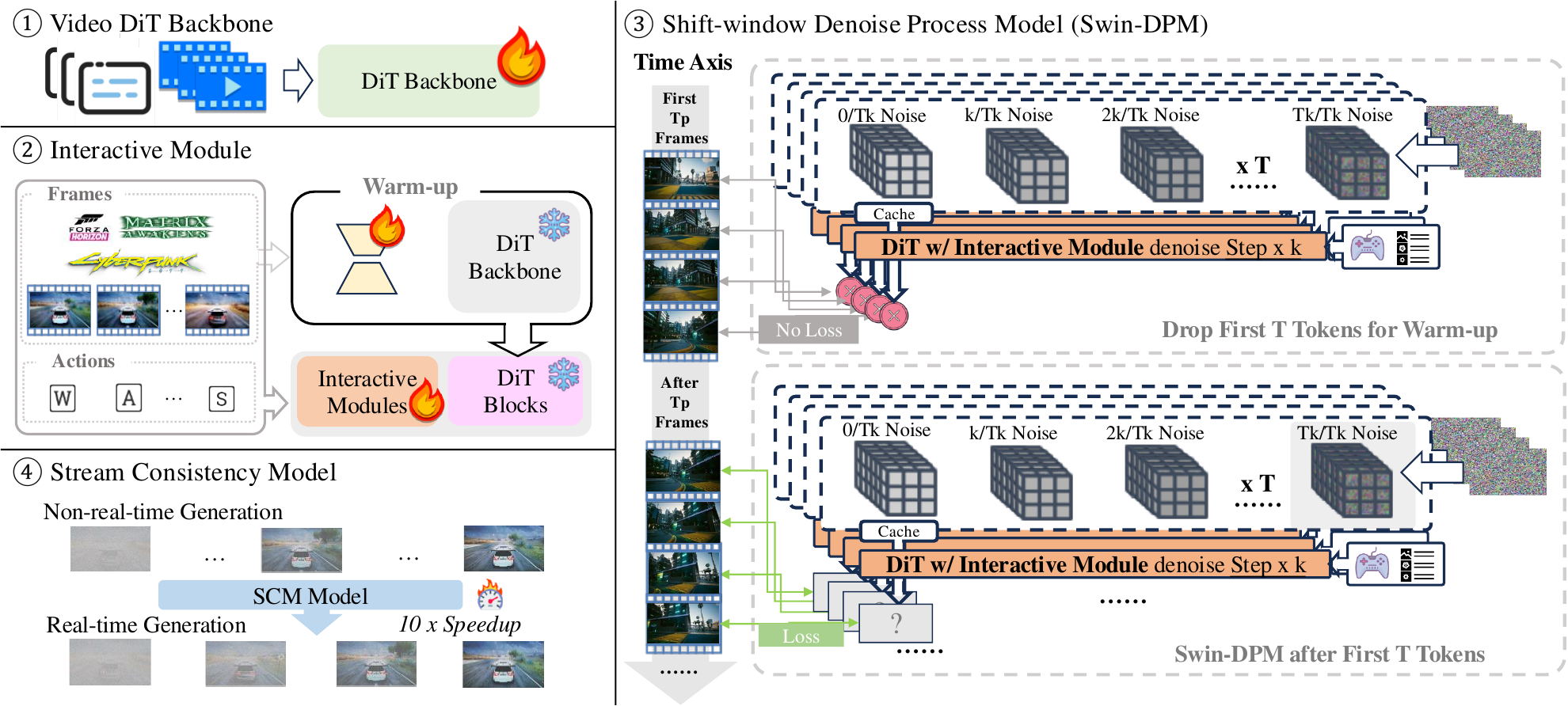}
    \caption{The training process of \method begins with a pretrained video DiT backbone. First, the \controller is warmed up using \dataset data with unsupervised LoRA to make subsequent training focus on movement, not visuals. Then, we train the \controller for precise frame-level control. \streamerabbr enables infinite-length generation, and \lcm is introduced to accelerate sampling to real-time speeds.}
    \label{fig:train}
\end{figure}
Achieving granular control is notoriously challenging, as labeling actions at the frame level is typically cost-prohibitive. To address this, we develop the \emph{GameData} platform, which autonomously captures paired data of in-game states (extracted directly from CPU memory) alongside corresponding video frames, significantly reducing labeling costs and complexity. Additionally, \method incorporates an advanced \controller that learns and generalizes game movement interactions from a limited amount of labeled data combined with extensive unlabeled data from both games and real-world environments. This enables \method to deliver exceptional accuracy across diverse scenarios, while maintaining robust performance in the gaming domain.

Generating high-quality, real-time, and generalizable video simulations for infinite sequences presents additional technical challenges, often forcing previous simulators to compromise on one or more essential aspects. \method overcomes these limitations by adapting the world model from a pre-trained video Diffusion Transformer (DiT) model~\cite{peebles2023scalable}, leveraging its extensive pre-existing knowledge and generation quality. To enable infinite-length generation, \method introduces a novel diffusion approach, the \streamer (\streamerabbr), which allows the DiT model to extrapolate for smooth, continuous, and indefinitely long video creation. Finally, to achieve real-time efficiency, we fine-tune a \lcm (\lcmabbr), accelerating inference to real-time.

\medskip
\tightpara{Video DiT Backbone.} 
As a preliminary, we introduce the video DiT backbone, adapted from the publicly available DiT models~\cite{opensora}. It employs a 3D Variational Auto-Encoder (VAE) to encode $T\times p$ video frames into $T$ video tokens. The backbone consists of 32 attention blocks, followed by a linear output head with LayerNorm~\cite{ba2016layer}. Each attention block includes a self-attention layer operating on network features, a cross-attention layer linking conditions with self-attention outputs, and an FFN layer composed of two linear layers with a GELU activation~\cite{hendrycks2016bridging} in between. See \secDiT for further details.

\subsection{Model Components}
\method comprises three main components: a) an \textbf{\controller} that interprets user intentions (e.g., keyboard inputs) and integrates them into video token generation; b) a \textbf{\streamer} (\streamerabbr) that enables infinite-length video generation; and c) a \textbf{\lcm} (\lcmabbr) that accelerates sampling to achieve real-time performance. As shown in \cref{fig:train}, the model is fine-tuned from a pre-trained video DiT model through a three-stage process: first, we fix the DiT model parameters and train the \controller; next, we train the \controller and the DiT together following the \streamerabbr; finally, we optimize an \lcmabbr to accelerate inference to real-time speeds. The first two stages leverage both labeled gaming and unlabeled internet video data to enhance generalization, while the final \lcmabbr training focuses on labeled gaming data to reduce optimization complexity.

\begin{figure}[t]
     \centering
     \begin{subfigure}[t]{0.45\textwidth}
       \includegraphics[width=\textwidth]{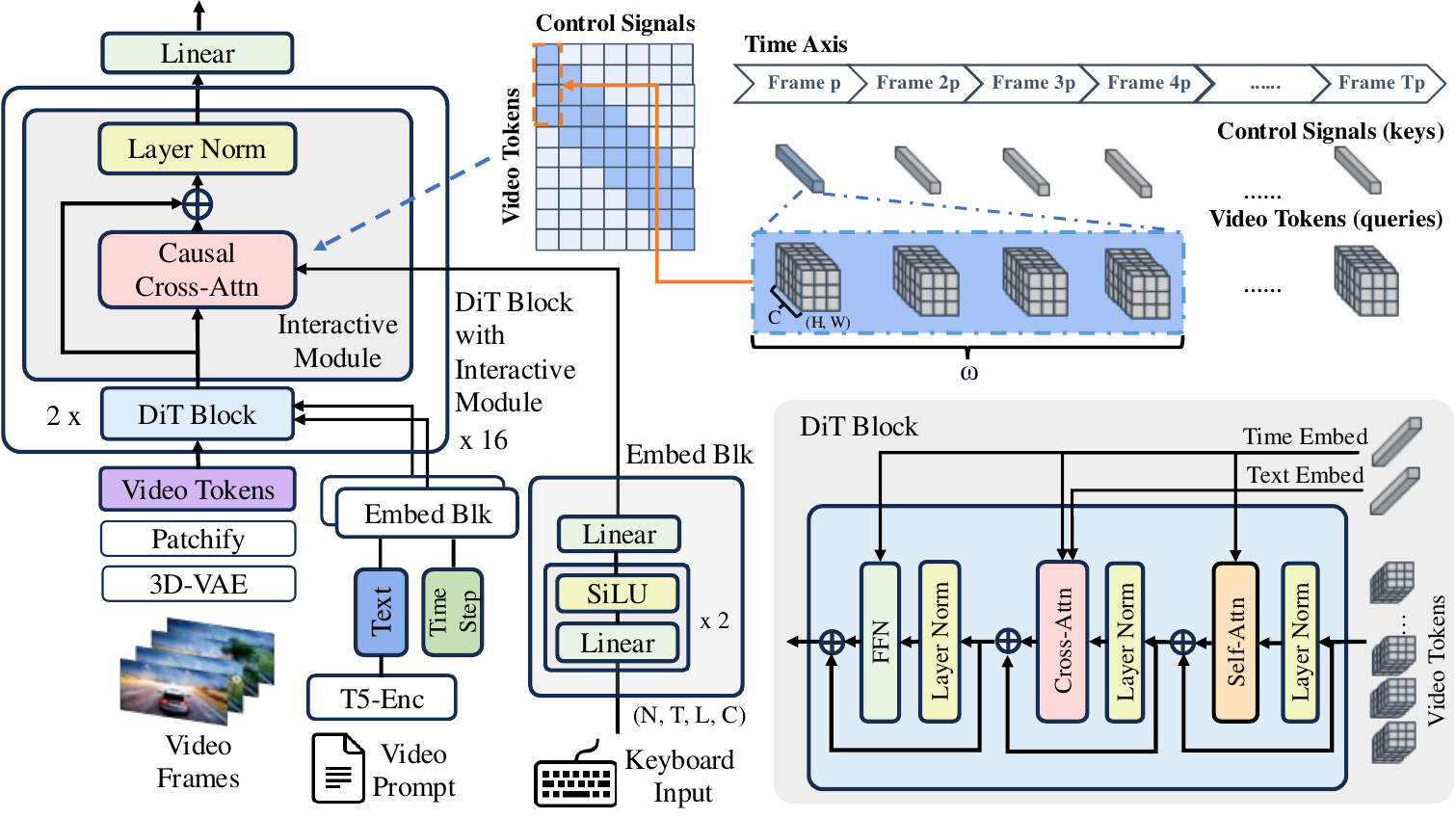}
       \caption{\textbf{The \controller:} After every two DiT blocks, the module merges the keyboard inputs into the video token feature through a \textbf{Causal Cross-Attention Layer}, where each keyboard input is limited to influence only the current and subsequent $\omega$ tokens. Here, every $p$ frames are condensed into a single token.}
       \label{fig:controller}
     \end{subfigure}
     \hfill
     \begin{subfigure}[t]{0.45\textwidth}
       \includegraphics[width=\textwidth]{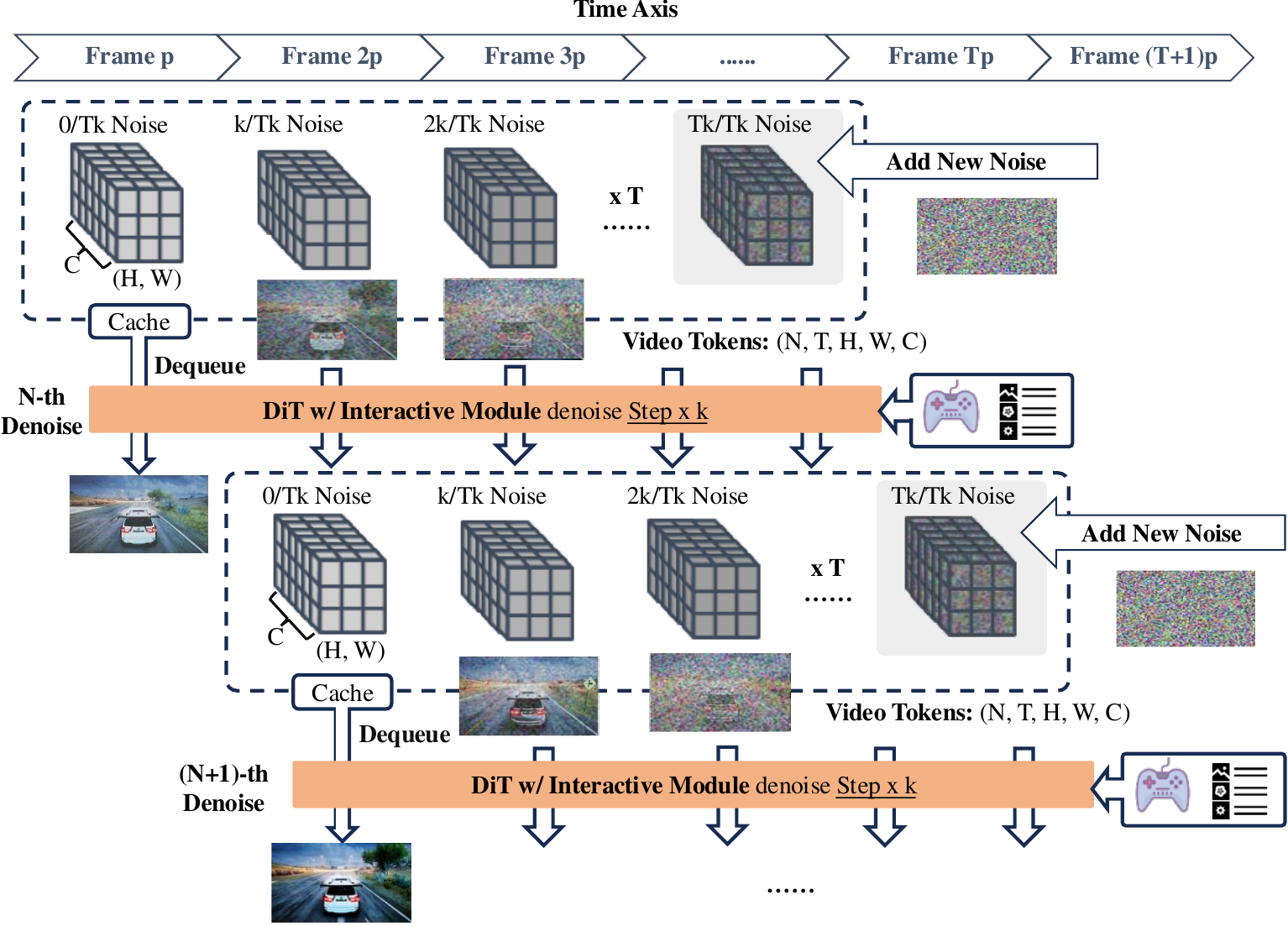}
       \caption{\textbf{\streamer:} The \streamerabbr transforms the traditional diffusion process into a streaming one, where $T$ video tokens with different noise levels are denoised simultaneously. After each token is fully denoised and dequeued for decoding, a new token of pure noise is added to the queue. The dequeued token is then copied to the cache, allowing it to continue participating in attention computations until the next token is dequeued.}
       \label{fig:swin-dpm}
     \end{subfigure}
     \caption{Main components of \method.}
     \label{fig:mainfig}
   \end{figure}

\medskip
\tightpara{\controller.} 
The \controller consists of an Embedding block (see \cref{fig:controller}) and a cross-attention layer. Its primary function is to translate keyboard inputs into natural language that guides video generation. For example, pressing `W' is interpreted as “The car is driving forward” in the \textit{Forza Horizon 5} scenario, or as ``The man is moving forward and looking up'' when combined with an upward mouse movement in \textit{Cyberpunk 2077}. For unlabeled real or game data, we apply a default description: ``The camera is moving in an unknown way.'' To enhance robustness, we randomly replace labeled keyboard inputs with this default sentence during training with probability \( q=0.1 \). 

To prepare for training, we first warmup the base DiT model for a few epochs using collected game and real-world data, fine-tuning a LoRA weight~\cite{hu2021lora}. This process ensures that the \controller focuses on learning interactions and movement patterns rather than simply fitting the video.

Once translated, these natural language descriptions are processed by a T5 encoder~\cite{raffel2020exploring} and transformed into a vector embedding through two linear layers and a SiLU layer~\cite{elfwing2018sigmoid} between them. This vector embedding is then concatenated with its corresponding video token and the next \( \omega \) video tokens, where \( \omega \) is a pre-defined causal relation range, typically set to \( \omega=4 \), as is shown in \cref{fig:controller}.

We perform this cross-attention operation each time the DiT model completes an odd-numbered self-attention step, enabling effective information exchange across frames and achieving precise, frame-level control for video generation.

\medskip
\tightpara{\streamer.}
Typical DiT models are limited to generating only a few seconds of video, even when substantial spatial and temporal compression is applied via VAEs. This limitation is largely due to the high computational cost and memory demands of attention mechanisms over extended time durations. To address this, it becomes crucial to assume that temporal dependencies are confined within a limited time window, beyond which attention computations are unnecessary. Building on this idea, we propose the \streamer (\streamerabbr), which leverages a sliding temporal window to manage dependencies effectively and enables the generation of long or even infinite videos by producing tokens with a stride of $s=1$. As is shown in \cref{fig:swin-dpm}, within each window, a queue of video tokens undergoes denoising at various noise levels. After $k$ denoising steps (where $k\times T$ is the number of diffusion solver steps), the leftmost, lowest-noisy token is dequeued into a cache. To maintain the queue length, a new token with Gaussian noise will be then added to the rightmost position. Each cached token is re-appended to the window’s token queue at noise level $0$ until the next token is cached, allowing it to continue participating in denoising and ensuring continuity between different windows. The network of \streamerabbr is fine-tuned from a pre-trained DiT model. During training, we sample $2w$ video tokens, where $w$ is the window size. We usually set $w=T$. The first $w$ tokens are used solely for warming up \streamerabbr and do not participate in backpropagation; loss is computed only on the last $w$ tokens. At inference time, we follow the same setup: the first $w$ tokens are for warmup and are discarded, with the generated video starting from the $(w+1)$-th token.
\begin{table*}[!ht]
    \centering
    \caption{Ablation study on the components of \method. Note that there is a trade-off between inference speed, control precision, and rendering quality. \textbf{Move-LPIPS} and \textbf{Move-PSNR} are computed between the generated videos and test videos with ground truth movements.}
    \resizebox{\linewidth}{!}{
    \begin{tabular}{l||c|c|c|c|c|c|c|c} 
        \hline
        \textbf{Component} & \textbf{Scene}&  \textbf{\#Params} &\textbf{Inference Speed}& \textbf{FVD$\downarrow$} & \textbf{FID$\downarrow$} & \textbf{CLIP$\uparrow$} & \textbf{Move-LPIPS$\downarrow$} & \textbf{Move-PSNR$\uparrow$}  \\ 
        \hline\hline
        DiT Backbone & - & 2.3B & 48 frames / 34 Seconds & 1016.30 & 318.10 & 0.30 & - & - \\
        \hline\hline
        \multirow{3}{*}{+ Warmup} & \emph{Cyberpunk 2077} & 2.3B & 64 frames / 34 Seconds & 1429.45 & 183.24  & 0.28 & 0.125 & 27.80 \\ 
         & \emph{DROID} & 2.3B & 48 frames / 34 Seconds & 1133.16 & 224.98 & 0.29 & 0.191 & 27.72 \\ 
        
         & \emph{Forza Horizon 5} & 2.3B & 48 frames / 34 Seconds & 1891.67 & 141.11 & 0.31& 0.128& 26.89 \\
        \hline\hline
        \multirow{3}{*}{+ \controller}
         & \emph{Cyberpunk 2077} & 2.7B & 48 frames / 55 Seconds & 1112.49 & 173.31 & 0.28 & 0.129& 28.24 \\
         & \emph{DROID} & 2.7B & 48 frames / 55 Seconds & 1200.82& 237.66 & 0.30& 0.180 & 27.90 
         \\
         & \emph{Forza Horizon 5} & 2.7B & 48 frames / 55 Seconds & 1211.30 & 119.20 & 0.27& 0.125& 28.98 \\
        \hline\hline
        + \streamerabbr & \emph{Forza Horizon 5} & 2.7B & 0.8 FPS & 1651.50 & 163.27 & 0.24 & 0.113& 29.90 \\
        \hline
        + \lcmabbr & \emph{Forza Horizon 5} & 2.7B & 8 - 16 FPS & 1936.79 & 153.80 & 0.23& 0.109& 29.73 \\
        \hline
    \end{tabular}
    }
    \label{tab:metrics_main}
\end{table*}

\medskip
\tightpara{\lcm.} After extending the DiT model to \streamerabbr, we further address the need for achieving real-time rendering of the simulated world. A promising approach is to combine \streamerabbr with Consistency Models~\cite{song2023consistency,songimproved}, a leading method for accelerating diffusion. We use the \lcm (\lcmabbr)~\cite{kodaira2023streamdiffusion}, which distills the original diffusion process and its class-free guidance into a four-step consistency model while incorporating the denoising window design from \streamerabbr. The training procedure is illustrated in \cref{fig:train}. This integration results in a 10 - 20$\times$ acceleration in inference speed, reaching a rendering rate of 8 - 16 FPS.

\subsection{Construction of the \textbf{\emph{Source}} Dataset}
To train \method model, we construct the \dataset (\datasetabbr) dataset, which consists of two components: synthetic game data from Unreal Engine and real-world, unlabeled footage. The synthetic game data, collected using the \dataengine, serves as supervised training data for precise motion control, while the real-world footage improves the model’s visual quality and generalization to real-world scenarios.

After collection, the data is segmented into 6-second clips of continuous scenes and captioned using GPT-4o~\cite{hurst2024gpt}, resulting in a dataset of 750k labeled samples and 1.2 million unlabeled samples, all with 60 FPS. The labeled game data is further refined to ensure a balanced distribution of all possible game states. For more details on the dataset, see \secData.

\medskip
\begin{figure}
    \centering
    \includegraphics[width=0.7\linewidth]{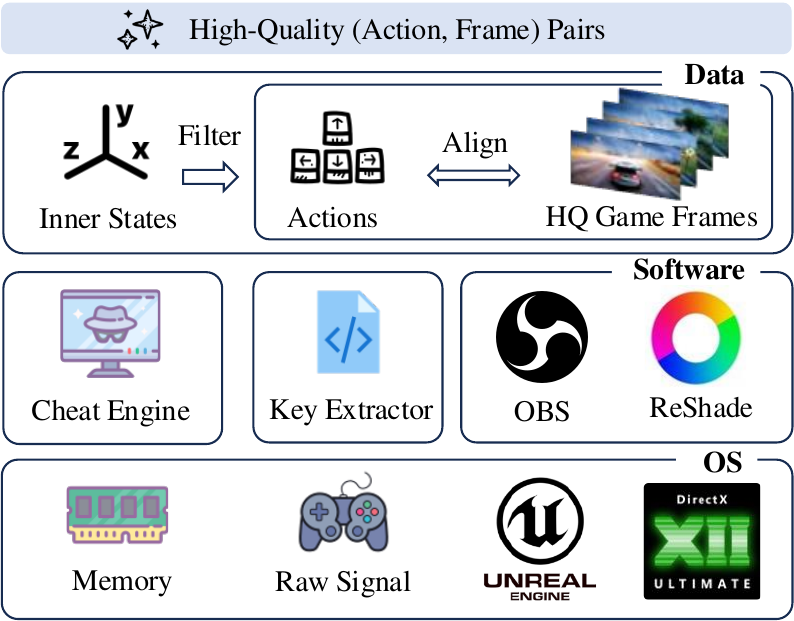}
    \caption{\textbf{The \dataengine} that creates the \datasetabbr dataset. It uses \emph{CheatEngine} to capture in-game status from CPU memory and filter out unsatisfactory frames, such as those with stuck characters or irregular movements. \emph{Reshade} removes game UIs and HUDs to ensure a more consistent data distribution. A Key Extractor then captures keyboard inputs and aligns them with frames recorded by \emph{OBS}.}
    \label{fig:dataengine}
\end{figure}
 \tightpara{The \dataengine.} As shown in \cref{fig:dataengine}, the \dataengine is built on open-source tools: \textit{Cheat Engine} software~\cite{cheatengine}, the \textit{Reshade} plugin~\cite{reshade} for DirectX, and \textit{OBS Recording} software~\cite{OBS}. \textit{Cheat Engine} is used to capture in-game world status data, such as character $(x, y, z)$ positions and camera movements. This status data is aligned with recorded video frames to create per-frame action-video pairs and is also used to check if the character or camera is stuck and requires a reboot. We employ the \textit{Reshade} plugin to remove all game UIs and HUDs and to standardize shading styles, providing a more consistent, low-complexity data source. Data for \textit{Forza Horizon 5} is collected using autonomous scripts with random walking algorithms, while \textit{Cyberpunk 2077} data is gathered manually with human operators running the \dataengine. See \secEngine for more details on the \dataengine.

\section{Experiments}

\begin{figure*}
    \centering
    \begin{tabular}{c}
    \includegraphics[width=0.2\linewidth]{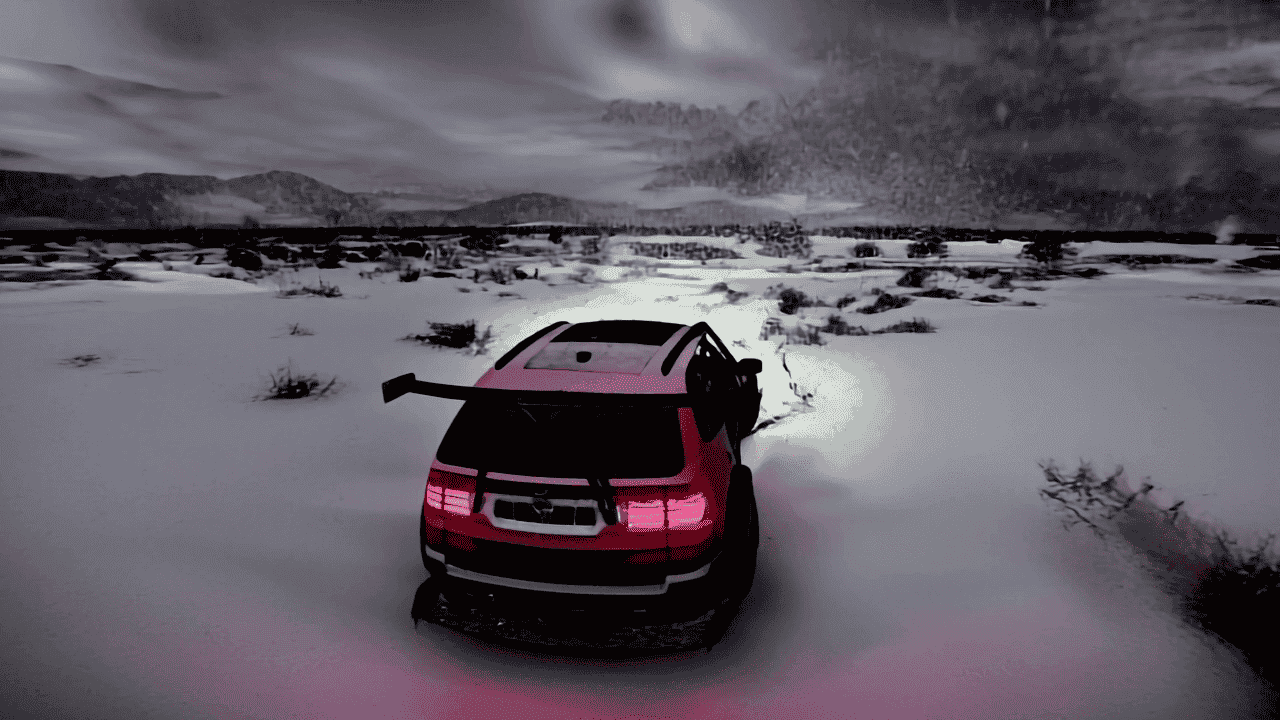}
     \includegraphics[width=0.8\linewidth]{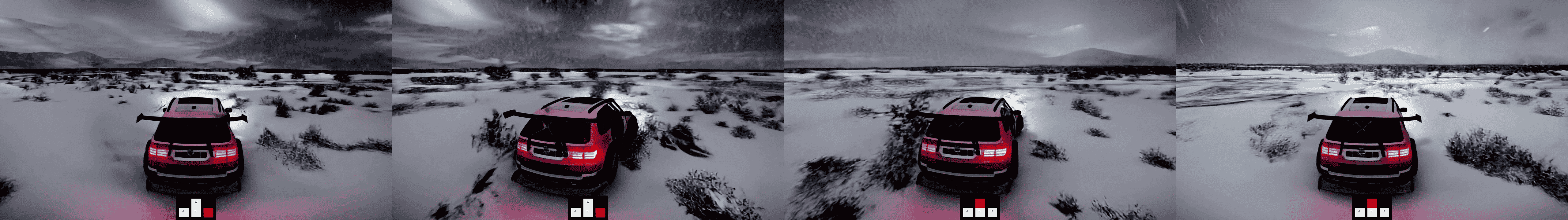}\vspace{-3.7pt}
     \\
     \includegraphics[width=0.2\linewidth]{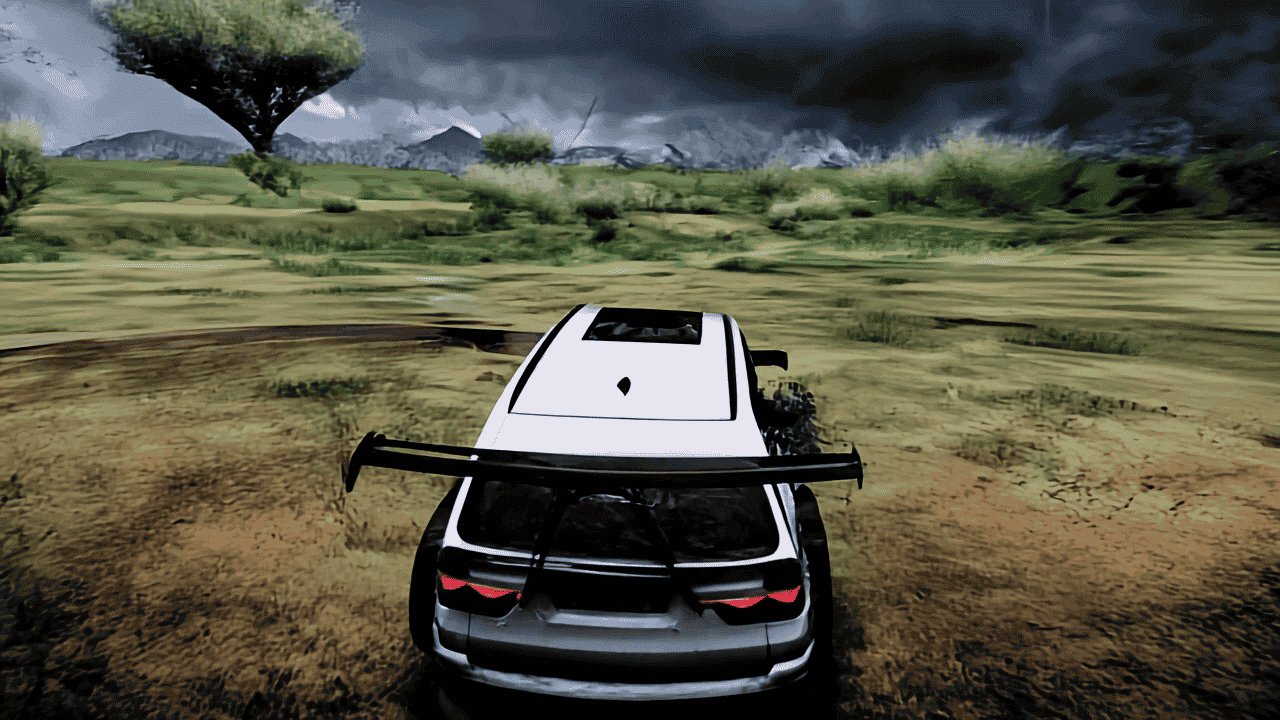}
     \includegraphics[width=0.8\linewidth]{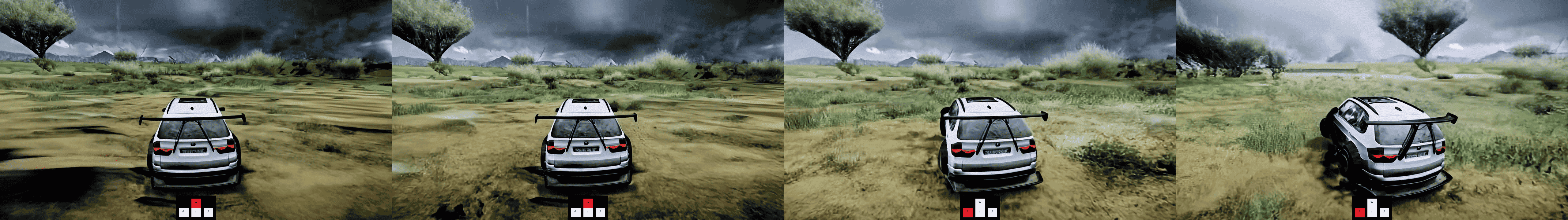}\vspace{-3.7pt}\\
     \includegraphics[width=0.2\linewidth]{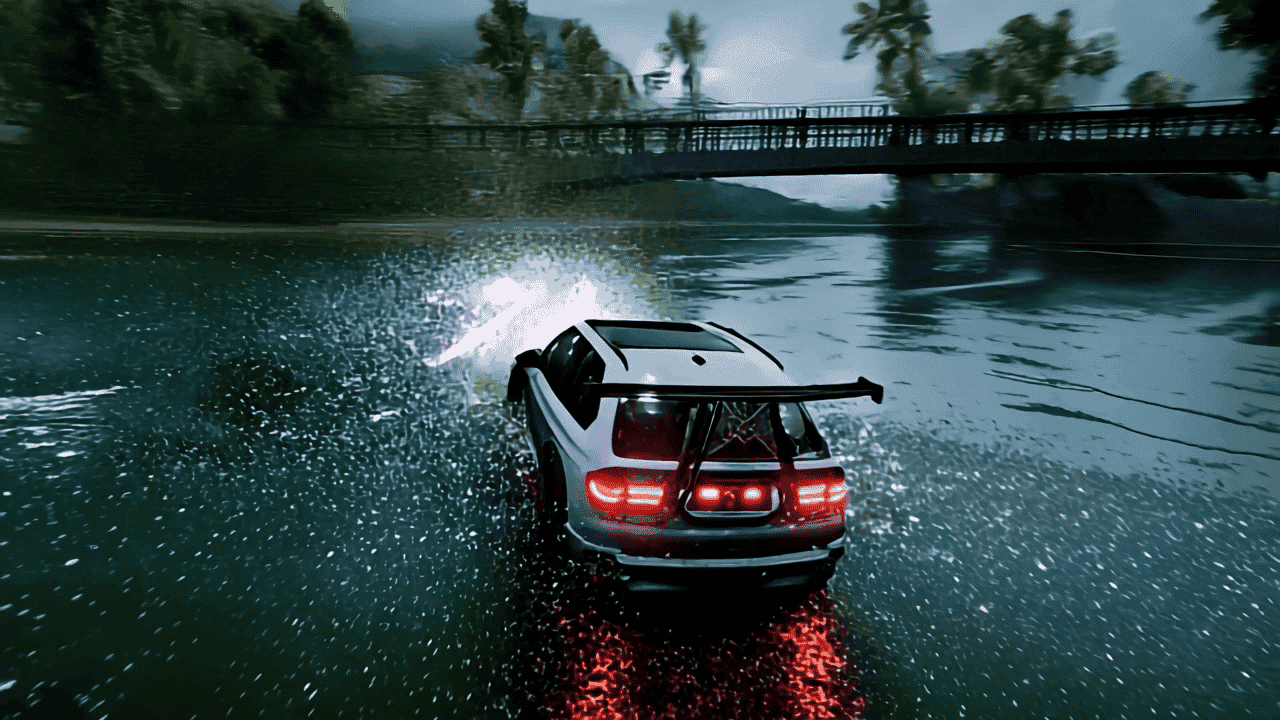}
     \includegraphics[width=0.8\linewidth]{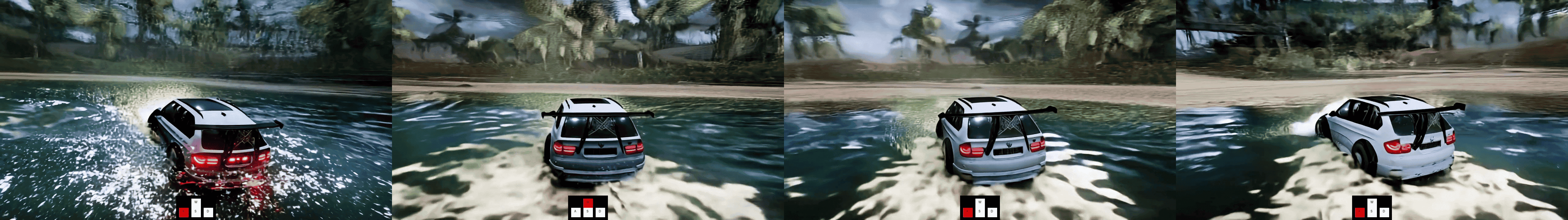}\vspace{-3.7pt}
     \\
     \includegraphics[width=0.2\linewidth]{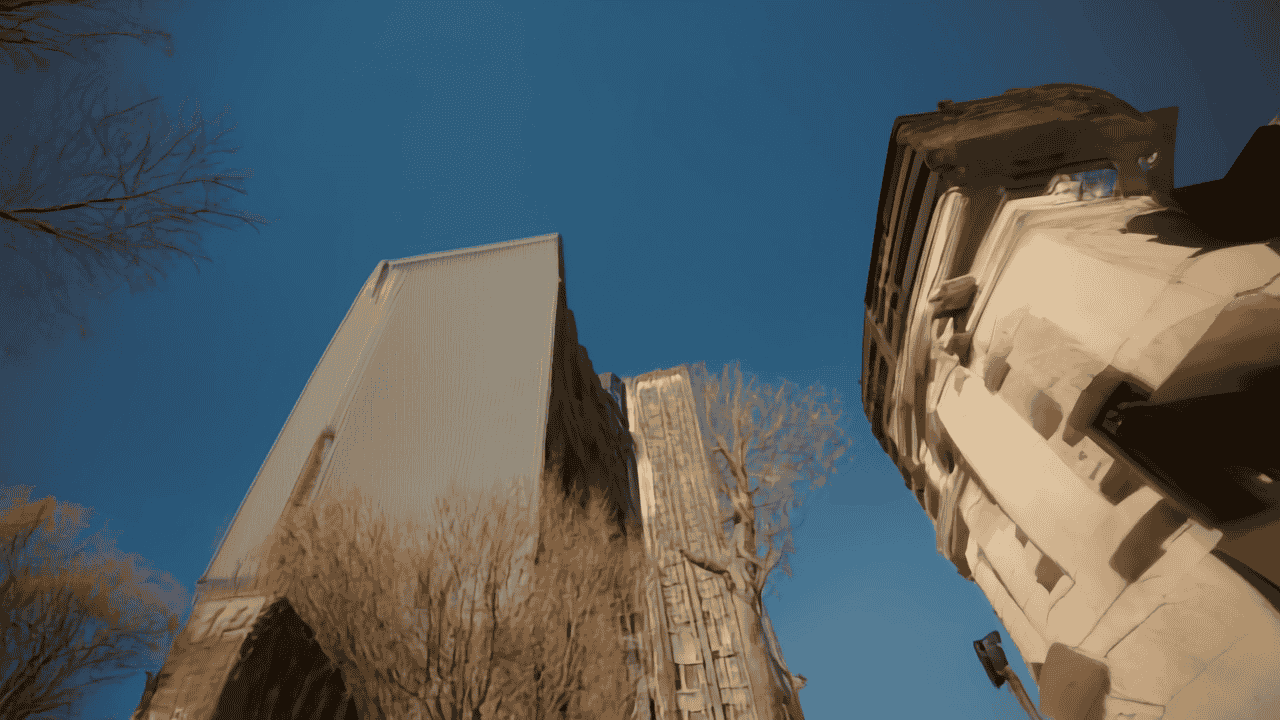}
     \includegraphics[width=0.8\linewidth]{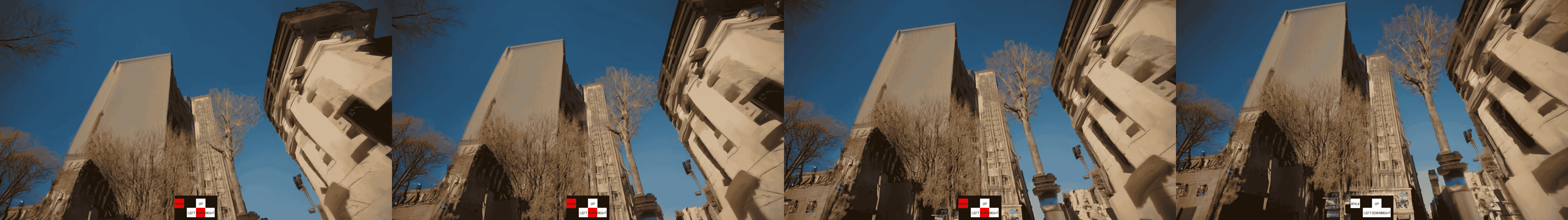}\vspace{-3.7pt}
     \\
     \includegraphics[width=0.2\linewidth]{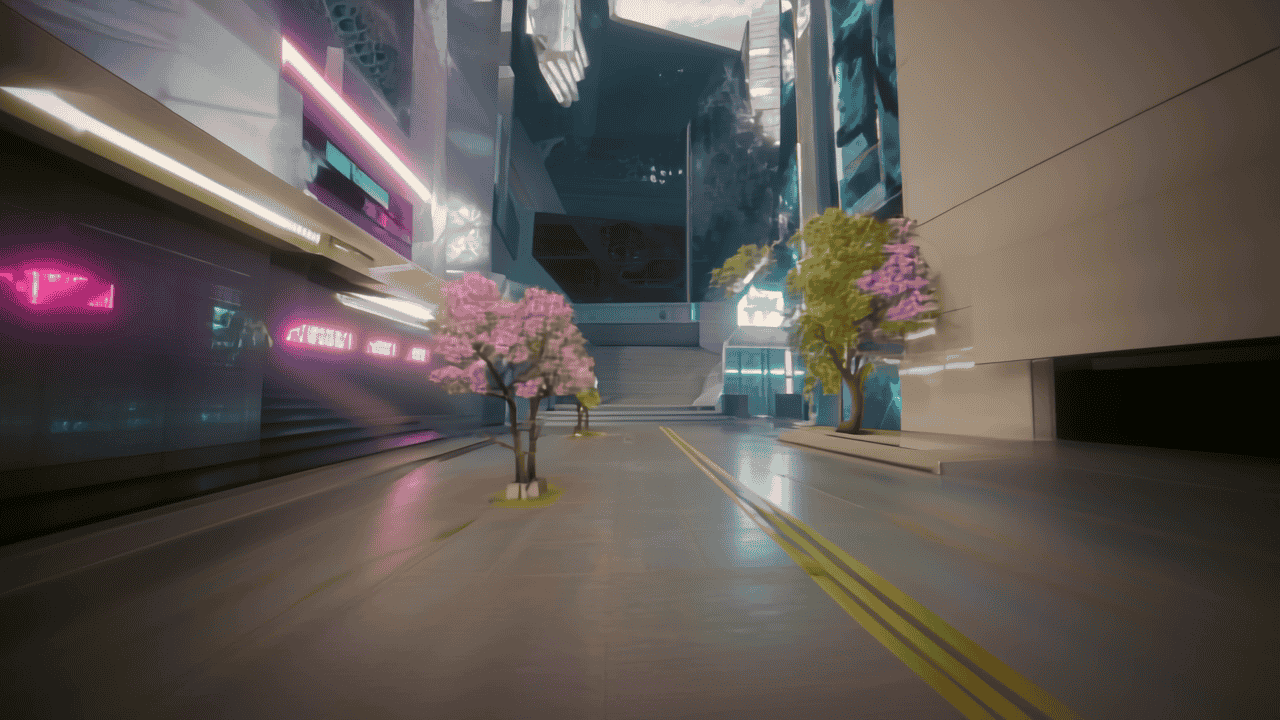}
     \includegraphics[width=0.8\linewidth]{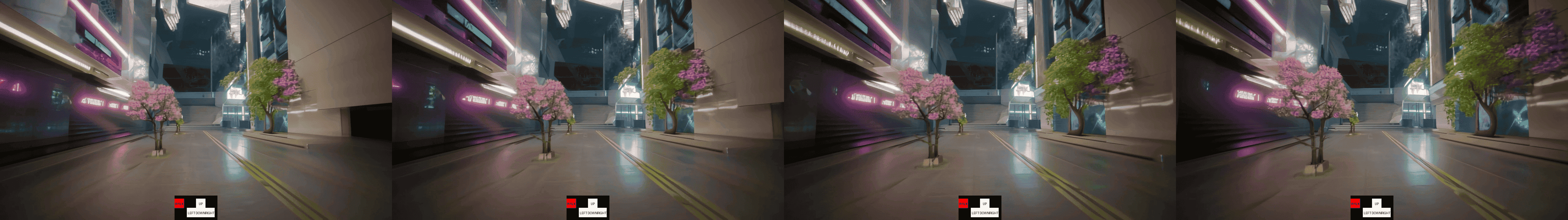}\vspace{-3.7pt}
     \\
     \includegraphics[width=0.2\linewidth]{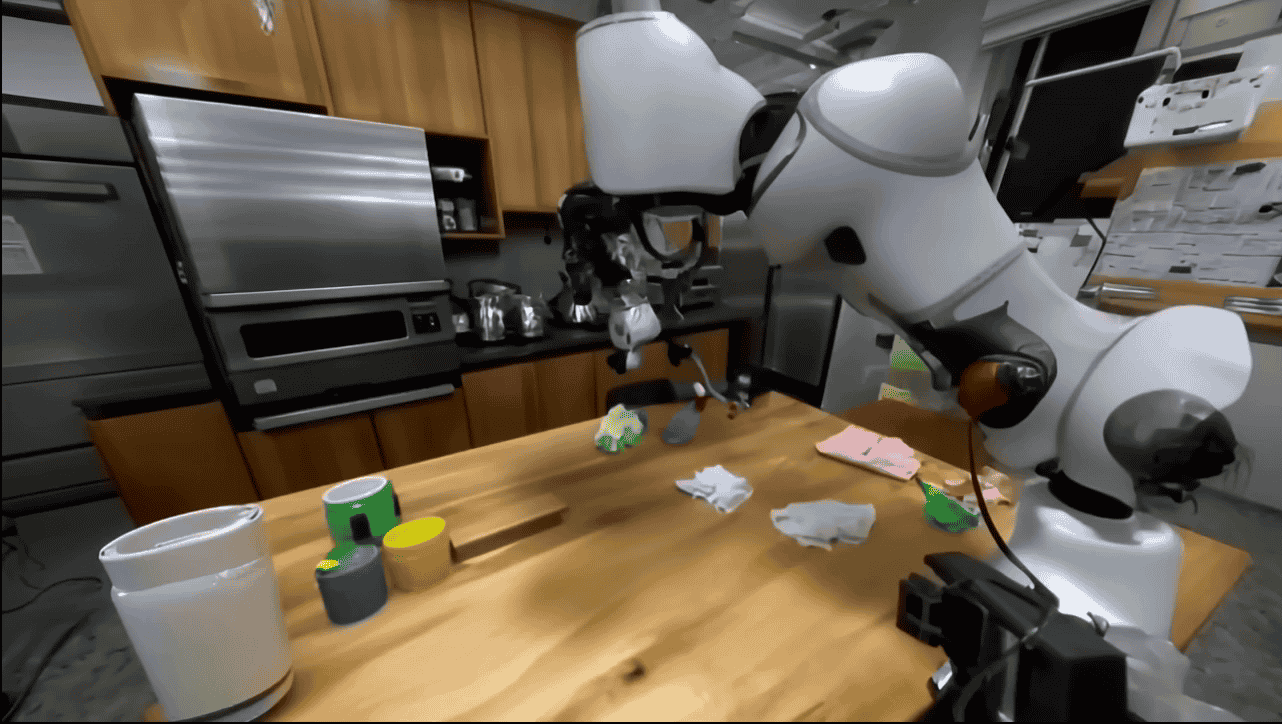}
     \includegraphics[width=0.8\linewidth]{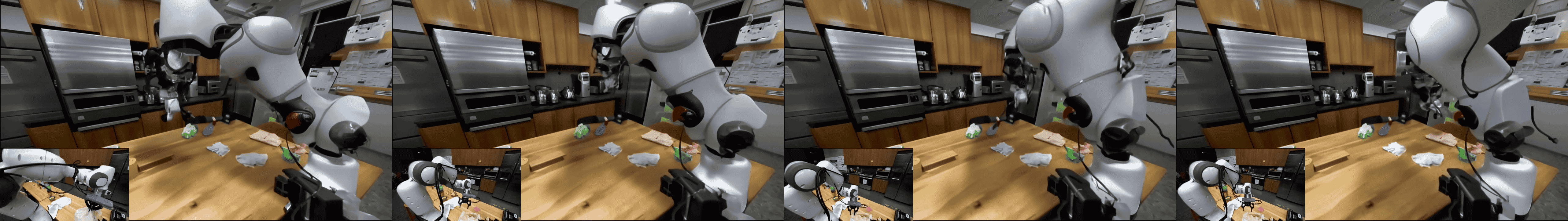}
    \end{tabular}
    \caption{The results demonstrate frame-level precise control achieved by the \controller across diverse scenes, weather conditions, and movement modes.}\label{fig:demo1}
\end{figure*}
\tightpara{Training Details.}
We train \method on the \datasetabbr dataset, using a pre-trained 2.3B parameter DiT model as the backbone, which generates 4 video tokens per second, each decoded into 4 frames by the VAE decoder~\cite{kingma2013auto}. To match this generation rate, we downsample the videos and keyboard inputs in the \datasetabbr dataset accordingly.
For all training cases, we first warm up the base DiT model on unlabeled \datasetabbr data for 20,000 steps with a batch size of 32. Following this, we train the \controller on labeled \datasetabbr data for an additional 20,000 steps with the same batch size, introducing another 0.4B parameter.
Next, we fine-tune \method model using \streamerabbr over 60,000 steps, also with a batch size of 32. For the final Consistency Model distillation, we use the \streamerabbr checkpoints as a teacher model and train the student network for 10,000 steps with a batch size of 32. More details can be found in \secTrain.
\begin{figure*}[htbp]
     \centering
     \begin{subfigure}[b]{1\textwidth}
       \includegraphics[width=\textwidth]{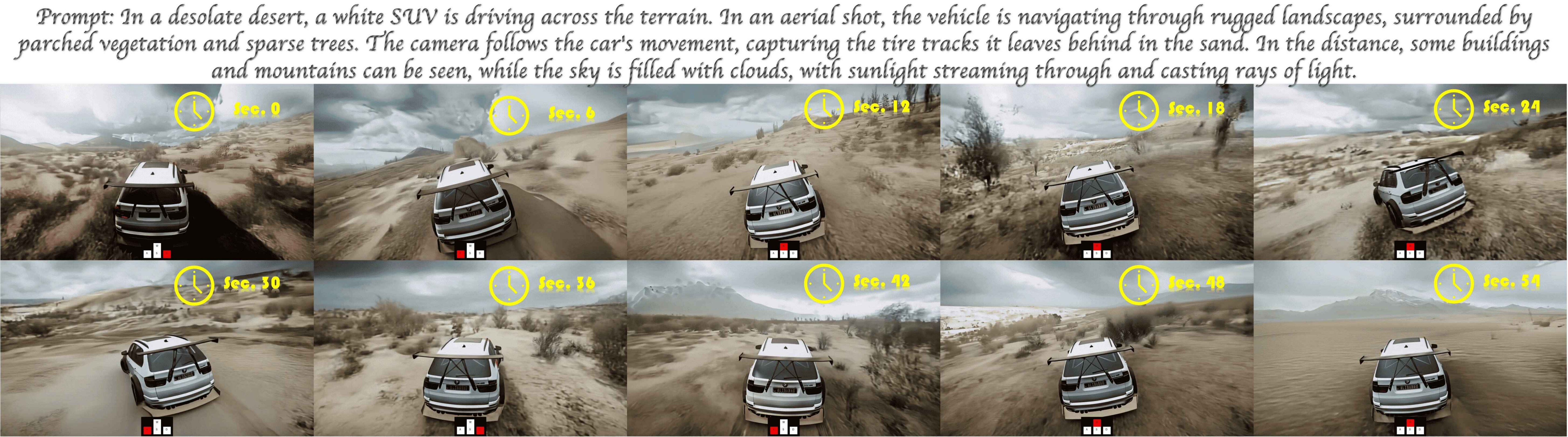}
       \caption{Long 1-minute video generated by \method.}
       \label{fig:longvideo}
     \end{subfigure}
     \hfill
     \begin{subfigure}[b]{1\textwidth}
       \includegraphics[width=\textwidth]{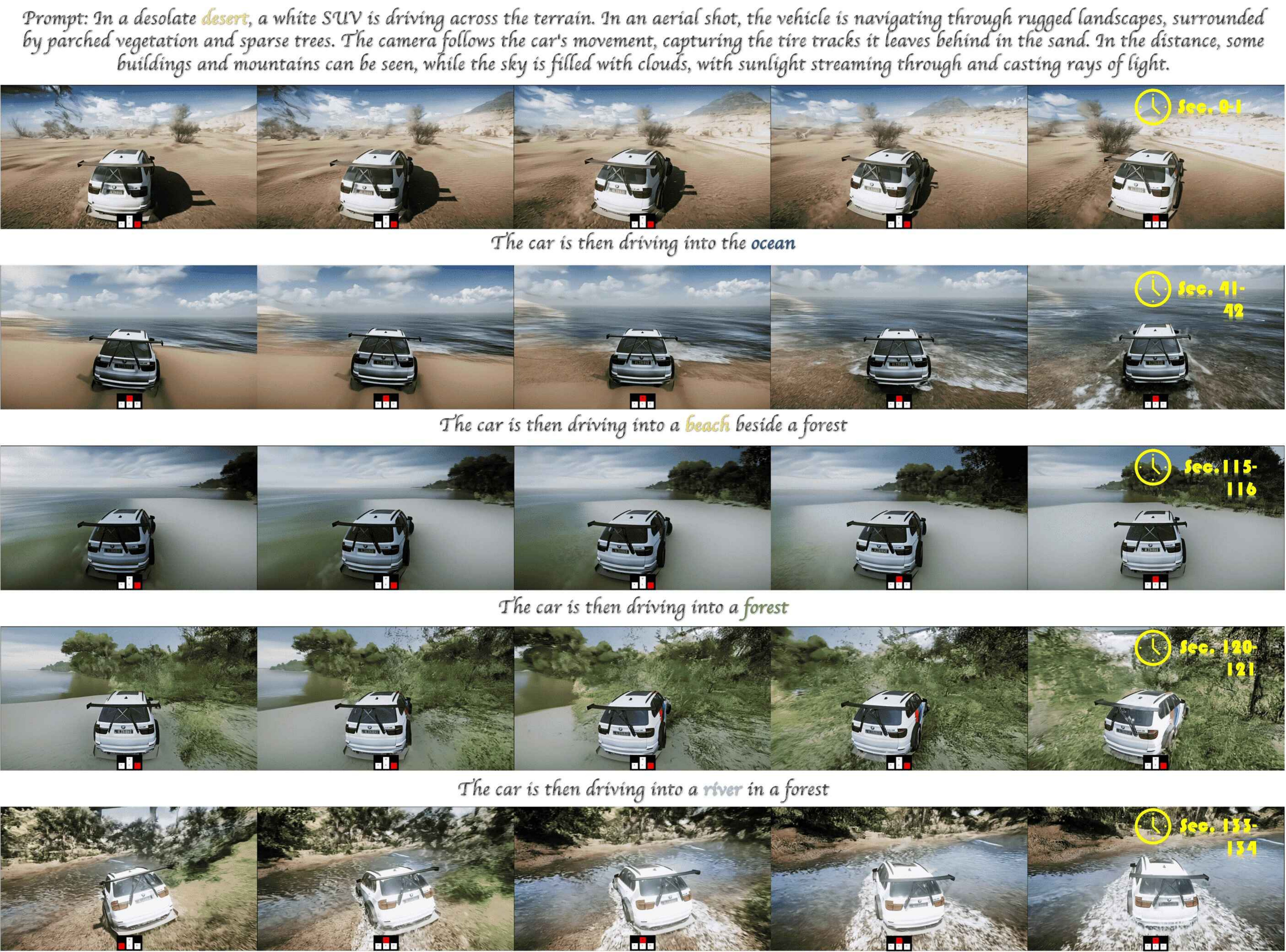}
       \caption{A continuous 2.5-minute video generated by \method, spanning multiple diverse scenes controlled through DiT text prompts.}
       \label{fig:switchscene}
     \end{subfigure}
     \caption{Long worlds generation results by \method. More examples are included in \supVideo.}
     \label{fig:demo2}
   \end{figure*}
\begin{figure*}[htbp]
     \centering
     \begin{subfigure}[b]{1\textwidth}
       \begin{tabular}{c}
     \includegraphics[width=0.2\linewidth]{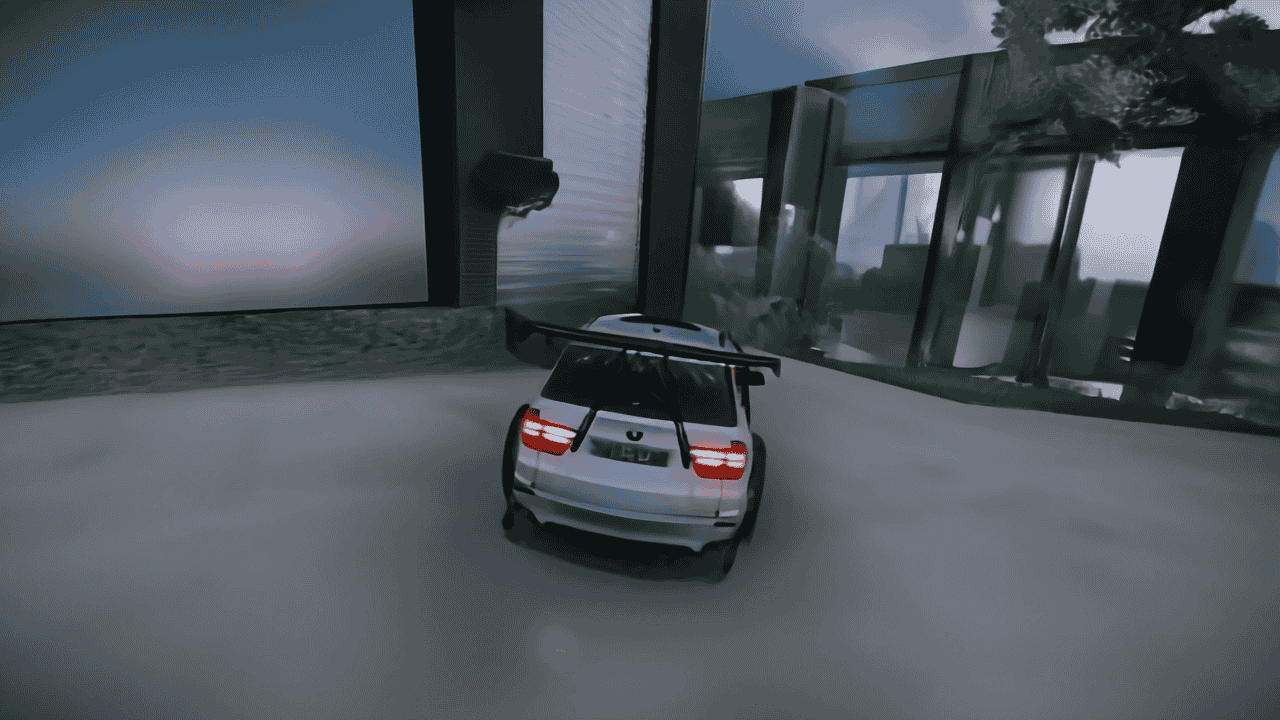}
     \includegraphics[width=0.8\linewidth]{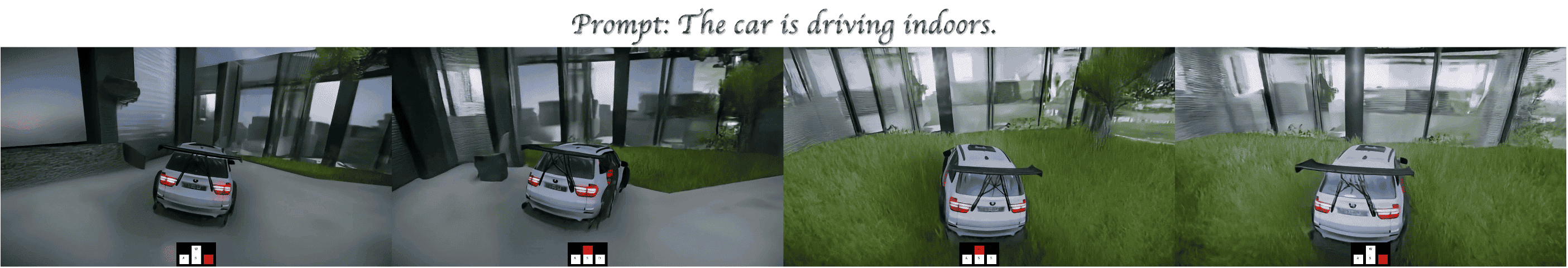}
     \\
     \includegraphics[width=0.2\linewidth]{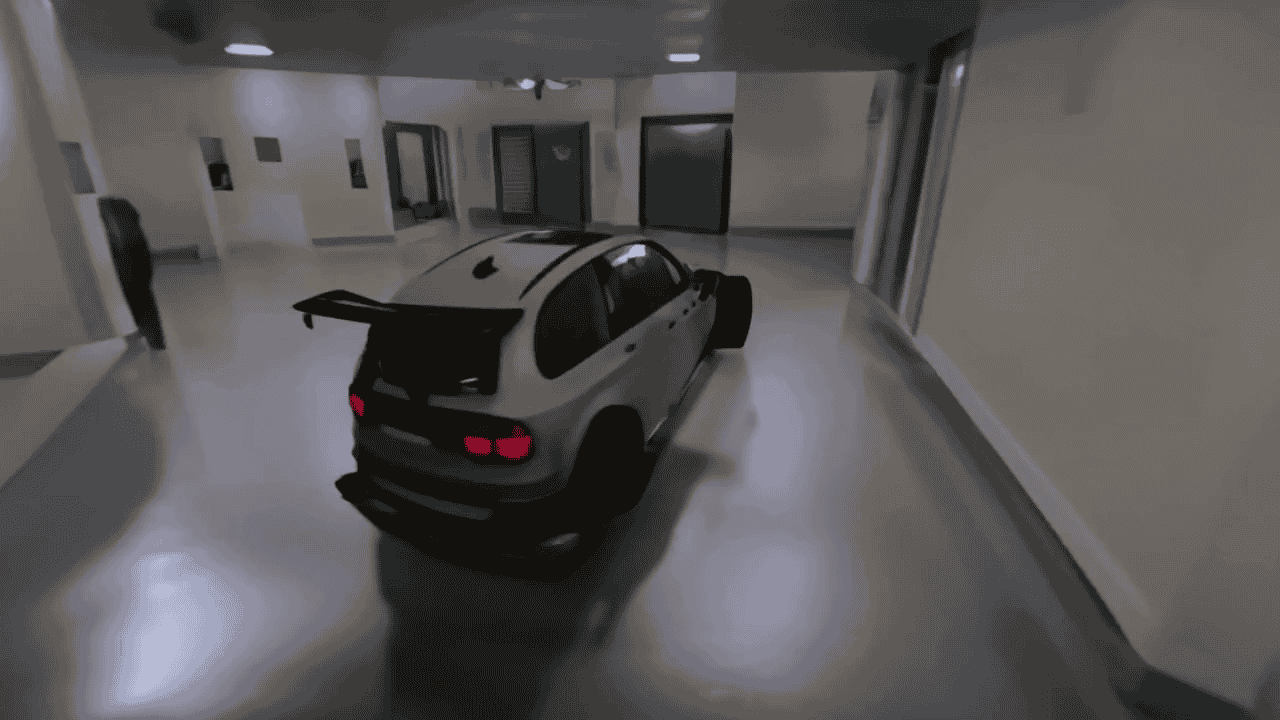}
     \includegraphics[width=0.8\linewidth]{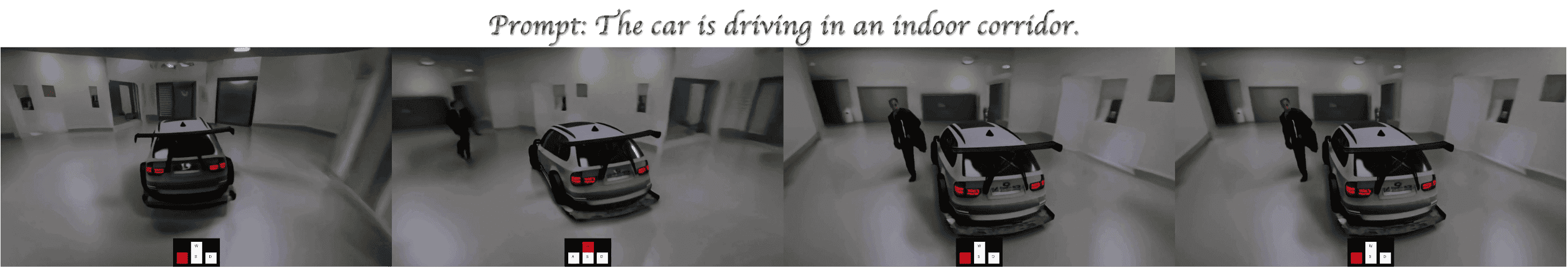}
     \\
     \includegraphics[width=0.2\linewidth]{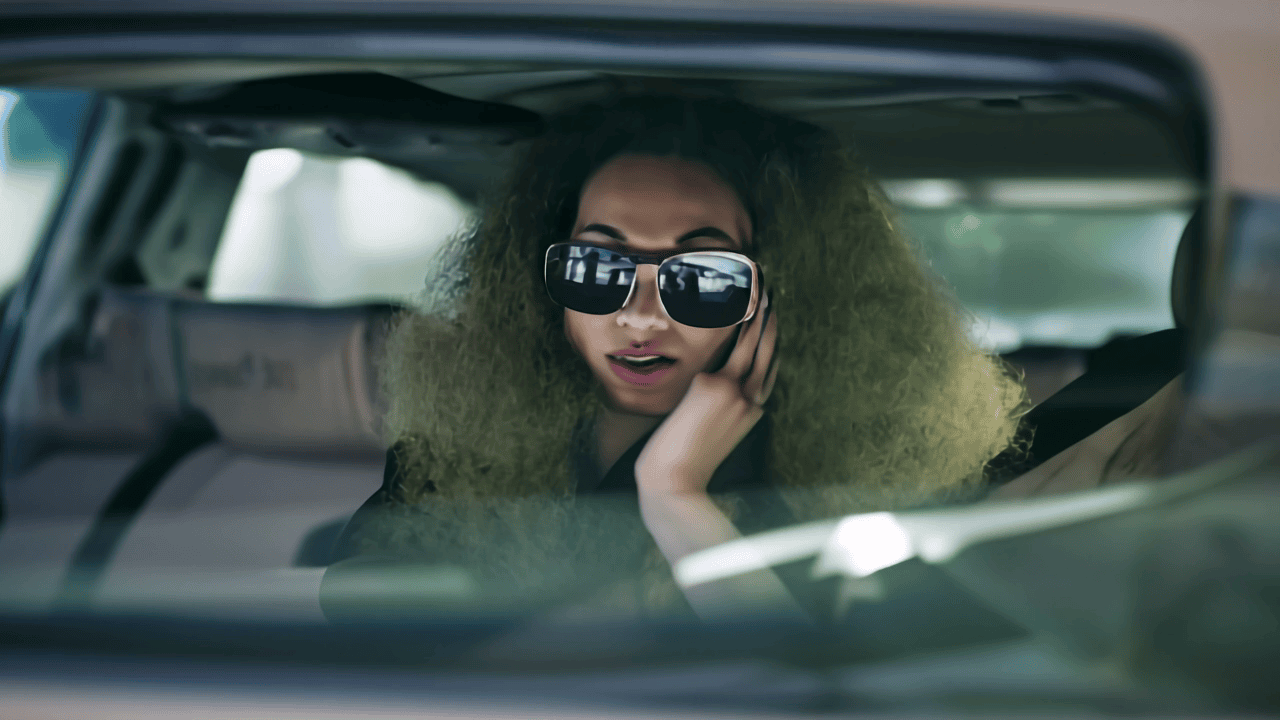}
     \includegraphics[width=0.8\linewidth]{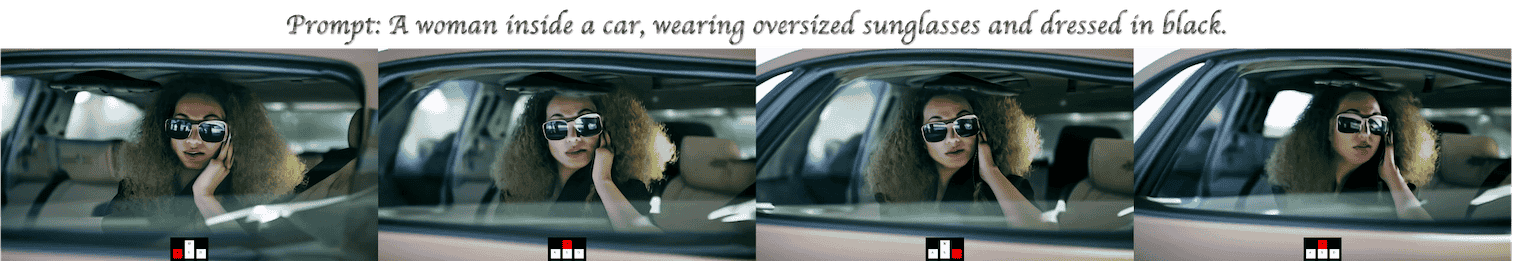}
     \\
     \includegraphics[width=0.2\linewidth]{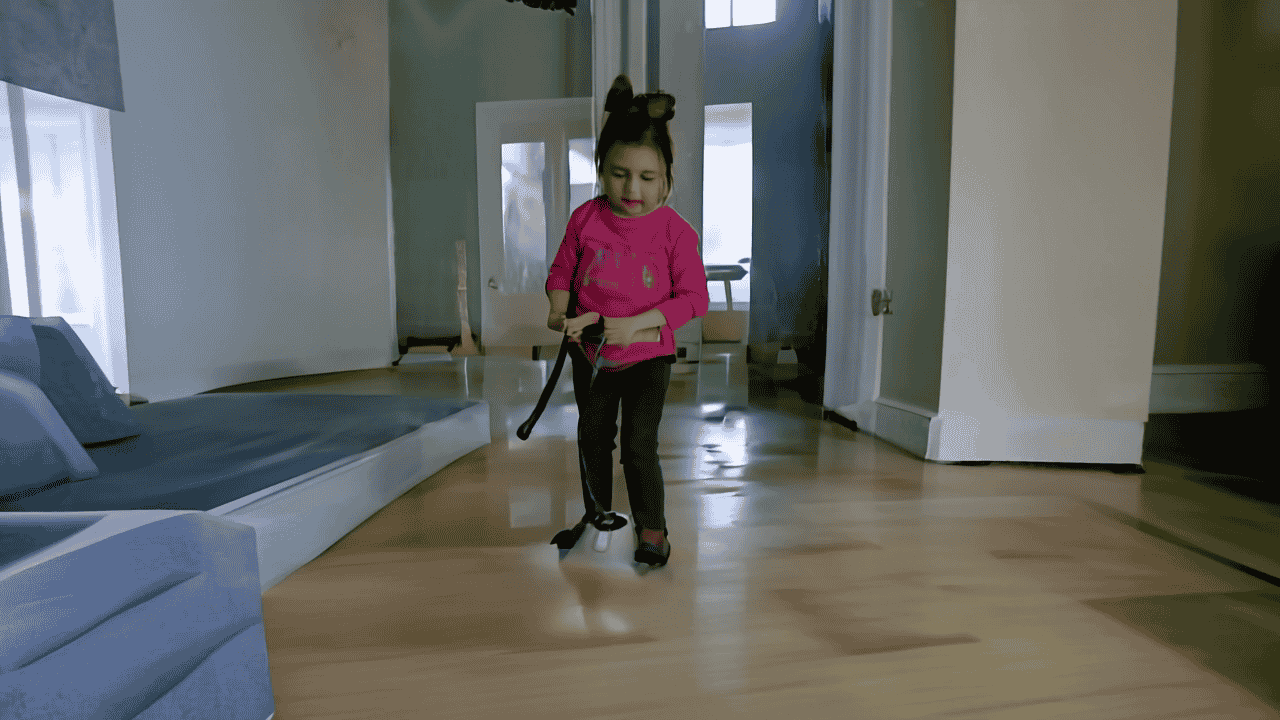}
     \includegraphics[width=0.8\linewidth]{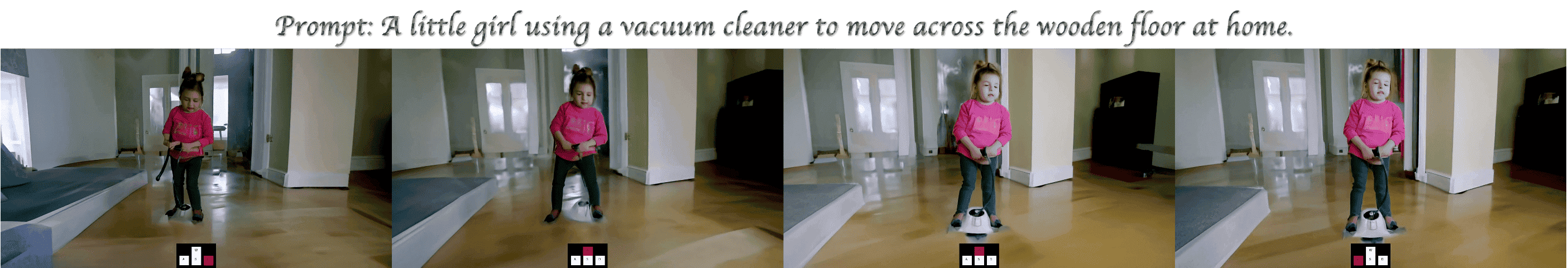}
     
    \end{tabular}
       \caption{\method can generalize its precise movement control to unlabeled scenes and objects, such as driving indoors or making people move as instructed.}
       \label{fig:general_move_and_scene}
     \end{subfigure}
     \hfill
     \begin{subfigure}[b]{1\textwidth}
       \includegraphics[width=\textwidth]{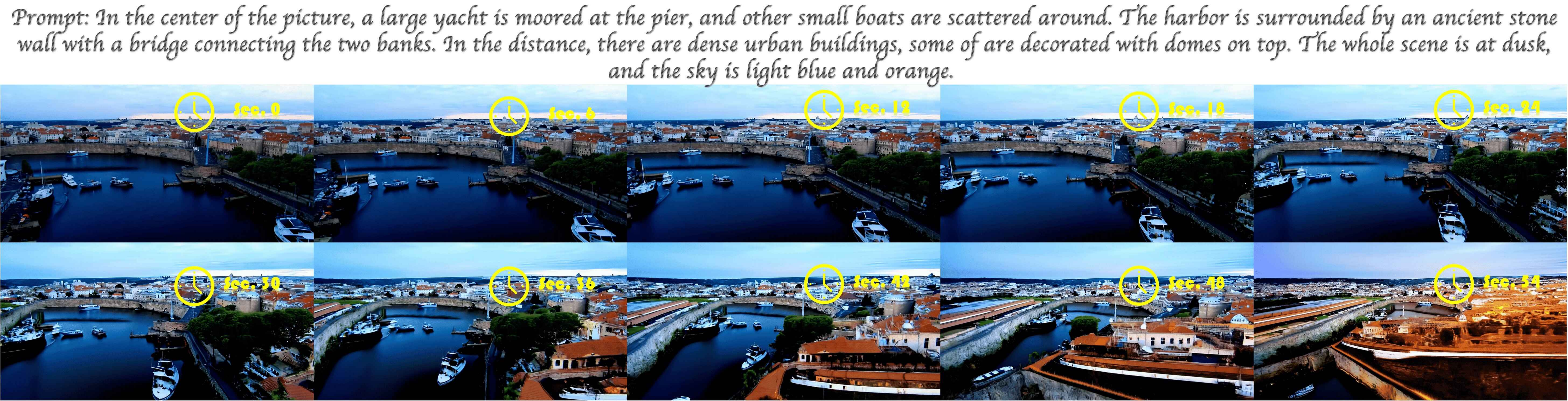}
       \caption{\method can also generate long, general videos by disabling the \controller, acting as a powerful video generator.}
       \label{fig:generallong}
     \end{subfigure}
     \caption{Generalization ability of \method on unseen scenes and objects.}
     \label{fig:generalization}
   \end{figure*}
\medskip
\tightpara{Metrics.} 
We evaluate performance using metrics for both general visual quality and movement control precision. For general visual quality, we use Fr\'echet Inception Distance (FID)~\cite{FID}, Fr\'echet Video Distance (FVD)~\cite{FVD}, and CLIP Score~\cite{CLIP} to assess text alignment. All metrics are evaluated on 2,048 seconds of randomly generated videos. To evaluate movement control precision, we generate 2,048 seconds of video based on keyboard inputs and text prompts from a fixed test set, then measure the Peak Signal-to-Noise Ratio (Move-PSNR)~\cite{PSNR} and Learned Perceptual Image Patch Similarity (Move-LPIPS)~\cite{LPIPS} between the generated videos and real videos with ground truth movements.

\subsection{Precise Frame-Level Interactions}
In this section, we evaluate the effectiveness of the \controller by testing its performance in three distinct scenarios: the \emph{Forza Horizon 5} car driving scenario, the \emph{Cyberpunk 2077} city walking scenario, and a robotic arm task from the \emph{DROID} dataset~\cite{khazatsky2024droid}. We select 50,000 6-second clips from the \emph{DROID} dataset, along with per-frame action labels of joint angles for seven joints, to form the training dataset. More details can be found in \secRobot. The third scenario is specifically designed to assess the effectiveness of \method in embodied AI tasks. For all scenarios, we follow the same training strategy: starting with a pre-trained DiT model, we first perform a warm-up using unlabeled data, followed by fine-tuning the \controller with labeled data.

\medskip
\tightpara{Qualitative Results.}
\cref{fig:demo1} illustrates examples of \method's generated outputs across all scenarios. \method demonstrates the ability to create vivid and dynamic worlds, accurately reflecting user interactions and intentions. It also models the physical behaviors within these environments, such as dust being kicked up when a car drives through a dry desert, or water splashing when it travels through a river. Additional examples of \method's generation capabilities are provided in \secExampleControl.

\medskip
\tightpara{Quantitative Results.}
The last two columns of \cref{tab:metrics_main} present the quantitative evaluation of interaction precision, using LPIPS and PSNR metrics. The results demonstrate that \controller significantly improves control precision, and this enhancement is maintained throughout the subsequent \streamerabbr and \lcmabbr processes.

\subsection{Infinete-Horizon World Generation}
Traditional world simulators focused on precise control often rely on small, auto-regressive generators trained from scratch to minimize the significant memory and time costs associated with pre-trained DiT models. However, this approach compromises visual quality and limits the full potential of world simulators. In this work, we introduce the first world simulator leveraging pre-trained video diffusion models, enabling infinite-length world generation with real-time rendering capabilities. In this section, we present our evaluation of these advancements.

\medskip
\tightpara{Generating Infinitely Long Videos.}
\cref{fig:demo2} showcases examples of generating 1-minute long worlds across diverse scenarios, including desert, river, grassland, snow, and day-to-night transitions. During generation, we switch the DiT prompt to adapt the environment, as shown in \cref{fig:switchscene}. \method's capability extends beyond this; it can generate truly \textbf{infinite-length videos}, with additional \textbf{half-hour} examples available in \supVideo. \cref{tab:metrics_main} reports the video quality and control precision of \method after training with \streamerabbr. While some visual quality is sacrificed, control precision remains strong, and the visual quality still surpasses previous world simulators, achieving a realistic AAA-level standard.

\medskip
\tightpara{Real-Time Rendering.}
We further investigate integrating \lcmabbr with \method. As reported in \cref{tab:metrics_main}, this integration highlights \method's real-time rendering capability, with a slight trade-off in visual quality and minimal loss in control precision, while significantly improving rendering speed from 0.8 FPS to 8 - 16 FPS.

\subsection{Generalization to Out-of-Distribution Worlds}
In addition to superior visual quality, a key advantage of using pre-trained video DiTs is their inherent ability to generalize across diverse scenes. We observe impressive generalization in \method, showcasing the potential of future research into building world simulators with pre-trained DiTs.

\medskip
\tightpara{Generating Unseen Scenes.}
With \method, we can control a car in previously unseen scenes by describing the scenario in the prompt. The first two rows of \cref{fig:general_move_and_scene} demonstrate this capability, where the car is driven through indoor environments, which were not part of the \datasetabbr dataset.

\medskip
\tightpara{Interacting with Unseen Objects.}
A more remarkable feature is \emph{The Matrix}’s ability to generalize interaction with real-world objects. As shown in the last two rows of \cref{fig:general_move_and_scene}, by specifying a human as the center object in the DiT prompt, we can make the person move in response to keyboard inputs.

\medskip
\tightpara{Generating Long Videos without Moving Control.}
Though \method is trained on the \datasetabbr dataset, it can also function as a general long video generator. By disabling the \controller and using only the DiT backbone trained after \streamerabbr, \method can generate long videos corresponding to ordinary prompts. \cref{fig:generallong} shows such an example, further proving \method's strength as a realistic world simulator.



\medskip
\section{Conclusion}
We introduce \method, a real-world simulator capable of generating infinitely long, high-fidelity video streams with precise real-time control. Trained on a blend of AAA game data and real-world footage, \method supports immersive exploration of dynamic environments, with zero-shot generalization to unseen scenarios. Operating at 8 - 16 FPS, it enables continuous, interactive simulations across diverse terrains, bridging the gap between virtual and real-world applications. This work highlights the potential of using game data to build robust world models with minimal supervision, and showcases the power of pre-trained video DiTs in enabling realistic, large-scale simulations.
\medskip

\section*{Acknowledgments}
We would like to express our gratitude to the Alibaba Cloud Elastic Computing-Heterogeneous Hypervisor \& Instance team for providing the Remote Gaming Console (RGC), which facilitates our data collection process. Special thanks to Bo Li, Min He, and their colleagues for their invaluable support and assistance. We also extend our appreciation to Microsoft and CD Projekt Red for developing their remarkable games, which played a key role in our data collection. A special acknowledgment goes to the Night City and Mr. V, who provided a peaceful environment for our research. Finally, we thank Professor Dr. Ping Luo of the University of Hong Kong for his insightful guidance on the collection of robotics data.
\clearpage
\maketitlesupplementary

\appendix
\newcommand{\AppendixPrefix}{A}
\setcounter{section}{0}
\renewcommand{\thefigure}{\AppendixPrefix\arabic{figure}}
\setcounter{figure}{0}
\renewcommand{\thetable}{\AppendixPrefix\arabic{table}}
\setcounter{table}{0}
\renewcommand{\theequation}{\AppendixPrefix\arabic{equation}}
\setcounter{equation}{0}

\section{Details in Experiments}

\subsection{DiT Backbone}
The DiT backbone is adapted from the publicly available DiT models~\cite{opensora}. It consists of a patch embedding module, a caption embedding module, a timestep embedding module, 32 DiT blocks, followed by a linear output head with LayerNorm~\cite{ba2016layer}. The followings provide details of each module within the DiT backbone.

\medskip
\tightpara{The Patch Embedding Module.}
The patch embedding module employs a 3D convolution with a kernel size of $1\times2\times2$, followed by a reshape operation. Thus, the convolution can effectively process the video latent from the VAE encoder, and the reshape operation can further transform the feature into a sequence of tokens with 2,048 feature size. By using a 3D convolution, the module captures both spatial and temporal features, ensuring that the token sequence retains essential information from the video data.

\medskip
\tightpara{Caption Embedding Module.}
The caption embedding module takes the caption token sequence encoded by the T5 model and further processes it through a two-layer FFN. Both the hidden feature size and the output feature size are set to 2,048, allowing the module to generate rich and high-dimensional representations of the caption data.

\medskip
\tightpara{Timestep Embedding Module.}
The timestep embedding module is implemented as a sinusoidal embedding module followed by a two-layer FFN. Both the hidden feature size and the output feature size of this FFN are set to 2,048.

\medskip
\tightpara{DiT Block.}
Each DiT block includes a self-attention layer operating on network features, a cross-attention layer linking conditions with self-attention outputs, and an FFN layer composed of two linear layers with a GELU activation~\cite{hendrycks2016bridging} in between.

\subsection{Training Details}
Upon obtaining the base DiT model, the training process consists of four distinct stages: (1) warm-up on unlabeled \datasetabbr, (2) training of the \controller, (3) fine-tuning using \streamerabbr, and (4) \lcm distillation. Below, we first outline the common training configurations utilized across all stages, followed by a detailed description of each individual phase.

\medskip
\tightpara{Common Settings.}
All training procedures were executed with an overall batch size of 32 and a learning rate of $1 \times 10^{-5}$
 . Mixed-precision training was employed using bfloat16 to enhance computational efficiency. During preprocessing, all video inputs were resized to a resolution of $1280\times 720$ pixels and set to 16 FPS. For sequences exceeding 25,200 frames in length, we used the Deepspeed Ulysses sequence parallelism strategy~\cite{jacobs2023deepspeed}, distributing the sequence across 8 GPUs to manage memory and computational demands effectively.

\medskip
\tightpara{Warm-Up on Unlabeled \datasetabbr Dataset.}
In the initial warm-up stage, we fine-tuned all linear layers of the base DiT model using Low-Rank Adaptation (LoRA) to tailor the model to the source data distribution~\cite{hu2021lora}. The LoRA rank was set to 128, and the model was trained for 20,000 steps. This adaptation ensures that the model parameters are suitably adjusted to the characteristics of the unlabeled source dataset before advancing to subsequent training phases.

\medskip
\tightpara{Training of \controller.}
The second stage focuses on training the \controller, each of which is integrated after every two consecutive DiT blocks, totaling 16 \controller. During this phase, the parameters of the base DiT model were frozen to concentrate the training solely on the \controller. This stage was conducted over 20,000 training steps, enabling the \controller to effectively interface with the base model without altering its foundational parameters.

\medskip
\tightpara{Fine-Tuning Using \streamerabbr.}
The third stage involves comprehensive fine-tuning of all model parameters, including both the base DiT model and the \controller, utilizing the \streamerabbr approach. This extensive fine-tuning was carried out over 60,000 steps, allowing for the refinement and optimization of the entire model to better capture data intricacies and enhance overall performance.

\medskip
\tightpara{Consistency Model Distillation.}
In the final stage, consistency model distillation was performed using the model from the preceding fine-tuning phase as the teacher model. The student model was initialized with the teacher's weights to facilitate knowledge transfer. During distillation, we employed a one-stage guided distillation technique~\cite{luo2023latent}, incorporating Classifier-Free Guidance (CFG) into the student model. For the Ordinary Differential Equation (ODE) solver within the consistency distillation framework, we utilized the Euler solver with a single-step size of $25 / 1000$. This distillation process was conducted over 10,000 training steps.
\section{The \datasetabbr Dataset}

\subsection{The \textbf{\dataengine}}\label{app:platform}
We build a framework, \dataengine, for data collection. The framework consists of three components: Controlling, Simulation, and Observation.

\medskip
\tightpara{Controling.} In most games, we need to control a character to go to different scenes and make interactions. Intrinsically, the collected data can be reconstructed with initial states of game worlds and a series of control signal. In order to make the collected data clean and meaningful, instead of being stuck in one corner, we designed two different control systems, namely the automatic one and the manual one. For the automatic control system, we use Cheat Engine for pivotal data access, such as XYZ coordinates in games. These data can be used to determine whether the game has been stuck for some time. We detect the coordinates of a past period of time and determine whether they are covered in a circle of a given size. If the game is detected as stuck, we will reset the game state and restart the recording. Generally, the automatically generated control signals will move randomly, change direction, and change perspective. This is good enough for games that move on a 2D-like surface. However, for games moving in a 3D space, random signals will struggle with generating meaningful content, so we have to change to the manual system. Since our game is running and captured on cloud servers, human data collectors will observe the game through a low-definition streaming and control manually. Signals (from keyboards, mice, and gamepads) are translated and delivered through the socket server, and cloud servers will generate keyboard events through the virtual keyboard. Here, the latency between the control signals and the OBS screen recording is crucial. We eventually found that the control signals recorded on the cloud sever and the actual action responses in the recorded videos were generally no more than three frames apart. and in general, this delay is stable and can be subtracted directly from the timeline.

\medskip
\tightpara{Simulation.} The game runs directly on the cloud servers. we can directly copy the server images to get a large number of running instances. The recorded videos and control signals will be uploaded to the data center. We set up a series of video quality checks to filter out samples of low quality (still or overly noisy videos, and some undefined scenes). All games run at the highest quality while ensure the OBS screen recording does not get stuck. In order to avoid overly complicated situations, we removed the NPCs and running vehicles in the game. We use the Reshade to adjust the game scenes to make it more reality-like.

\medskip
\tightpara{Observation.} We use OBS as the screen recorder. One can use scripts to control OBS for automatic recording. We recorded the game at native resolution of $2560\times1600$ (higher resolutions may cause the game and recording to lag). For the reality of the recorded videos, we removed GUIs and texts in the game through a Reshade plugin, namely ReshaderEffectShaderToggler.\footnote{https://github.com/4lex4nder/ReshadeEffectShaderToggler} It can turn off the rendering of GUI related shaders in the game while left the native video untouched. 

\medskip
\tightpara{Forza Horizon 5.} In Forza Horizon 5, a telemetry mechanism can be used for game status retrieving. We can access the real-time game data through socket after checking on the telemetry option in settings. An example script for data listening can be found here.\footnote{https://github.com/jasperan/forza-horizon-5-telemetry-listener} We can access XYZ coordinates, velocities and accelerations. We use these data for stuck detection and sample filtering. Since Forza Horizon 5 is a game that mainly takes on 2D area, we apply automatic pipeline that randomly walking on different game scenes (like dessert, grassland, the watery and the snowy areas). Control signals are simplified to going forward, turning left and turning right. During the data collection stage, if XYZ of is still for several seconds, the controller will try to move back. And if the $40$ position points collected during the last $40$ seconds can be covered with a circle with radius of $80$ meters, the controller will try to teleport the car to a random position. After raw sample collection, we apply some strategies to filter out samples of low quality. We use the acceleration data to detect if the car has collided with anything, and drop these video clips with collision. Sometimes the car is moving backward while the controlling input is moving forward, this is because the direction of movement in the game is to provide acceleration. We filter out data with a large angle between acceleration and velocity. Due to some problems in the game itself, the video often changes suddenly at some time. We filter out video clips with large average error between any two adjacent frames.

\medskip
\tightpara{Cyberpunk 2077.} Cyberpunk 2077 is a game that offers realistic visuals and lighting effects. Due to the complexity of the game terrain, we have to choose the manual pipeline. For simplicity, the actions in game are reduced to two separated inputs. The first one makes the character move forward or stop. And the second one makes the direction of the character's sight move up, down, left and right. We disable the NPCs and moving vehicles with game mod. During the data collection, players observe the game through low-definition OBS streaming and send control signals. The signals are then mapped into ``W'' (moving forward) / ``U'' ``D'' ``L'' ``R'' (up, down, left and right) on the cloud servers. We access and record the XYZ coordinates of player through Cheat Engine. These coordinate sequences are then used for filtering out video clips where collisions occur between the character and the game scene.

\subsection{The \textbf{\datasetabbr} Dataset}

We present the \textbf{\datasetabbr} dataset from three perspectives: basic information, the annotation method used to convert the original data from \textbf{\dataengine} to our desired format, and the filtering method applied to remove undesirable data.

\subsubsection{Basic Information}\label{app:basic_info}
\begin{figure*}[t]
\centering
\includegraphics[width=\linewidth]{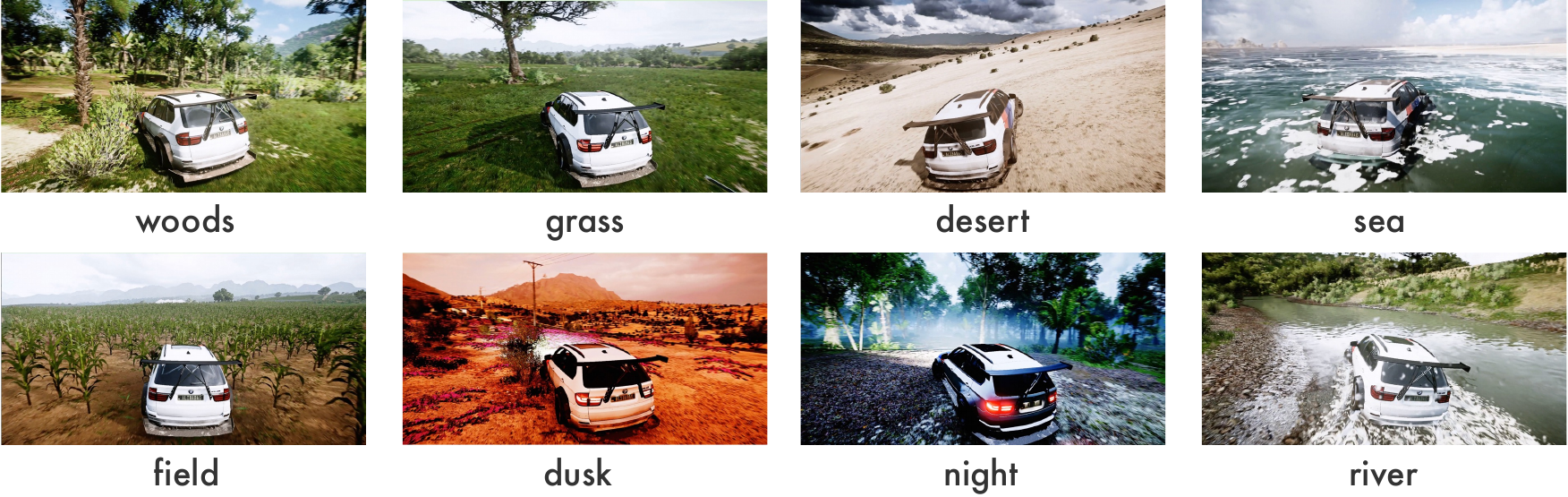}
\vspace{-20pt}
\caption{
Examples of Horizon5 across different scenarios.
}
\label{fig:app-categories}
\vspace{-2pt}
\end{figure*}

The \textbf{\datasetabbr} comprises data from both Forza Horizon 5 and Cyberpunk 2077. For Forza Horizon 5, we collected approximately 1,200,000 pairs of video and control signals, while for Cyberpunk 2077, we gathered around 1,000,000 such pairs. All collected videos have a duration of approximately 6 seconds, recorded at 60 FPS. For Forza Horizon 5, we specifically collected data from multiple scenes, including deserts, oceans, water bodies, grasslands, and fields. The videos from different scenes are illustrated in \cref{fig:app-categories}, along with the distribution of data volume for each scene \cref{fig:app-distribution} (a). For Cyberpunk 2077, we focused on gathering data from urban environments that feature a significant number of tall buildings. 

In Forza Horizon 5, the dataset includes only three distinct control signals: ``moving forward'' (denoted by ``D''), ``moving forward and turning left'' (denoted by ``DL''), and ``moving forward and turning right'' (denoted by ``DR''). In contrast, the data for Cyberpunk 2077 encompasses five different control signals: ``moving forward'' (denoted by ``W''), ``turning left'' (denoted by ``L''), ``turning right'' (denoted by ``R''), ``looking upward'' (denoted by ``U''), and ``looking downward'' (denoted by ``D'').

\subsubsection{Annotation Methods}
The original data from \textbf{\dataengine} typically has a duration of around 10 minutes, which is excessively long for training \method. Therefore, we use FFmpeg~\citep{ffmpeg} to segment these videos into 6-second clips. Next, we extract the corresponding control signals from the complete set of signals. We then use InternVL~\citep{chen2024internvl} to generate captions based on 12 uniformly extracted key frames from the videos. After the captioning process with InternVL, we perform manual corrections on the generated captions to eliminate obvious errors.

\subsubsection{Filtering Methods}
After the annotation step, a significant amount of undesired data remains, which could disrupt the training of \method. To address this, we employ five filtering methods to eliminate these problematic data points, which we introduce as follows. Note that for Cyberpunk 2077, since we utilize human data collection rather than automatic methods, many of the following issues do not exist.

\begin{figure*}[t]
\centering
\includegraphics[width=\linewidth]{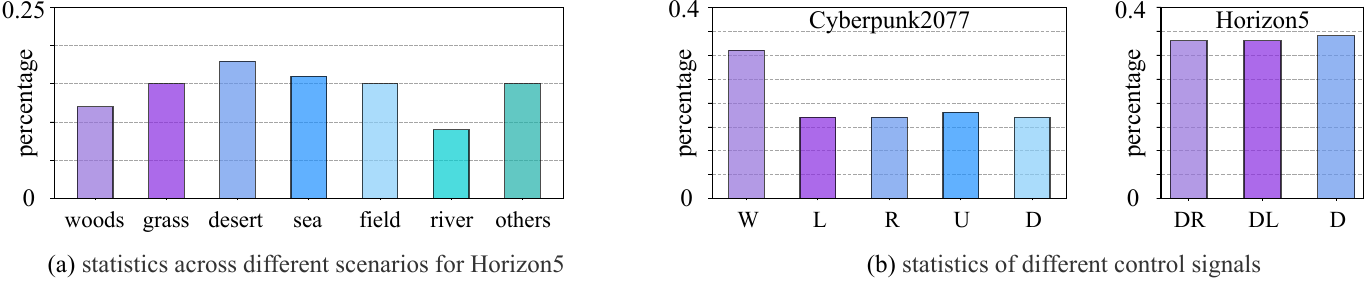}
\vspace{-20pt}
\caption{
(a) The statistics for Horizon5 across different scenarios include woods, grass, desert, sea, fields, rivers, and others. The results indicate that the quantity of data across these scenarios is relatively balanced.
(b) The statistics of different control signals for Cyberpunk 2077 and Horizon5. In Cyberpunk 2077, the percentage of the "move forward" signal is relatively high, while other steering control signals are distributed more evenly. In Horizon5, all three control signals are evenly distributed.
}
\label{fig:app-distribution}
\vspace{-2pt}
\end{figure*}

\medskip
\noindent\textbf{Balance Control Signals.}
Balancing the number of different control signals is beneficial for the training of \method. The process of balancing control signals consists of three steps: 1) First, we analyze the distribution of control signals for each 6-second video and record the results. 2) Next, we assess the overall distribution of control signals across the entire dataset to identify the most frequently occurring control signal. 3) Finally, we remove some data points that contain the highest proportion of this predominant control signal. We repeatedly implement the second and third steps until the distribution is relatively balanced. The distribution results for Forza Horizon 5 and Cyberpunk 2077 are reported in \cref{fig:app-distribution} (b). We provide the pseudocode for the algorithm in \cref{alg:control_signal_balancing}.

\begin{algorithm}
\caption{Control Signal Balancing Algorithm}
\label{alg:control_signal_balancing}
\begin{algorithmic}[1]
\REQUIRE Dataset $D$ containing control signals from 6-second videos
\ENSURE Balanced Dataset $B$
\STATE Initialize $B$ as an empty set
\FOR{each video $v$ in $D$}
    \STATE Analyze the distribution of control signals in $v$
    \STATE Record the results for $v$
\ENDFOR

\WHILE{not isBalanced($B$)}
    \STATE overallDistribution $\gets$ Assess the overall distribution of control signals in $D$
    \STATE mostFrequentSignal $\gets$ Identify the most frequently occurring control signal from overallDistribution
    \STATE $D \gets$ Remove data points from $D$ that contain mostFrequentSignal
\ENDWHILE

\STATE Set $B \gets D$
\RETURN $B$
\end{algorithmic}
\end{algorithm}

\medskip
\noindent\textbf{Detect and Remove the Data with Collisions.}
In Forza Horizon 5, randomly generated control signals often cause the car to collide with walls or rocks. Additionally, the car may be struck by other vehicles. These collisions can severely disrupt the training process, making it essential to identify and remove collision-affected data. Our analysis revealed that collisions consistently result in abrupt changes in acceleration over a very short time. Thus, we use significant variations in acceleration as a reliable indicator of collision events and discard any corresponding data to maintain the integrity of the training process.

\medskip
\noindent\textbf{Detect and Remove Stuck Data.}
In Forza Horizon 5, after colliding with walls or rocks, the car often gets stuck; even when the ``D'' key is pressed, the car fails to move. This stuck situation complicates the training data and negatively impacts the performance of \method. Therefore, we need to detect and remove such instances. Detecting when the car is stuck is relatively straightforward—we simply calculate the distance the car has traveled within the video. If this distance falls below a certain threshold, we conclude that the car is stuck and discard the corresponding data.

\medskip
\noindent\textbf{Detect and Remove the Data with Mismatched Motion and Control.}
As introduced in \cref{app:platform}, to quickly resolve a stuck situation, the car will move backward when stuck. As a result, it is possible for the car to still move backward at a slower speed even when the ``D,'' ``DL,'' or ``DR'' keys are pressed. Similar situations may arise when ``DL/DR'' is pressed for a long period and then switched to ``DR/DL.'' Although the acceleration is directed to the right/left, the car may continue to move in the opposite direction for a brief period. We refer to this as mismatched motion and control, which complicates the training process. To address this issue, we calculate the directions of both the acceleration and the car's movement. If the angle between these two directions is too large, we discard the corresponding data.

\medskip
\noindent\textbf{Detect and Remove Artifacts.}
In Forza Horizon 5, visual artifacts can occur when a car collides with obstacles like trees, introducing distortions into the generated videos. To filter out such corrupted data, we detect variations in pixel values across consecutive frames. Our analysis shows that applying a high threshold effectively identifies all videos containing these artifacts, enabling their removal.

\subsection{The \textbf{\emph{DROID}} dataset}

\subsubsection{Basic Information}
DROID is a large, diverse robot manipulation dataset containing 76k demonstration trajectories (350 hours of interaction) collected across 564 scenes and 86 tasks over 12 months by 50 collectors worldwide. It aims to improve the performance, robustness, and generalization of robotic manipulation policies. 
DROID uses the same hardware setup across all 13 institutions to streamline data collection while maximizing portability and flexibility. The setup consists of a Franka Panda 7DoF robot arm, two adjustable Zed 2 stereo cameras, a wristmounted Zed Mini stereo camera, and an Oculus Quest 2 headset with controllers for teleoperation. Everything is mounted on a portable, height-adjustable desk for quick scene changes.
\subsubsection{Filtering Methods}

\medskip
\noindent\textbf{Remove Overly Complex Scenes and Balance Different Scenes}
DROID is a collaborative effort involving multiple laboratories and contains data from 11 different environments, including domestic scenes like kitchens and bedrooms, as well as industrial settings such as factories and laboratories. Due to the complexity of these scenes, which poses challenges for subsequent captioning and video learning, we first classify the data based on scene labels. For each category, we manually select 50 less complex scenes. We then use DINO to encode and extract semantic features to calculate the mean, removing outliers within each scene based on this semantic mean. To balance the number of training samples across scenes, we ensure that the final dataset contains an approximately equal number of samples from each scene after outlier removal.

\medskip
\noindent\textbf{Filtering Frames Without Arm Presence and Removing Failed Executions}
Since some frames in the videos do not contain the robotic arm or have only a small visible area, we use Grounding DINOv2~\cite{liu2023grounding} to remove such frames. If more than 20\% of the frames in a video meet this condition, the entire video is discarded. Additionally, to ensure accuracy in control, we remove data where the robotic arm fails to follow the intended trajectory successfully. Finally, we use the spatial position of the robot gripper as a condition for each frame in the training.

\section{More Examples}
\subsection{Precise Frame-Level Control}
\begin{figure*}
    \centering
    \begin{tabular}{c}
     \includegraphics[width=0.2\linewidth]{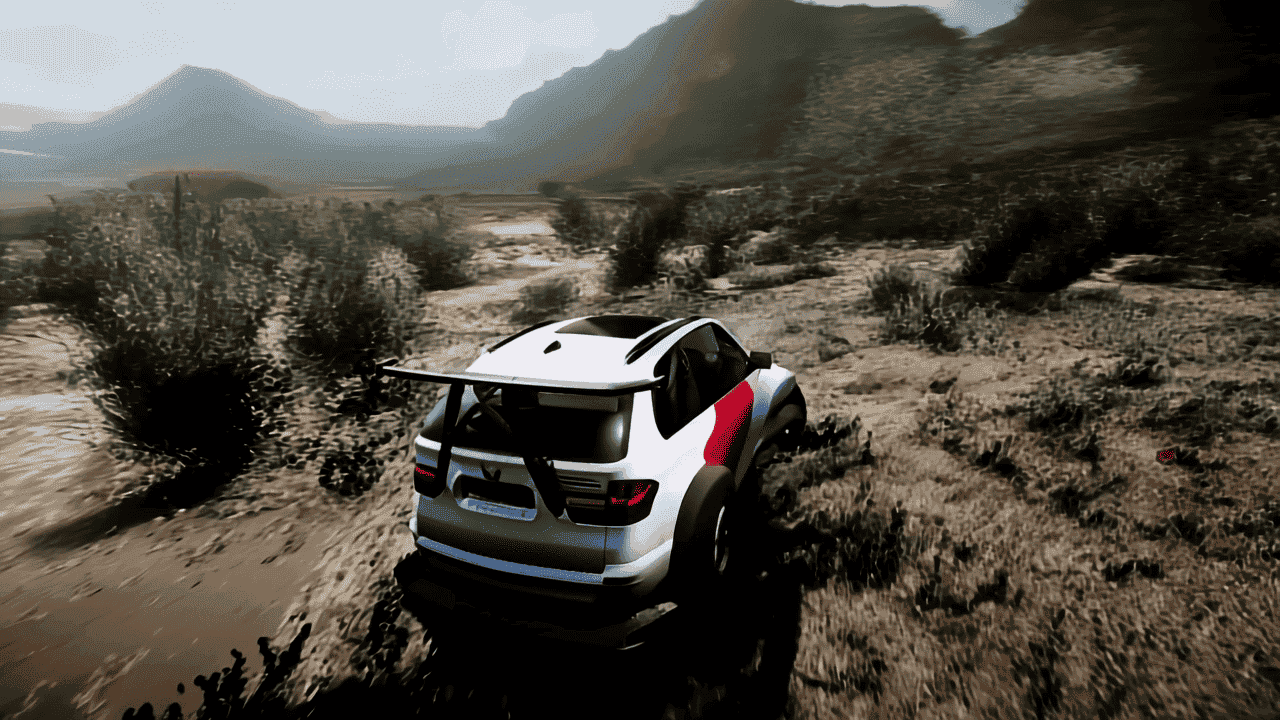}
     \includegraphics[width=0.8\linewidth]{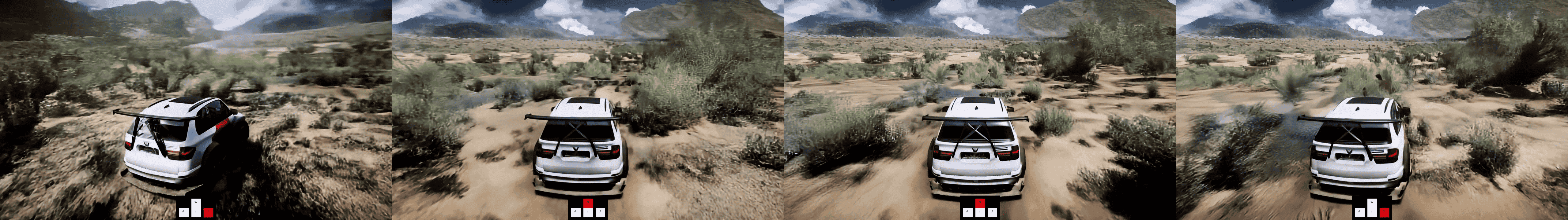}\vspace{-3.7pt}
     \\
     \includegraphics[width=0.2\linewidth]{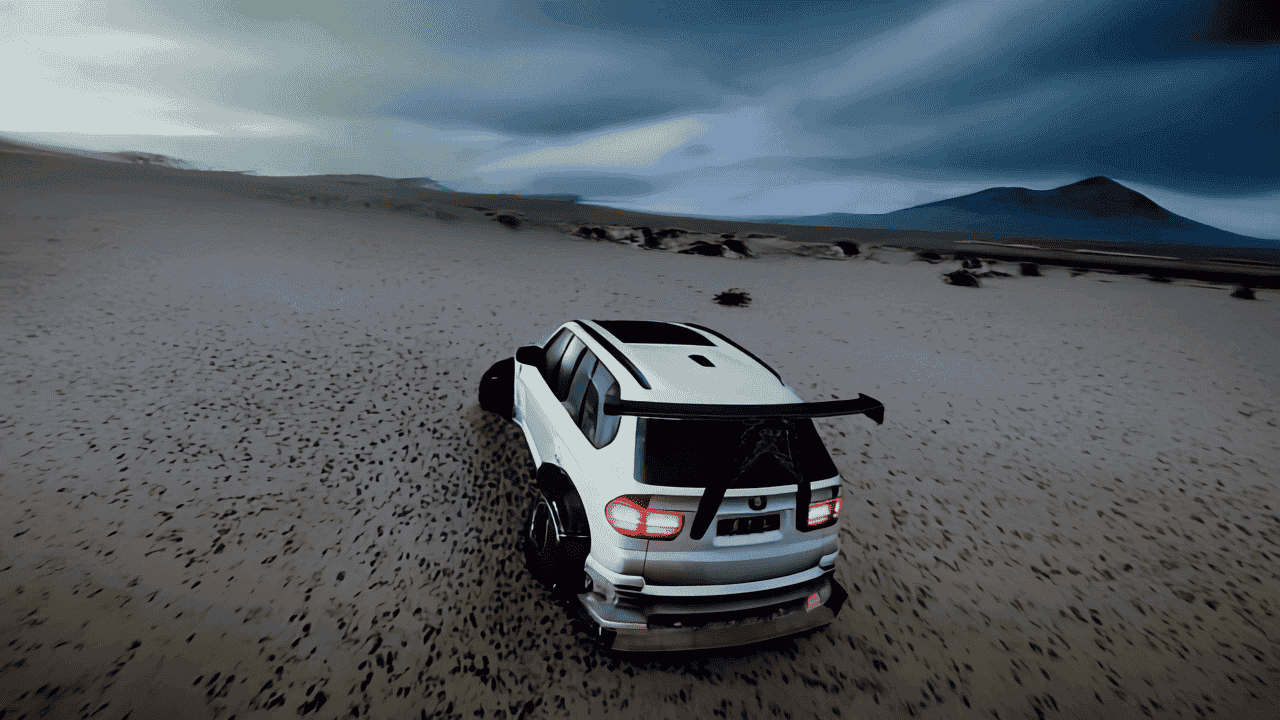}
     \includegraphics[width=0.8\linewidth]{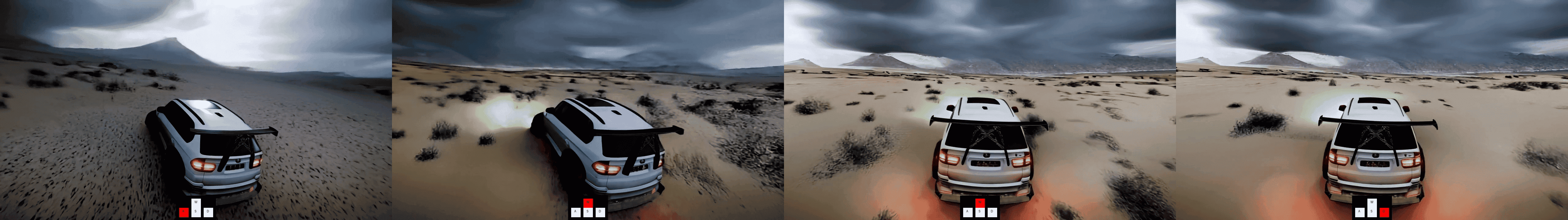}\vspace{-3.7pt}\\
     \includegraphics[width=0.2\linewidth]{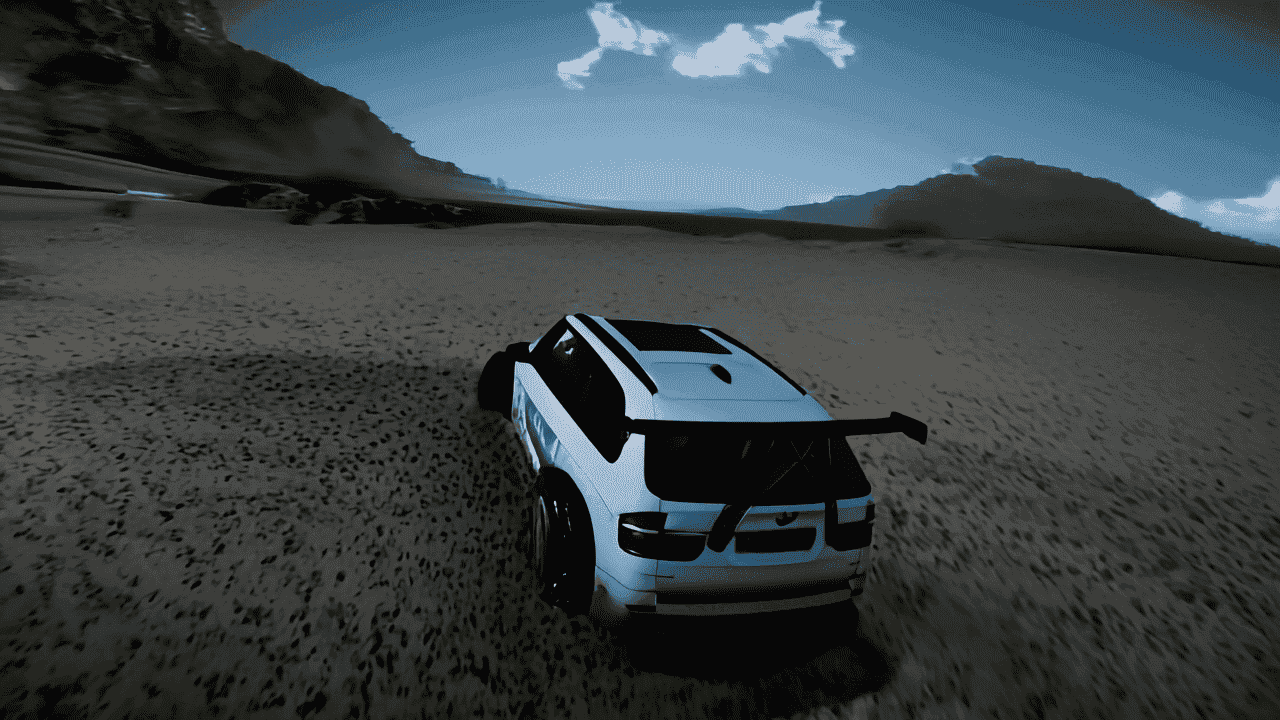} 
     \includegraphics[width=0.8\linewidth]{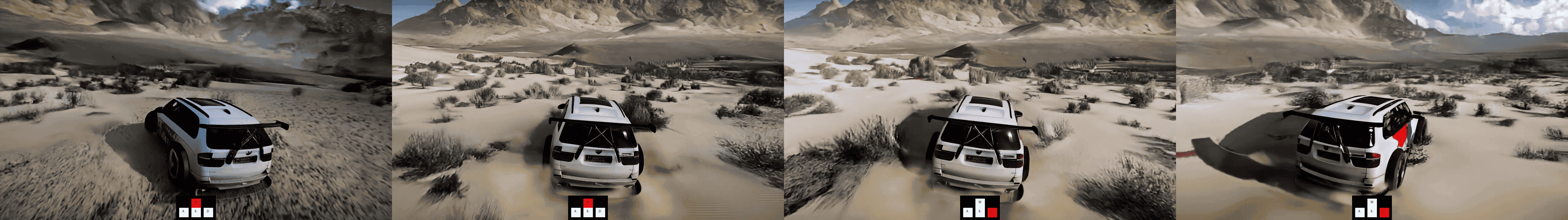}\vspace{-3.7pt}
     \\
     \includegraphics[width=0.2\linewidth]{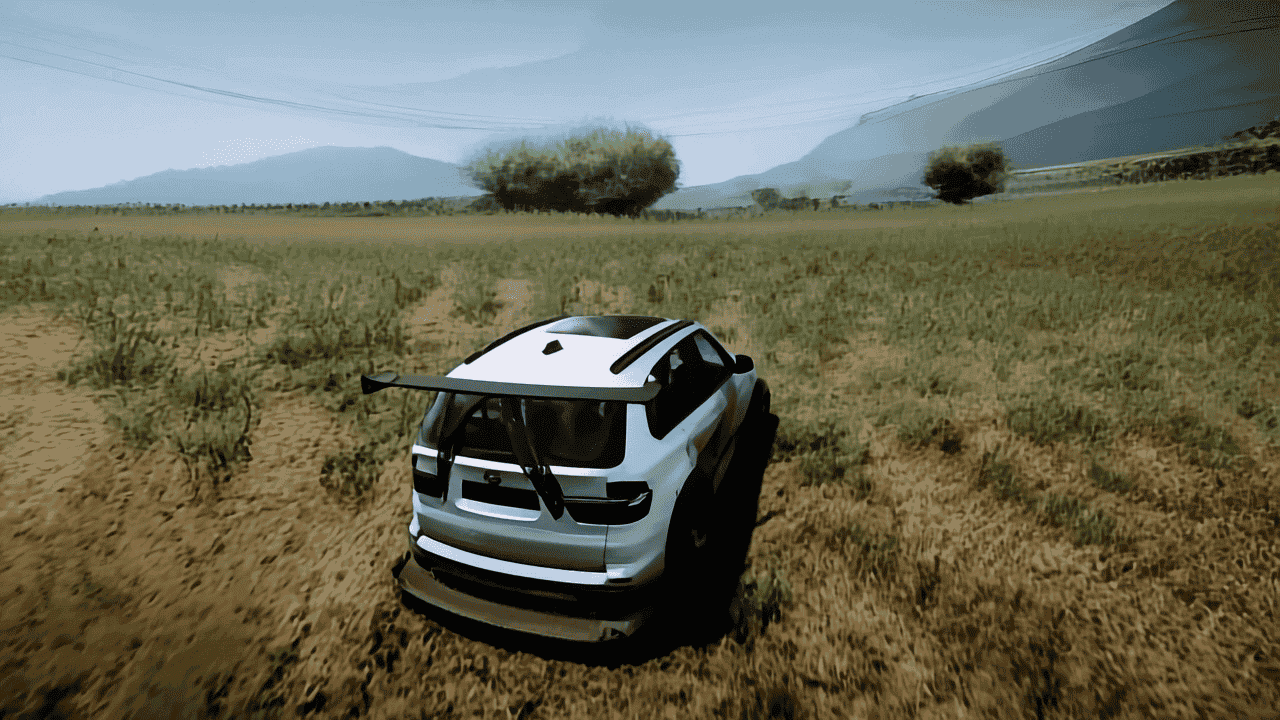} 
     \includegraphics[width=0.8\linewidth]{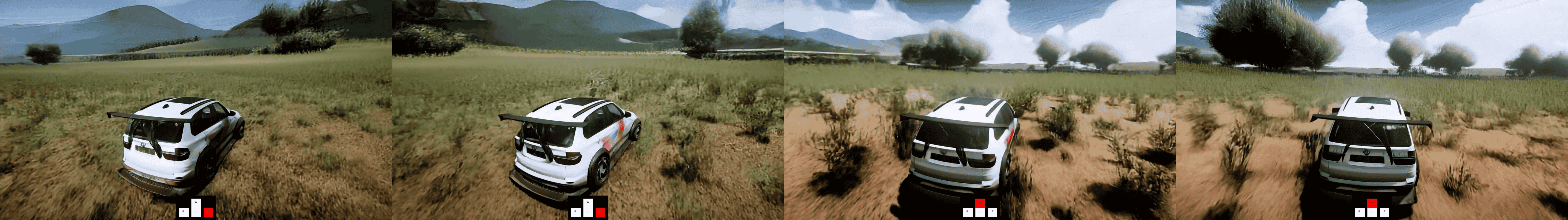}\vspace{-3.7pt}
     \\
     \includegraphics[width=0.2\linewidth]{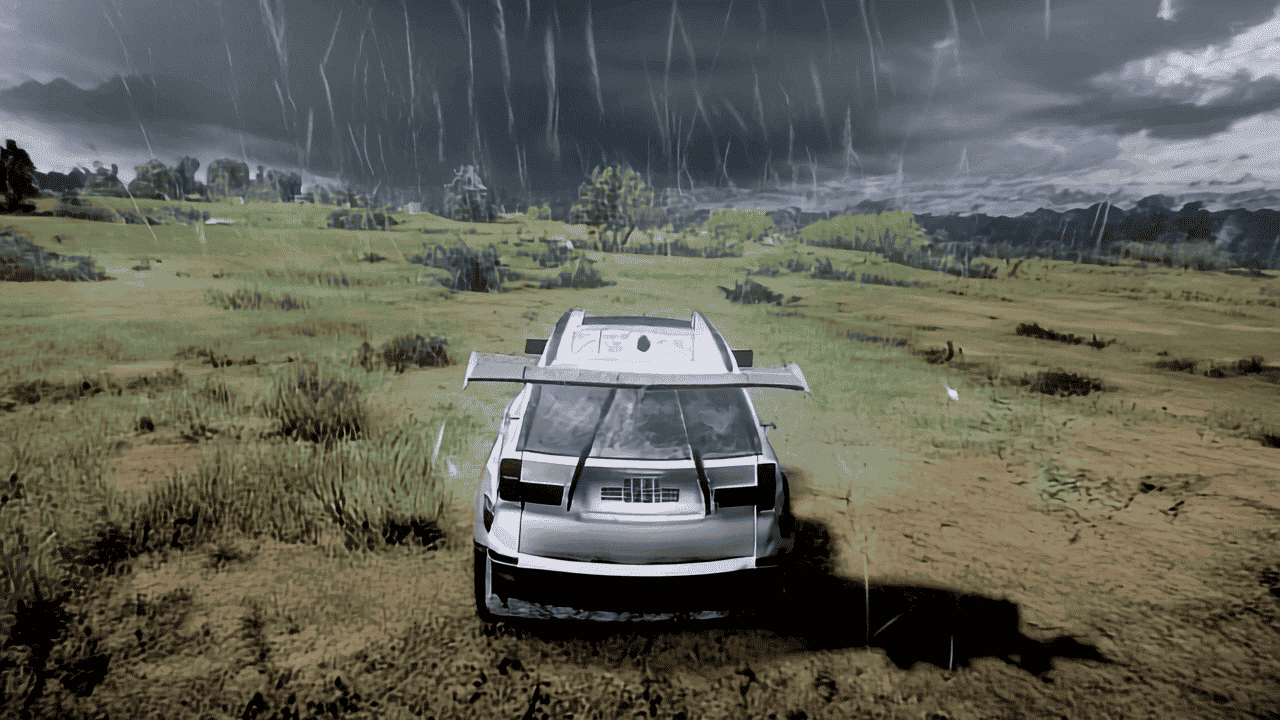} 
     \includegraphics[width=0.8\linewidth]{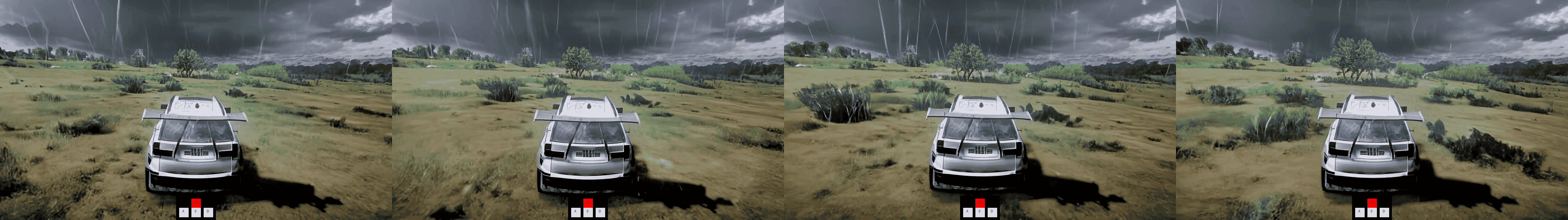}\vspace{-3.7pt}
     \\
     \includegraphics[width=0.2\linewidth]{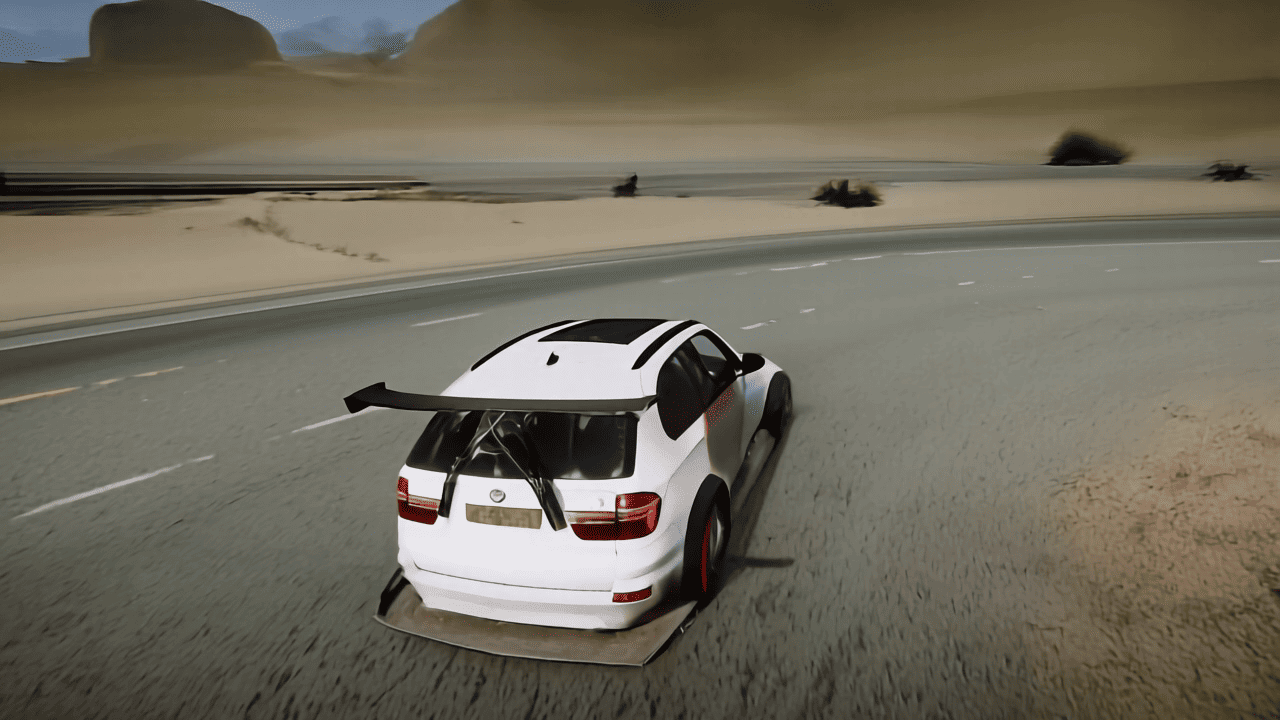} 
     \includegraphics[width=0.8\linewidth]{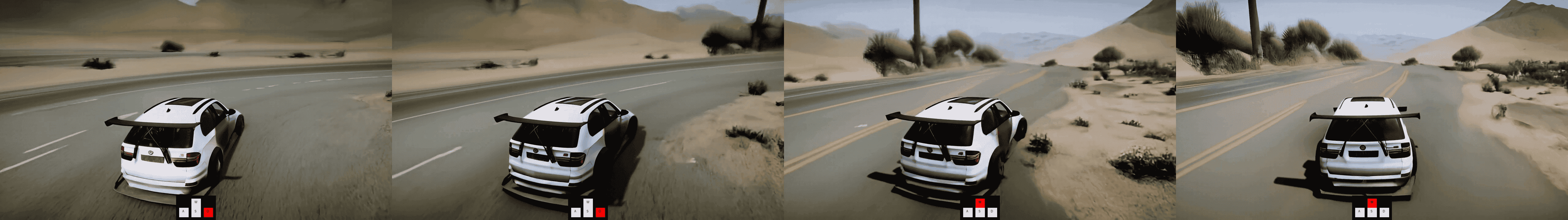}\vspace{-3.7pt}
     \\
     \includegraphics[width=0.2\linewidth]{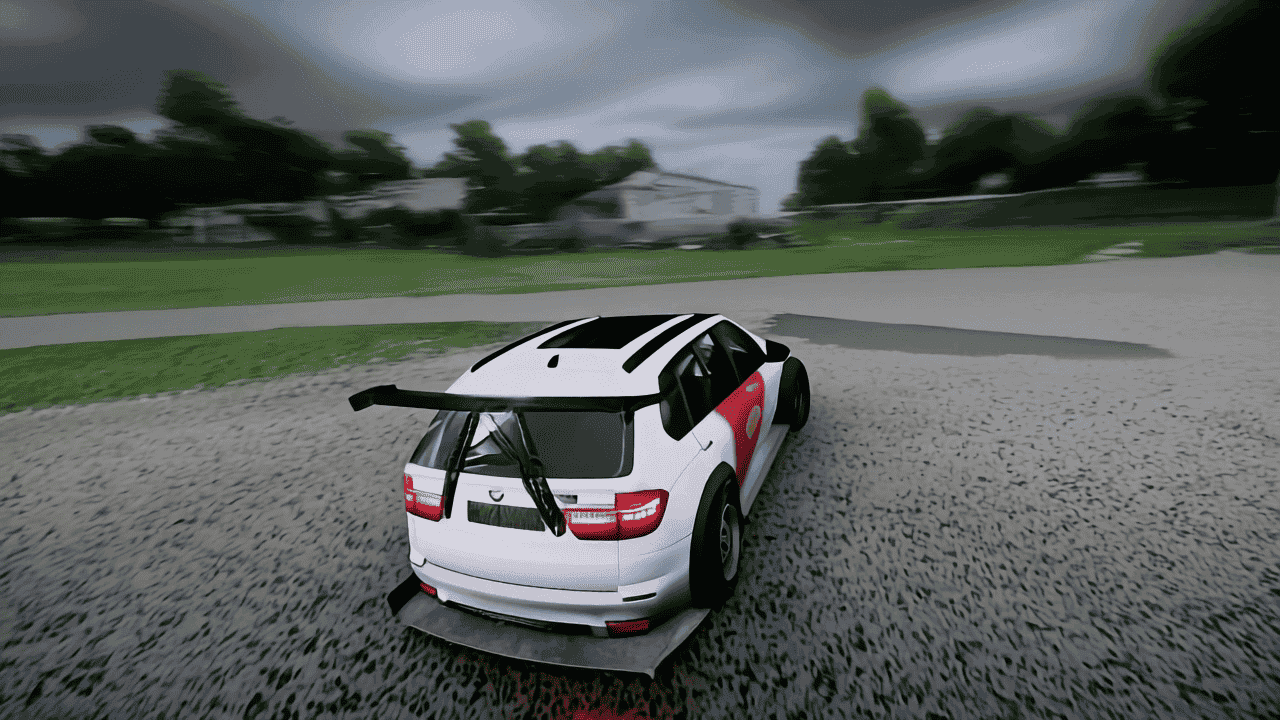} 
     \includegraphics[width=0.8\linewidth]{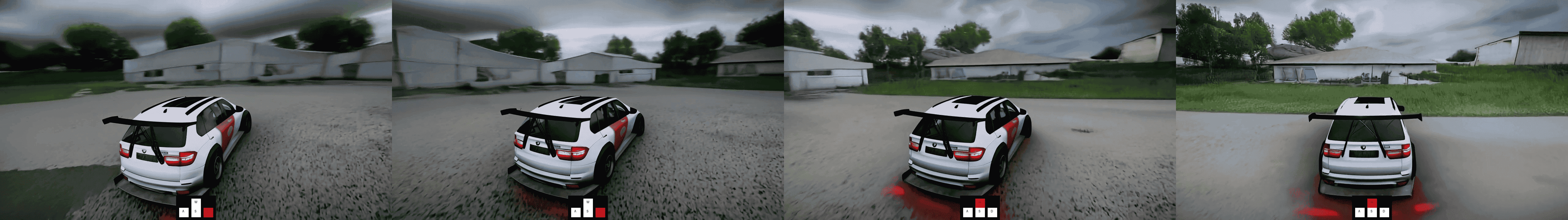}\vspace{-3.7pt}
     \\
     \includegraphics[width=0.2\linewidth]{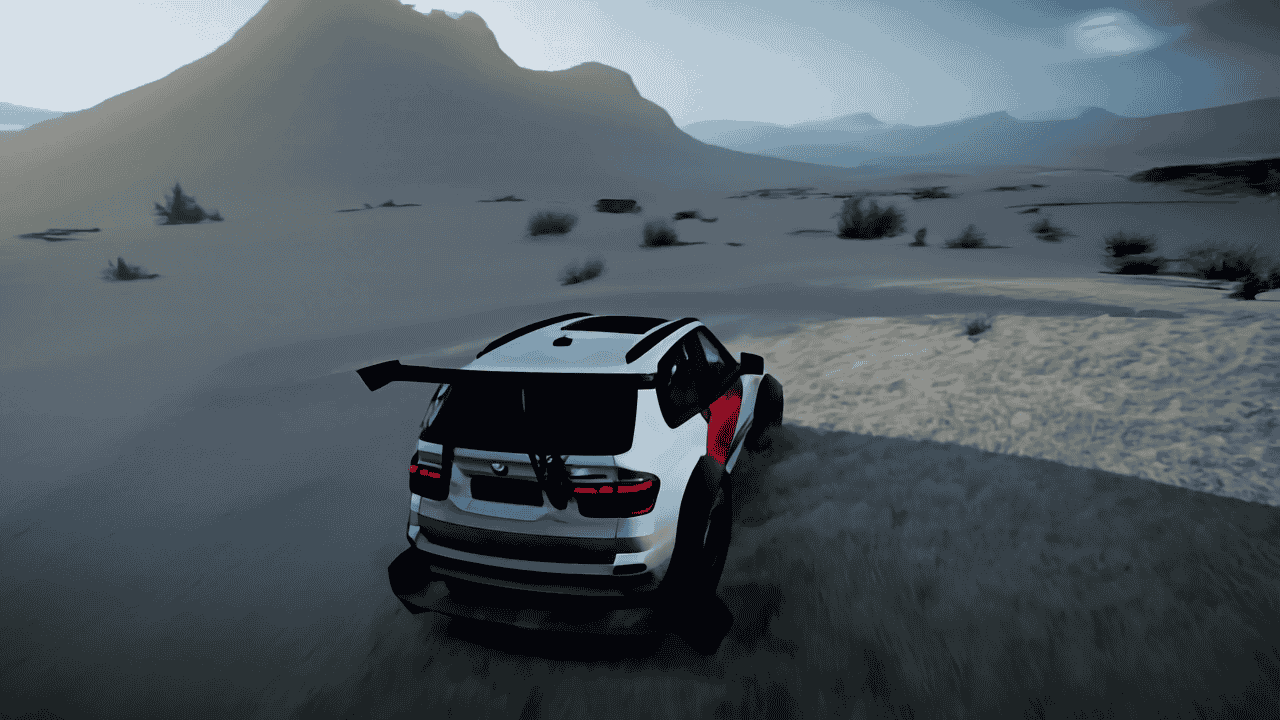} 
     \includegraphics[width=0.8\linewidth]{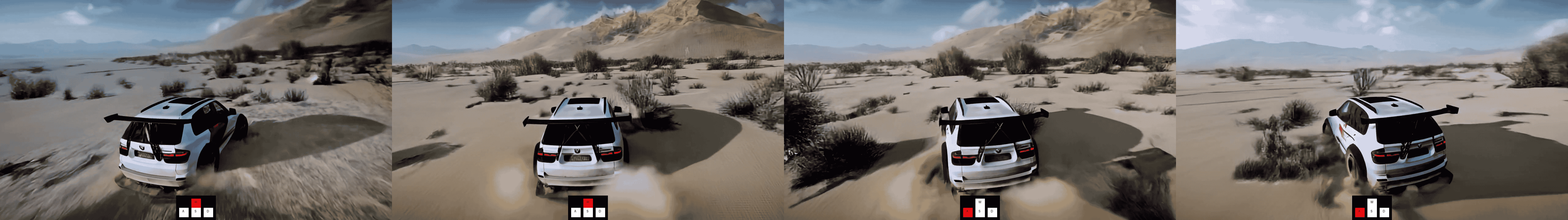}\vspace{-3.7pt}
     \\
     \includegraphics[width=0.2\linewidth]{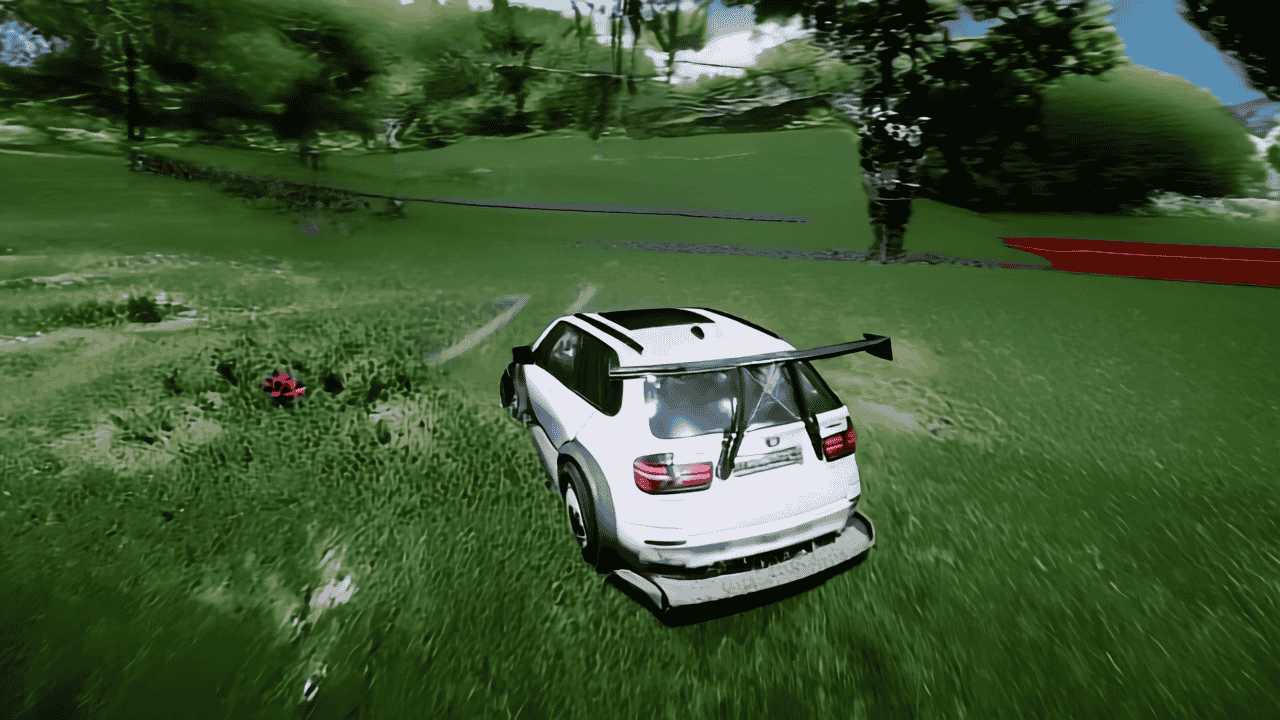}    
     \includegraphics[width=0.8\linewidth]{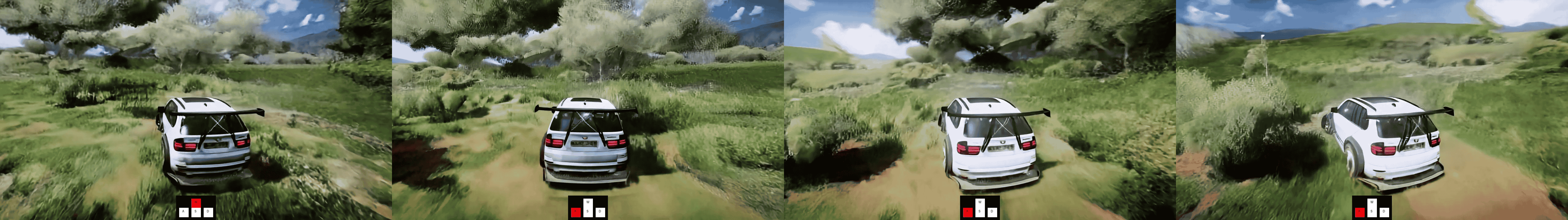}\vspace{-3.7pt}
     \\
     \includegraphics[width=0.2\linewidth]{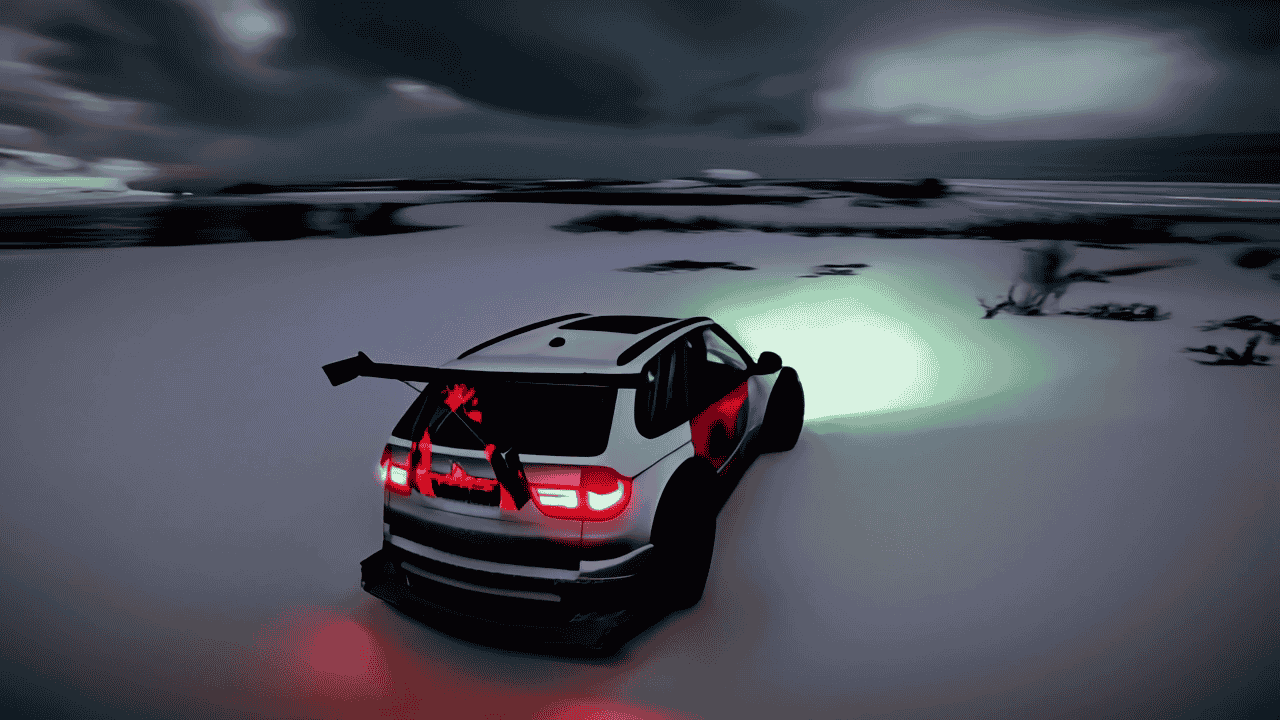}    
     \includegraphics[width=0.8\linewidth]{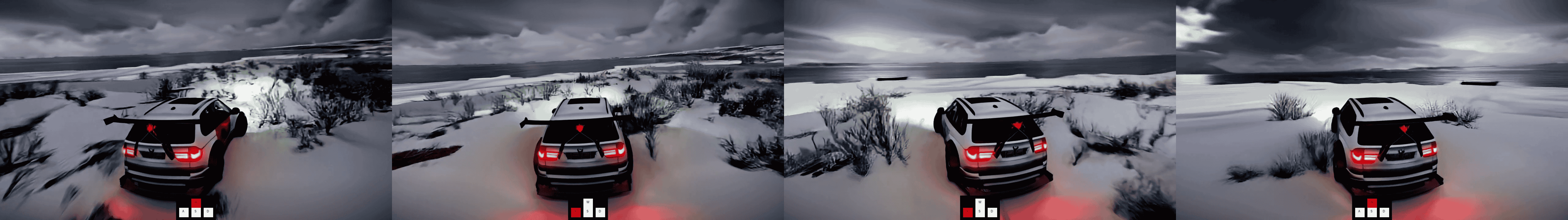}\vspace{-3.7pt}
     \\
    \includegraphics[width=0.2\linewidth]{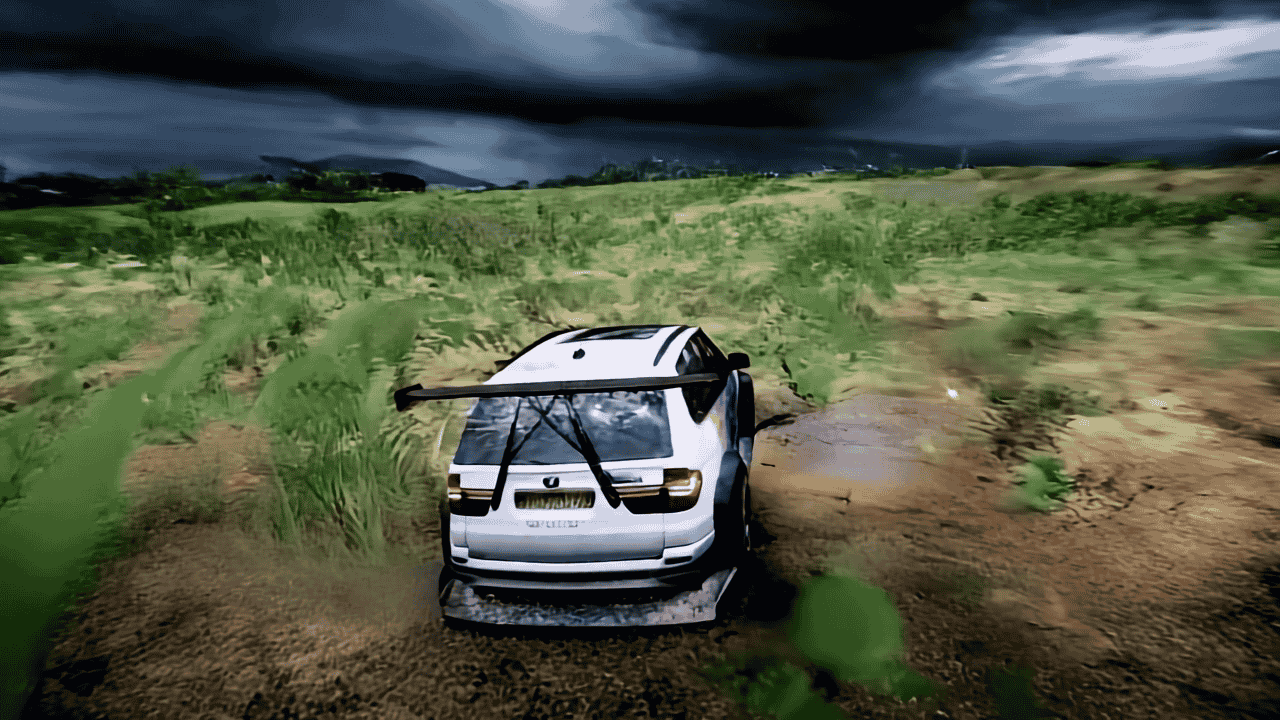}    
     \includegraphics[width=0.8\linewidth]{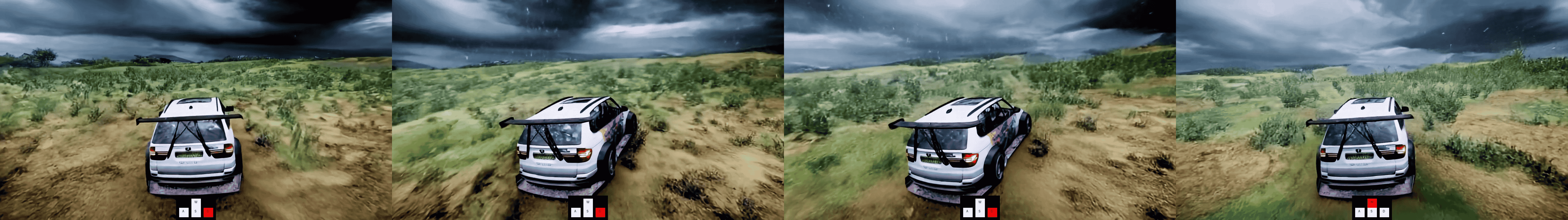}\vspace{-3.7pt}
    \end{tabular}
    \caption{More results demonstrate frame-level precise control achieved by the \controller across diverse scenes, weather conditions, and movement modes.}\label{fig:supp_4s_demo}
\end{figure*}

\begin{figure*}
    \centering
    \begin{tabular}{c}
     \includegraphics[width=0.2\linewidth]{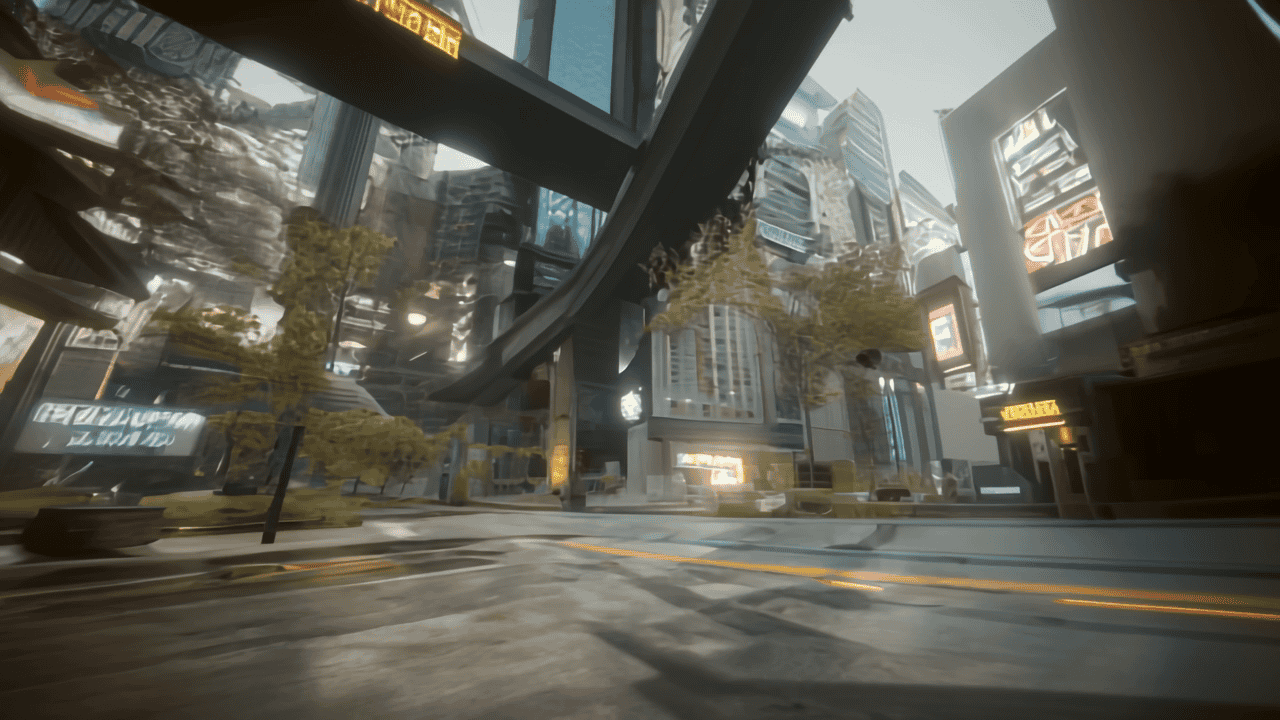} 
     \includegraphics[width=0.8\linewidth]{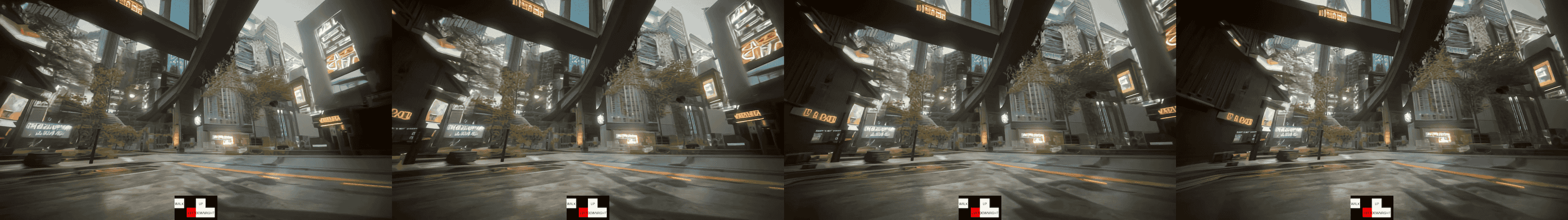}\vspace{-3.7pt}
     \\
     \includegraphics[width=0.2\linewidth]{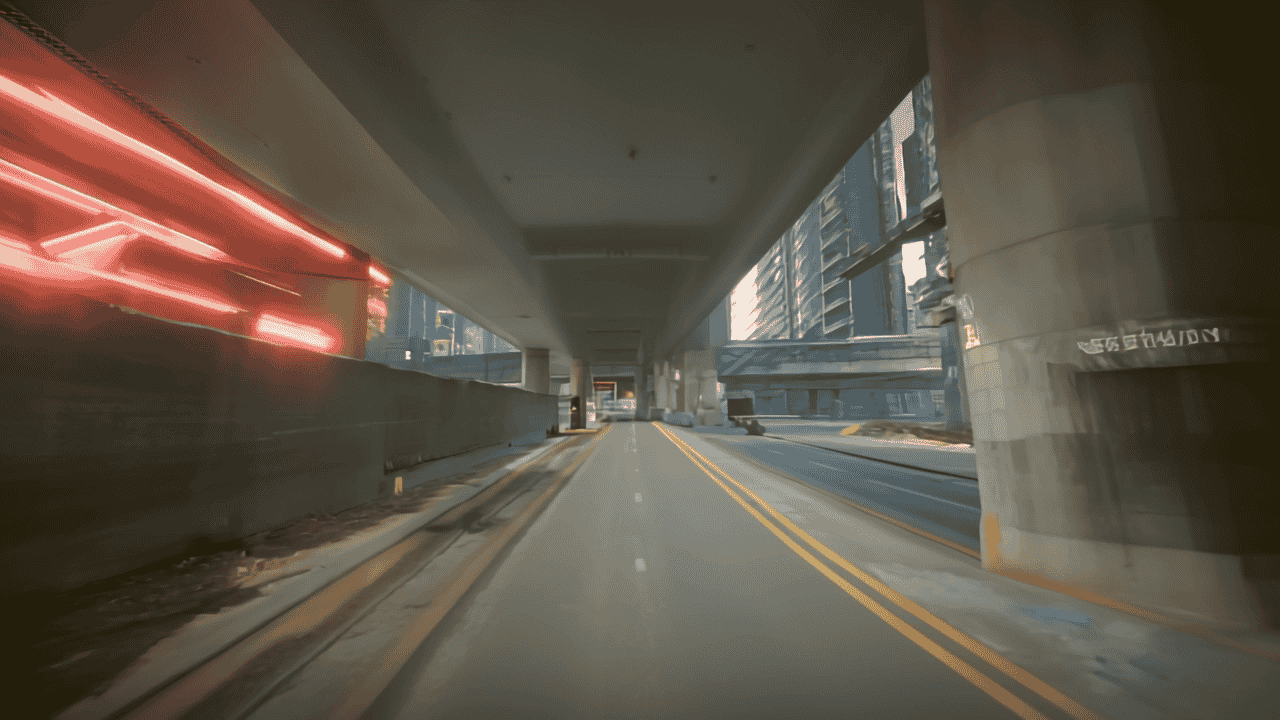} 
     \includegraphics[width=0.8\linewidth]{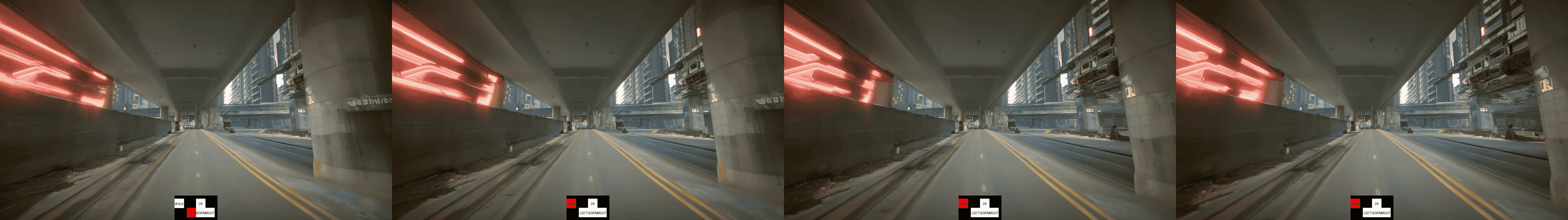}\vspace{-3.7pt}\\
     \includegraphics[width=0.2\linewidth]{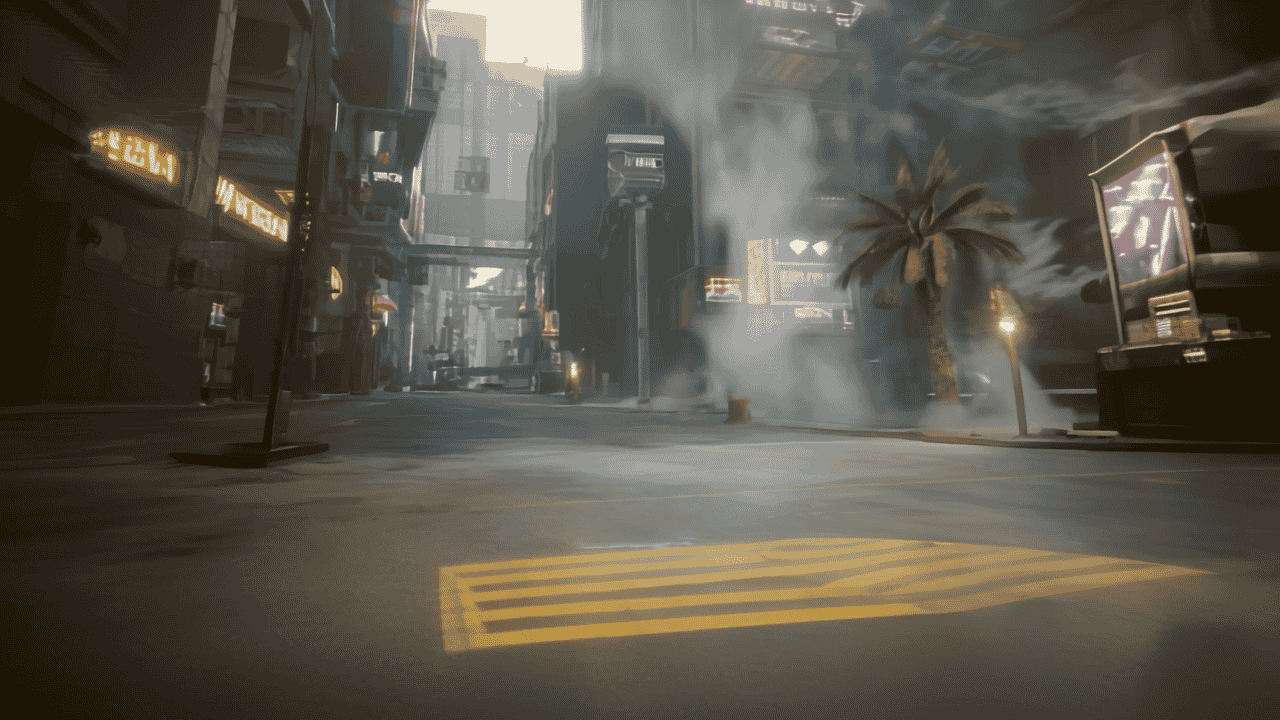} 
     \includegraphics[width=0.8\linewidth]{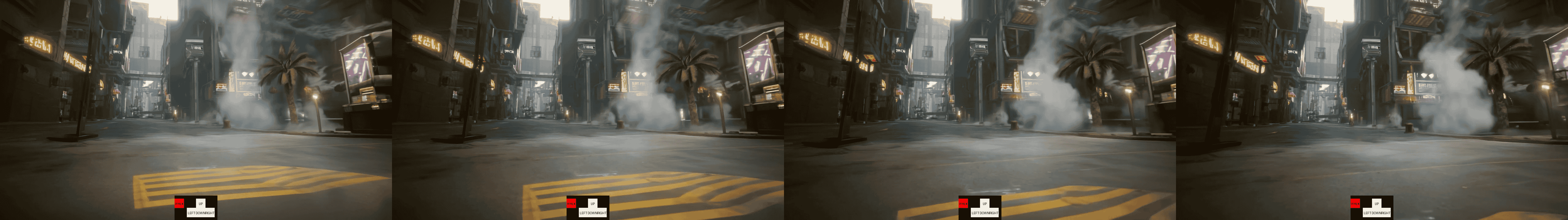}\vspace{-3.7pt}\\
     \includegraphics[width=0.2\linewidth]{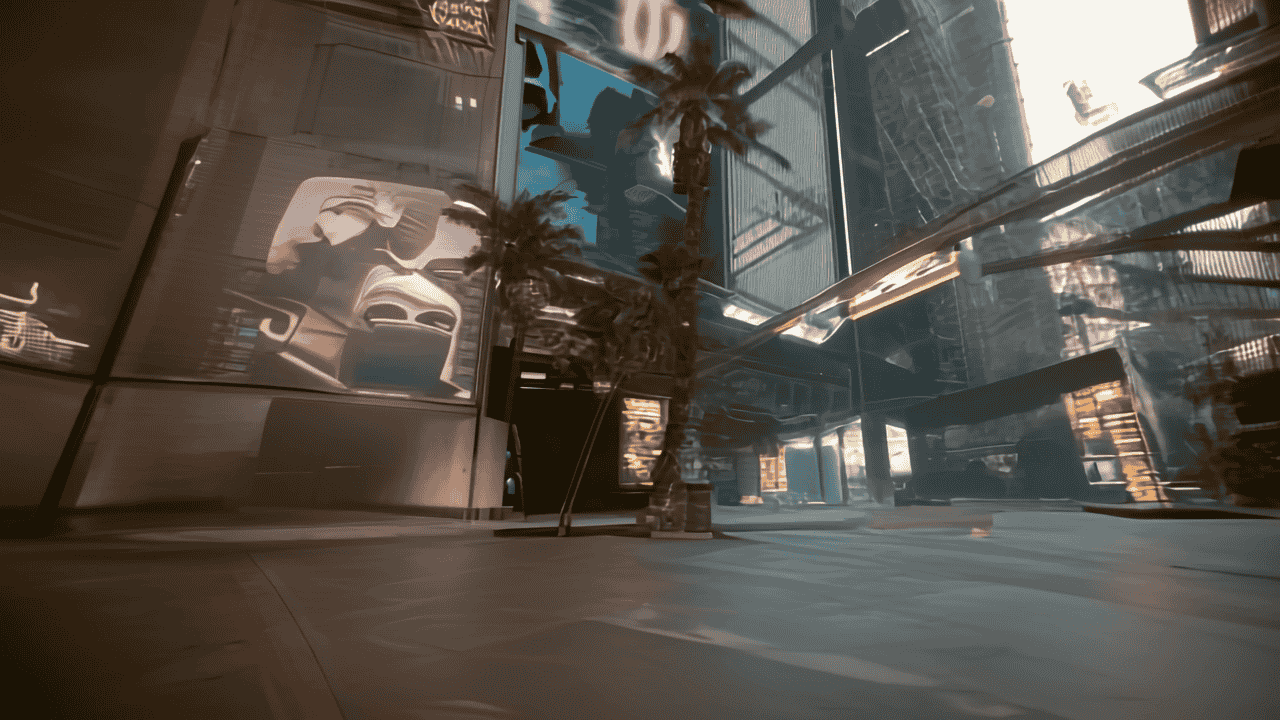} 
     \includegraphics[width=0.8\linewidth]{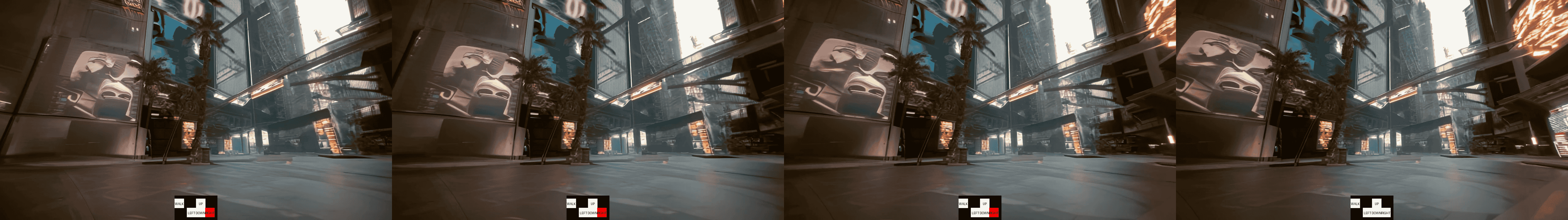}\vspace{-3.7pt}\\
     \includegraphics[width=0.2\linewidth]{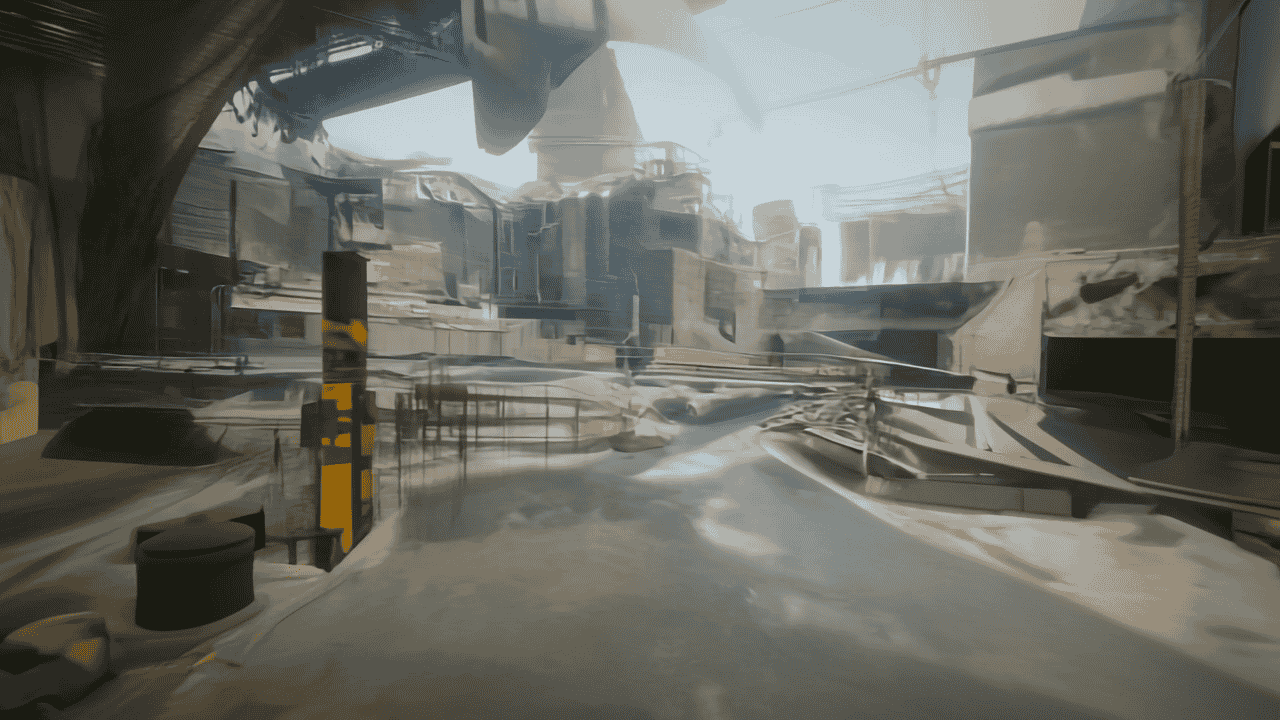} 
     \includegraphics[width=0.8\linewidth]{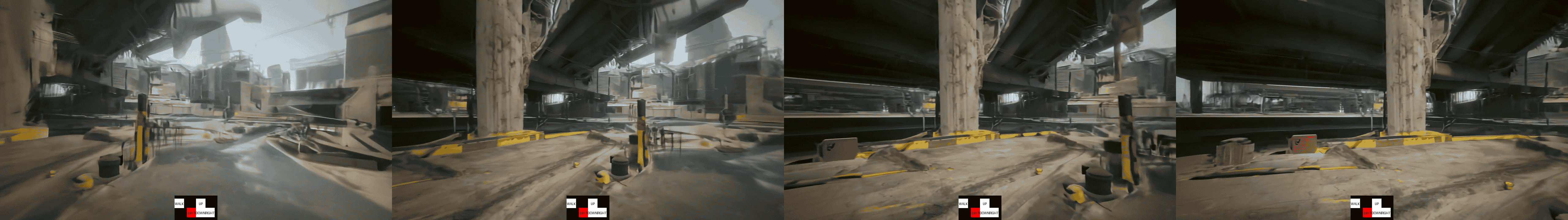}\vspace{-3.7pt}\\
     \includegraphics[width=0.2\linewidth]{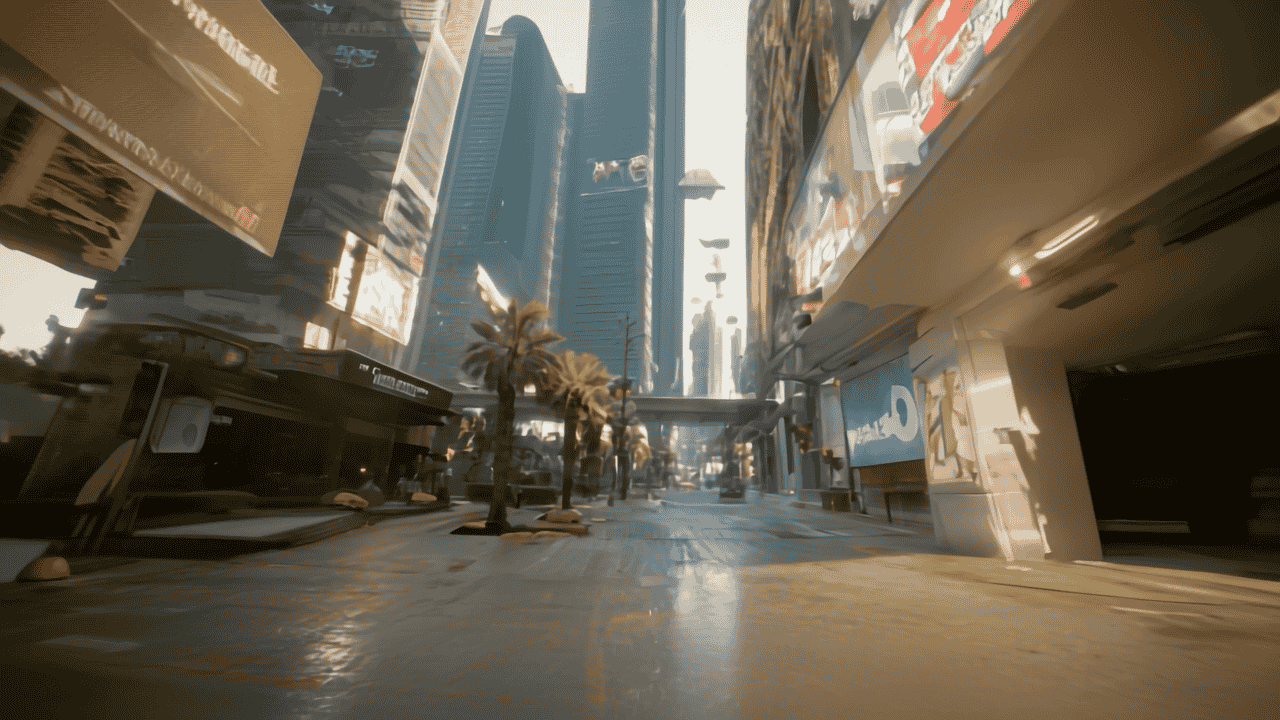} 
     \includegraphics[width=0.8\linewidth]{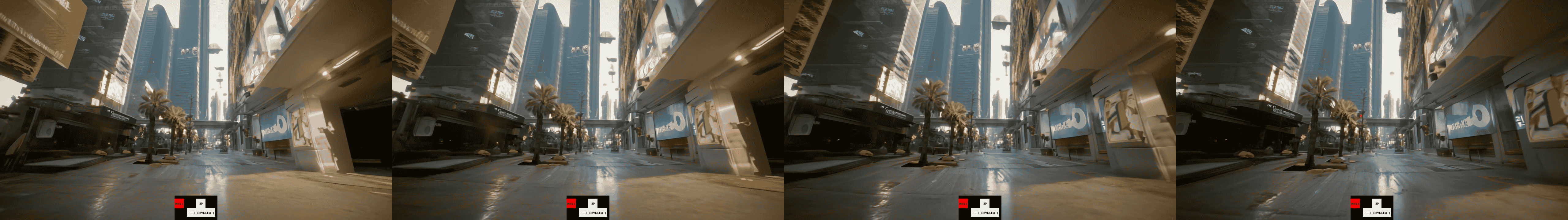}\vspace{-3.7pt}\\
     \includegraphics[width=0.2\linewidth]{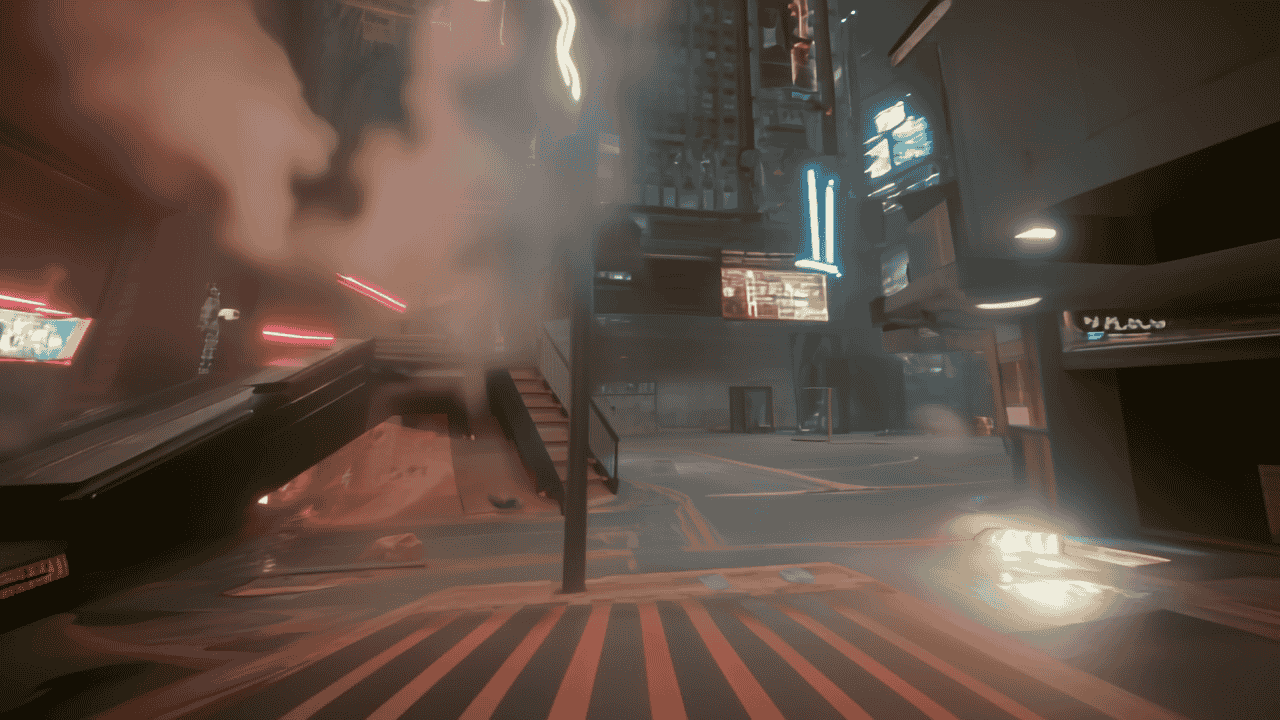} 
     \includegraphics[width=0.8\linewidth]{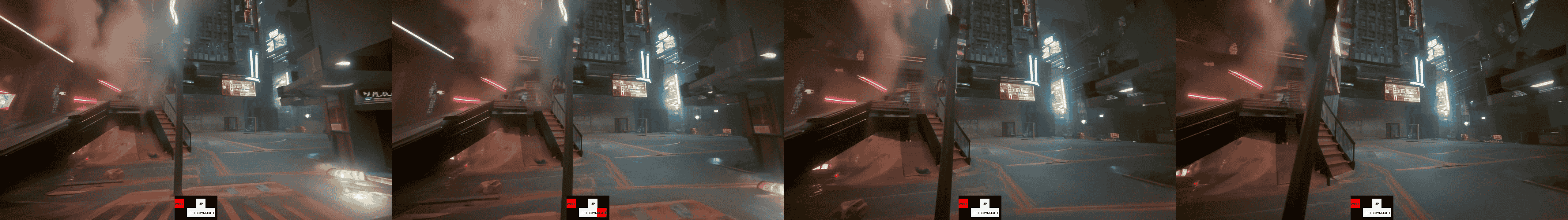}\vspace{-3.7pt}\\
     \includegraphics[width=0.2\linewidth]{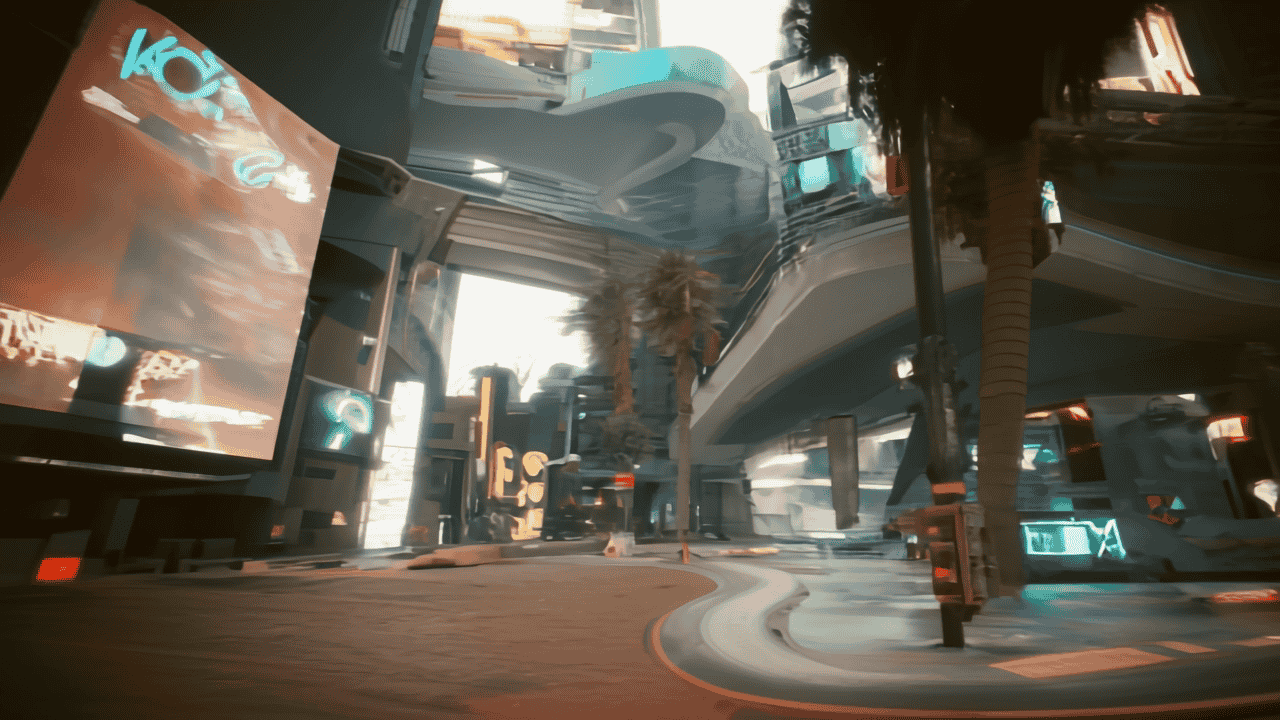} 
     \includegraphics[width=0.8\linewidth]{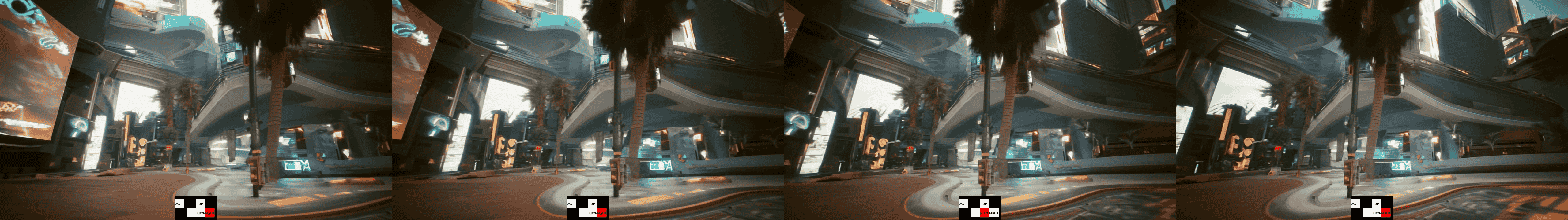}\vspace{-3.7pt}\\
     \includegraphics[width=0.2\linewidth]{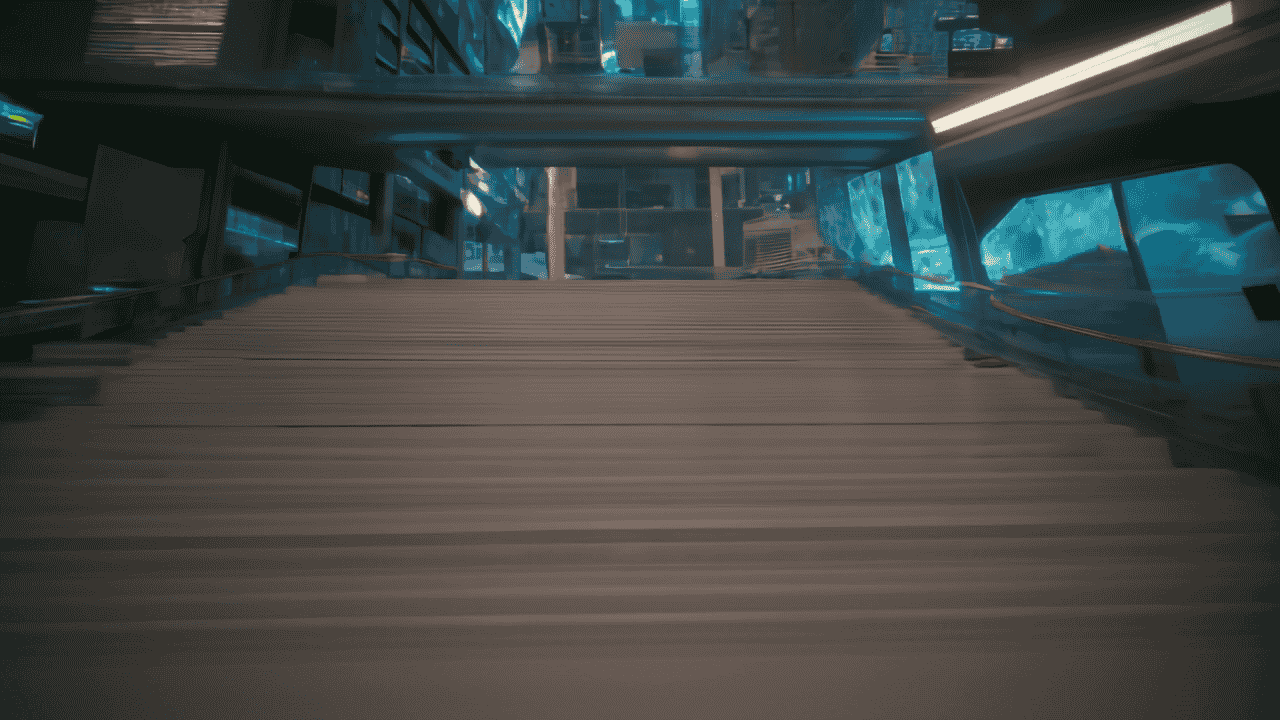} 
     \includegraphics[width=0.8\linewidth]{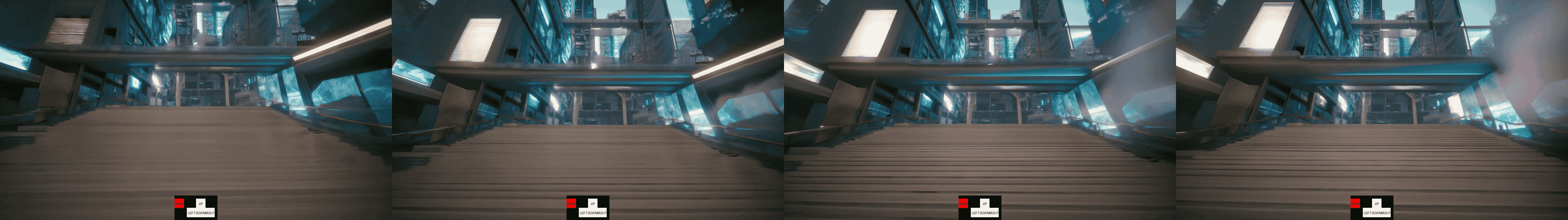}\vspace{-3.7pt}\\
     \includegraphics[width=0.2\linewidth]{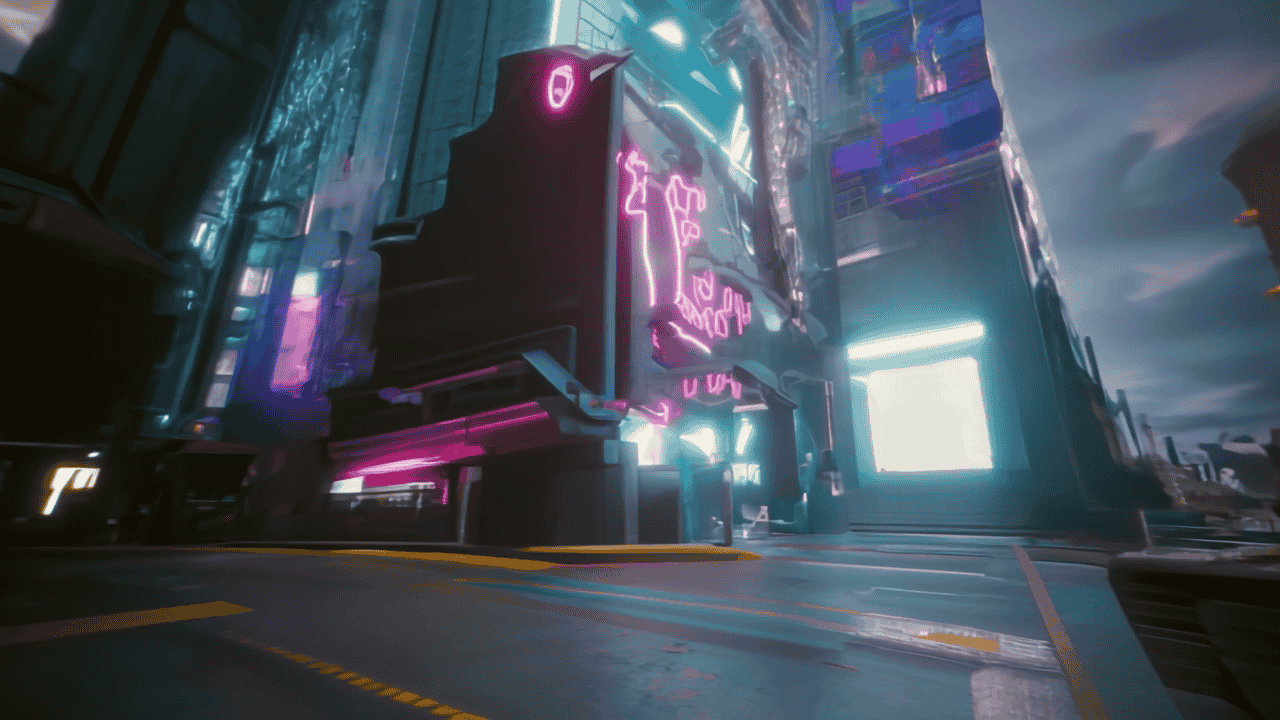}  
     \includegraphics[width=0.8\linewidth]{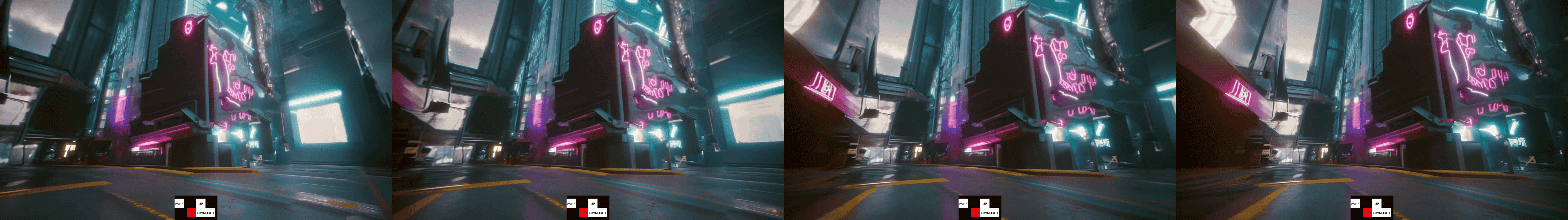}\vspace{-3.7pt}\\
     \includegraphics[width=0.2\linewidth]{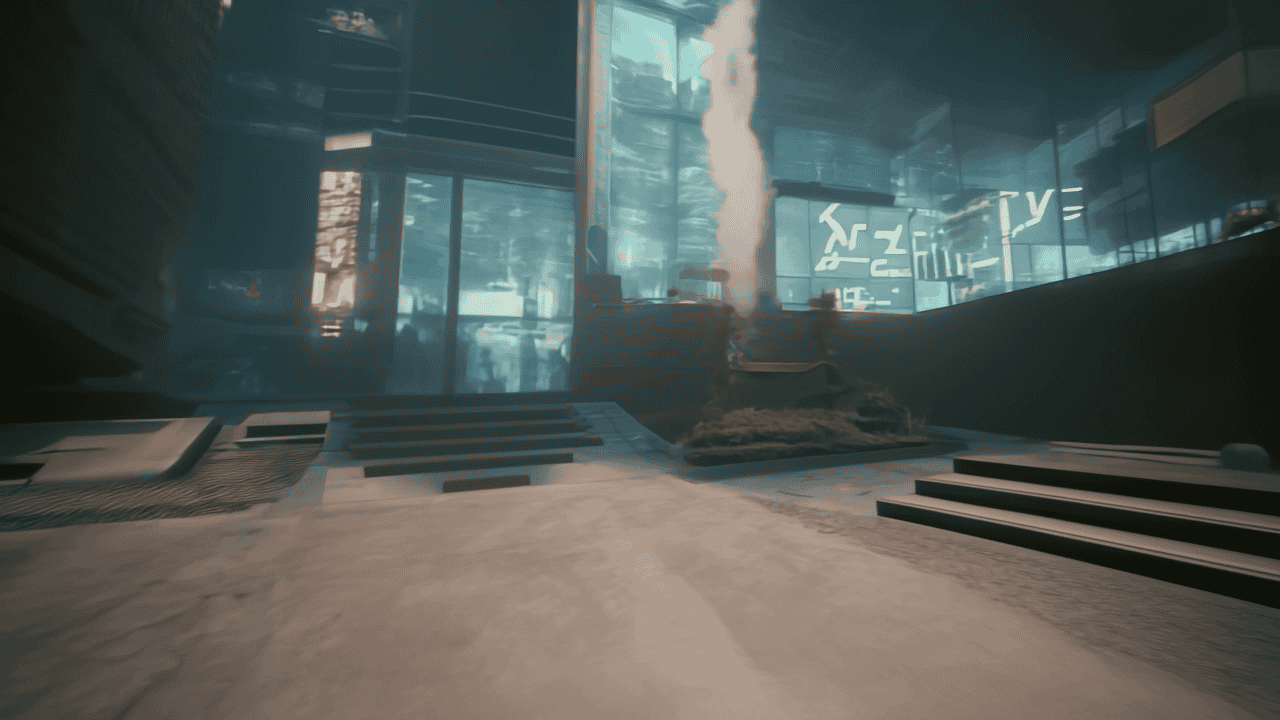}  
     \includegraphics[width=0.8\linewidth]{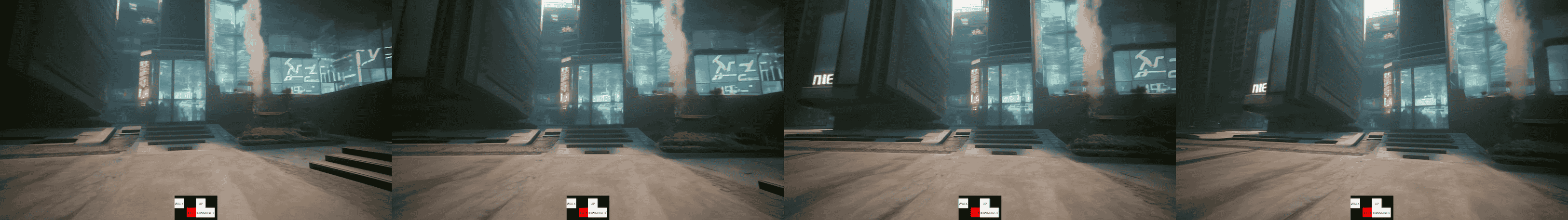}\vspace{-3.7pt}
    \end{tabular}
    \caption{More results demonstrate frame-level precise control achieved by the \controller across diverse scenes, weather conditions, and movement modes.}\label{fig:supp_cyber_demos}
\end{figure*}

We provide more results on precise frame-level moving control in \cref{fig:supp_4s_demo} and \cref{fig:supp_cyber_demos}.

\subsection{Generalization}
\begin{figure*}[htbp]
     \centering
       \begin{tabular}{c}
     \includegraphics[width=0.2\linewidth]{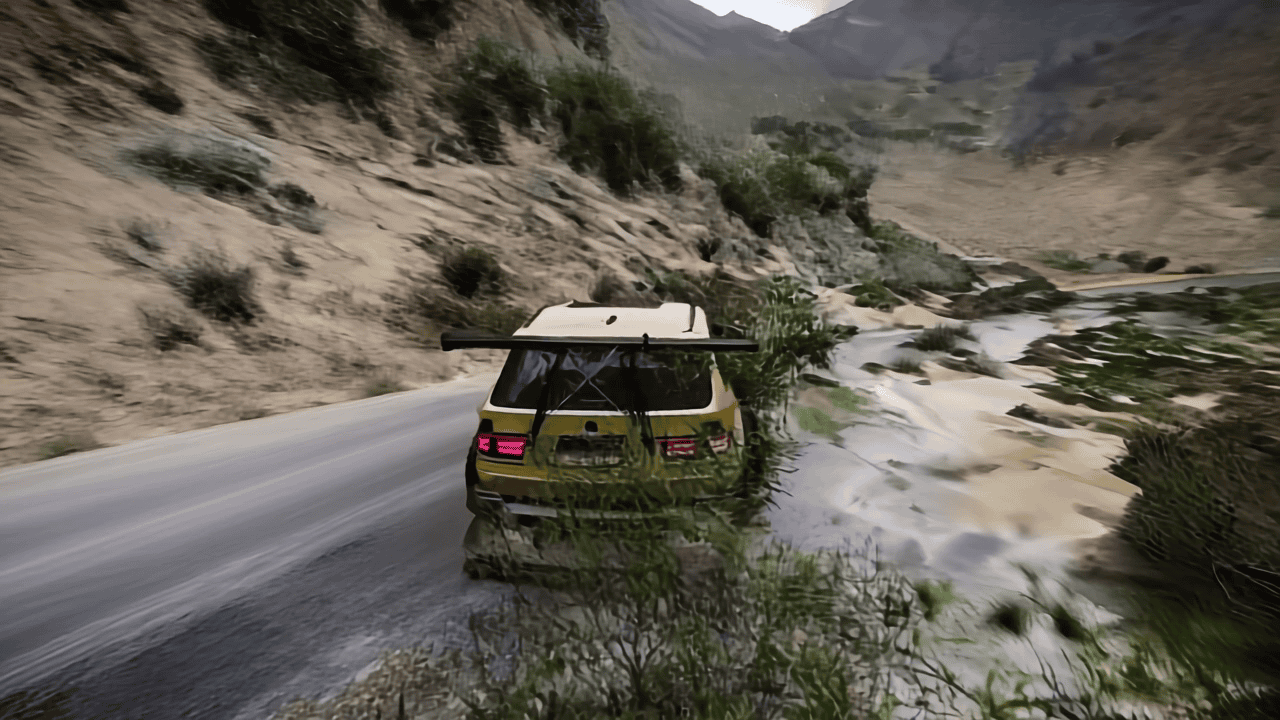} 
     \includegraphics[width=0.8\linewidth]{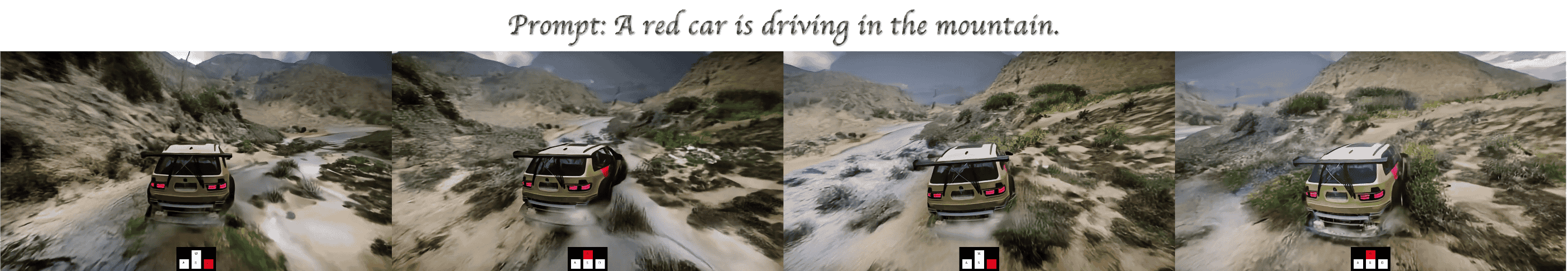}
     \\
     \includegraphics[width=0.2\linewidth]{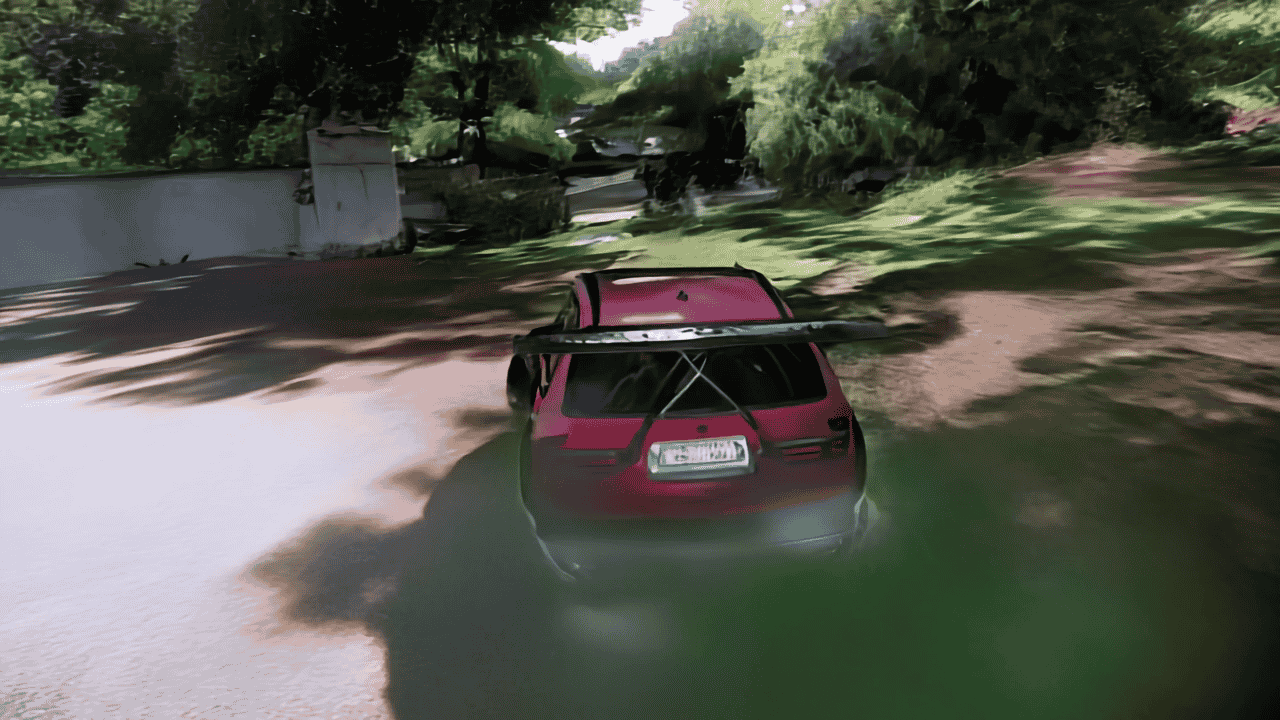} 
     \includegraphics[width=0.8\linewidth]{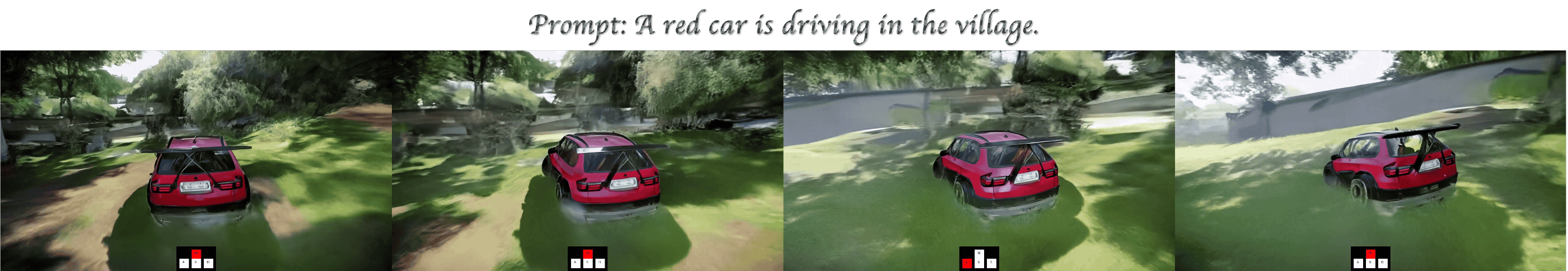}
     \\
     \includegraphics[width=0.2\linewidth]{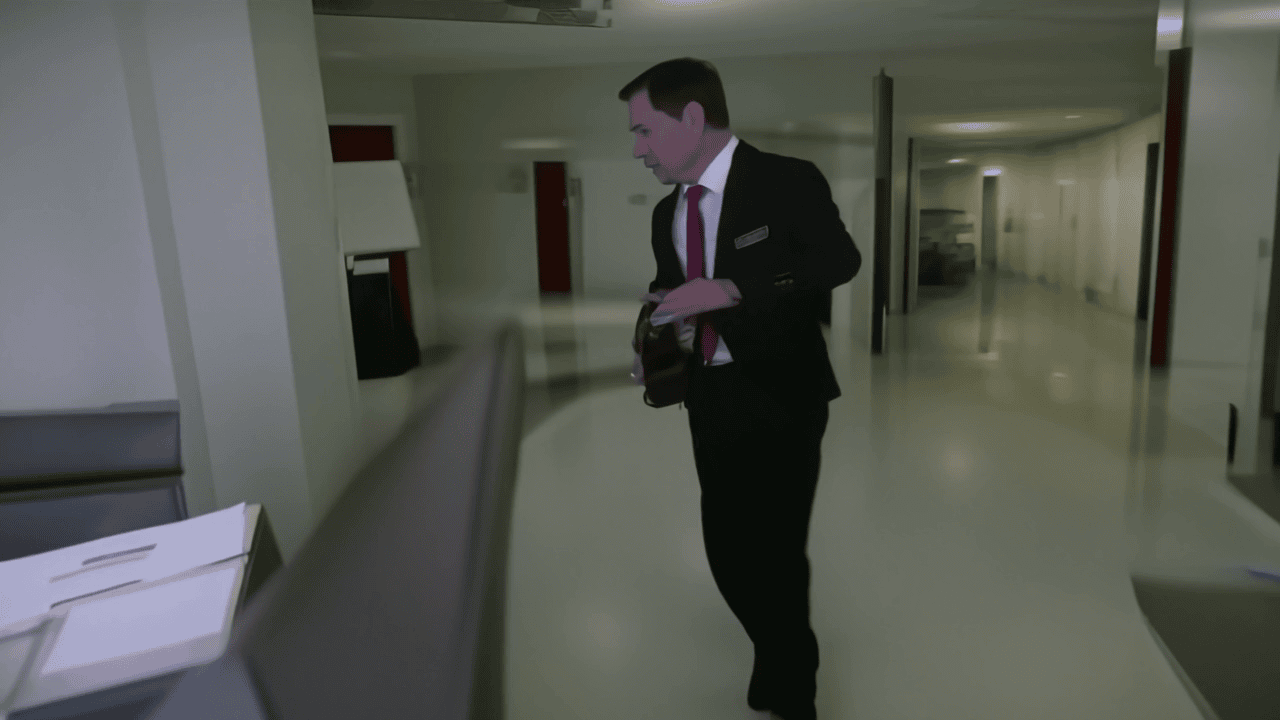} 
     \includegraphics[width=0.8\linewidth]{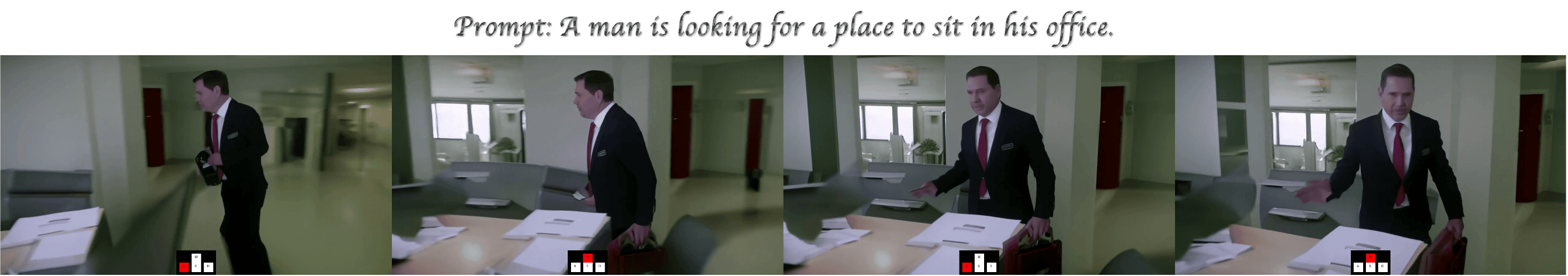}
     \\
     \includegraphics[width=0.2\linewidth]{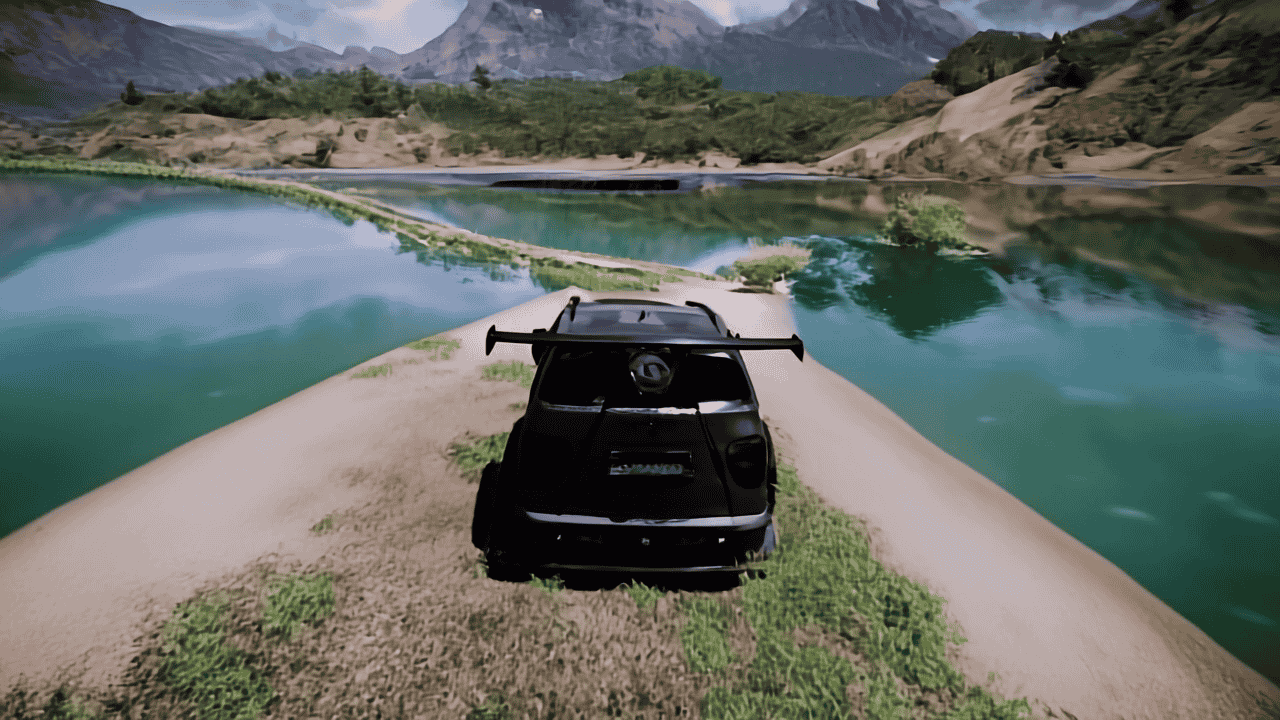} 
     \includegraphics[width=0.8\linewidth]{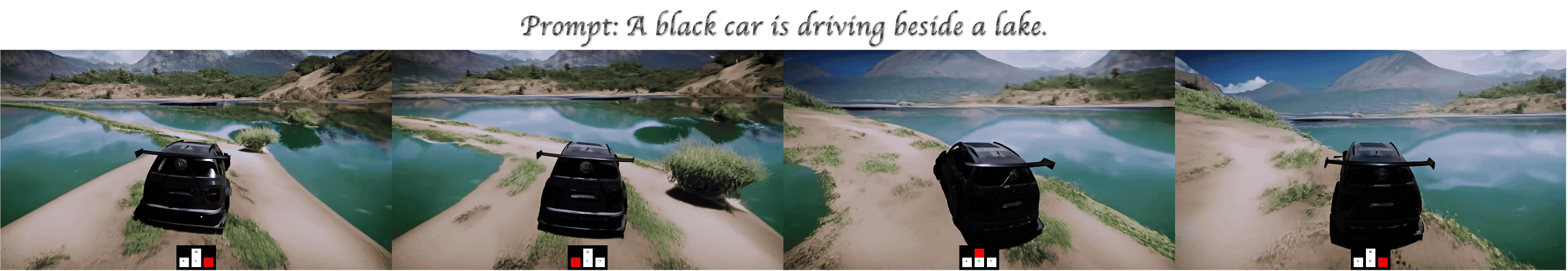}
     \\
     \includegraphics[width=0.2\linewidth]{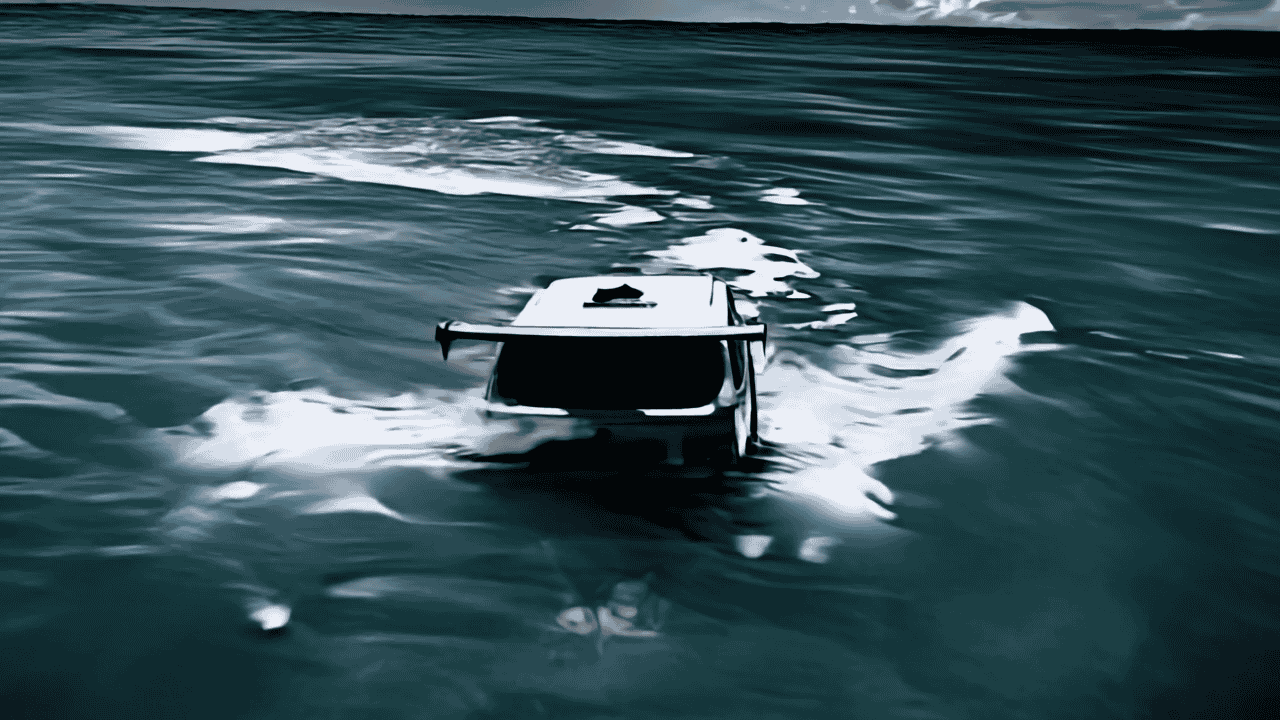} 
     \includegraphics[width=0.8\linewidth]{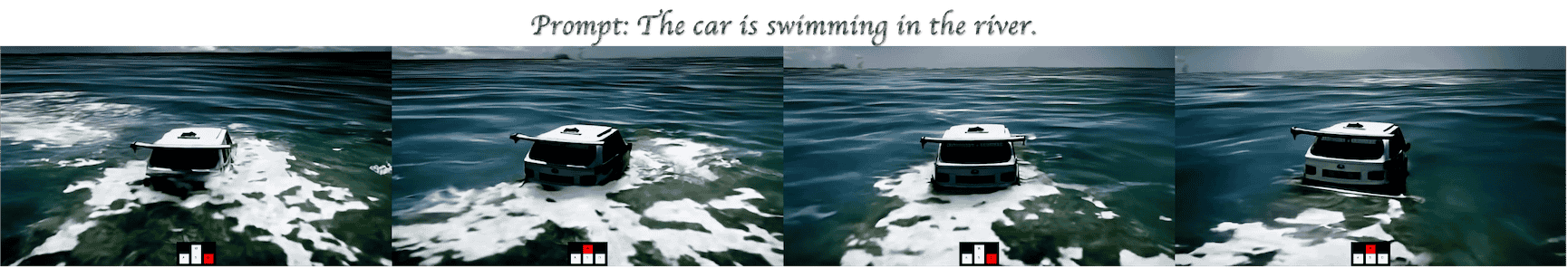}
     
     \\
     \includegraphics[width=0.2\linewidth]{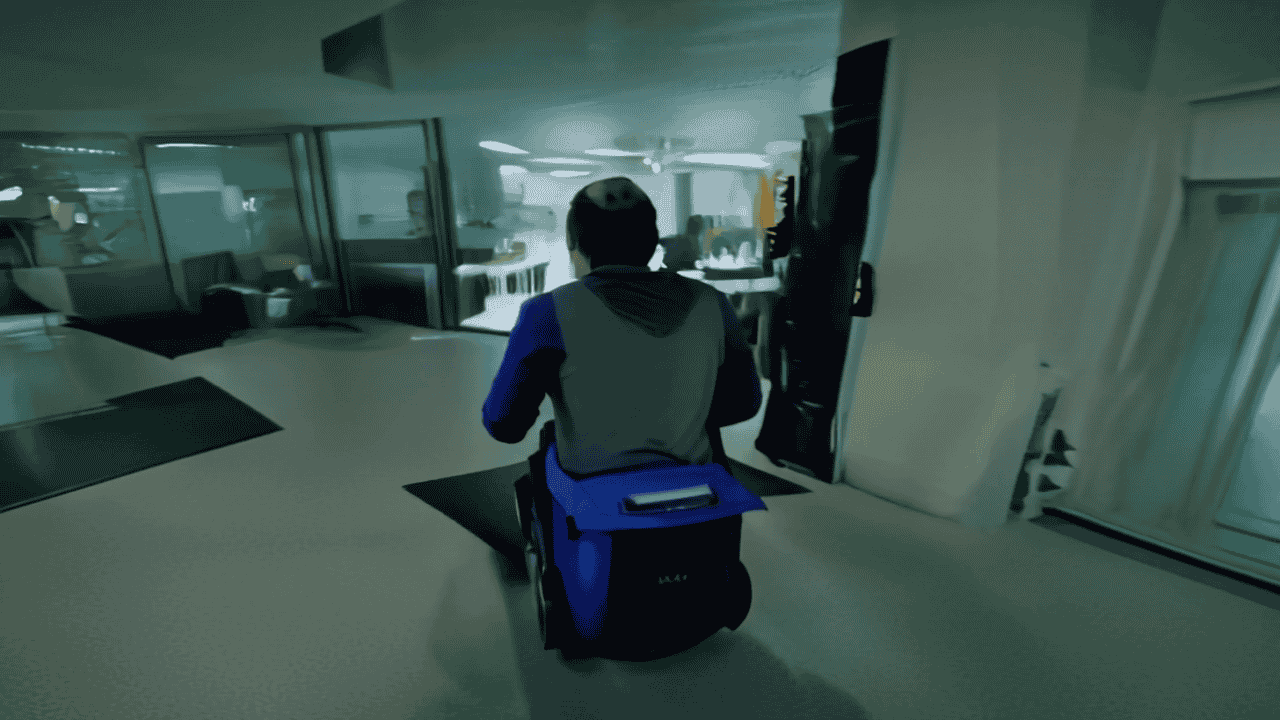} 
     \includegraphics[width=0.8\linewidth]{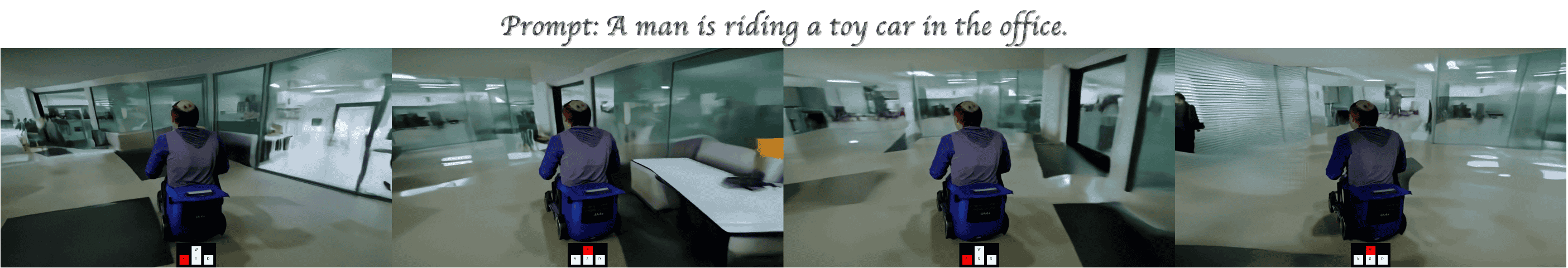}
     \\
     \includegraphics[width=0.2\linewidth]{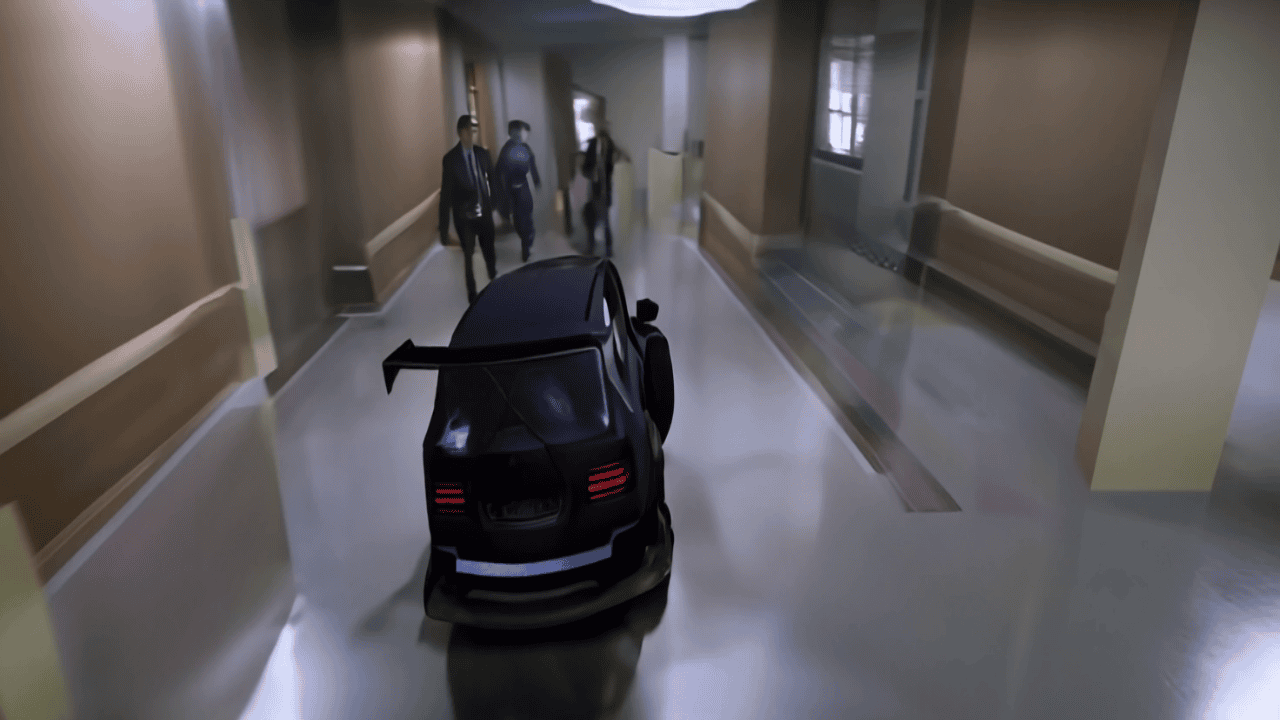} 
     \includegraphics[width=0.8\linewidth]{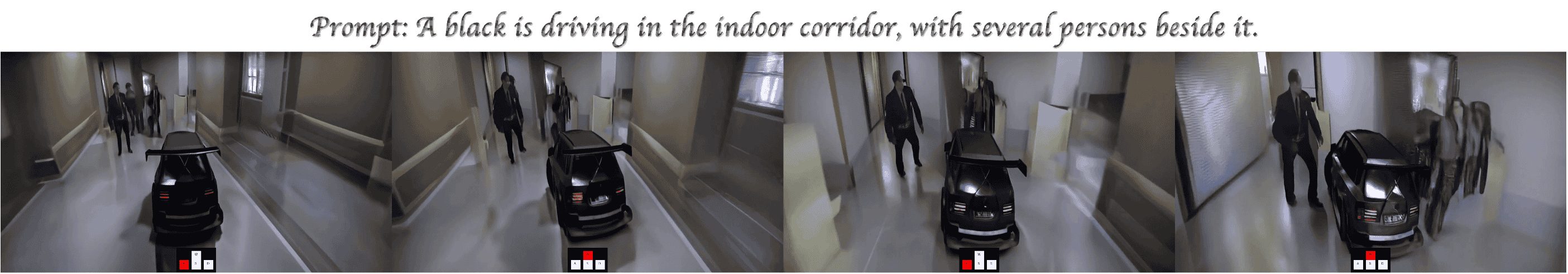}
     \\
     \includegraphics[width=0.2\linewidth]{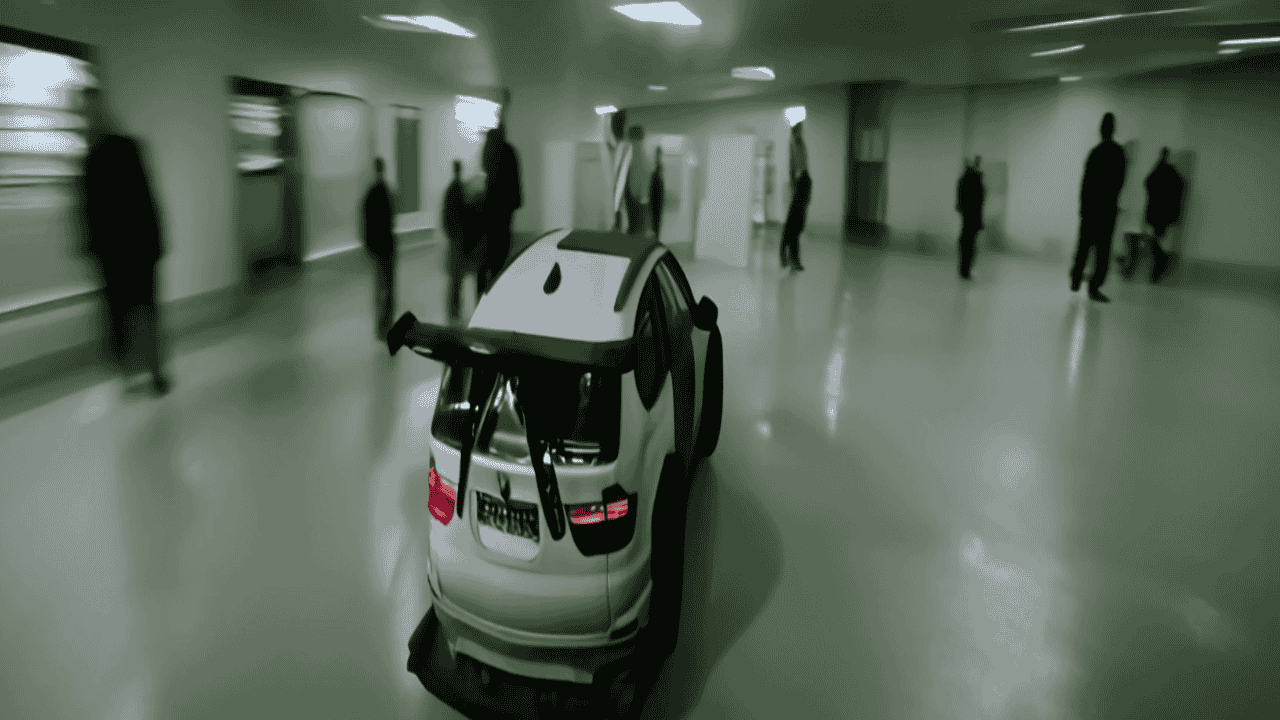} 
     \includegraphics[width=0.8\linewidth]{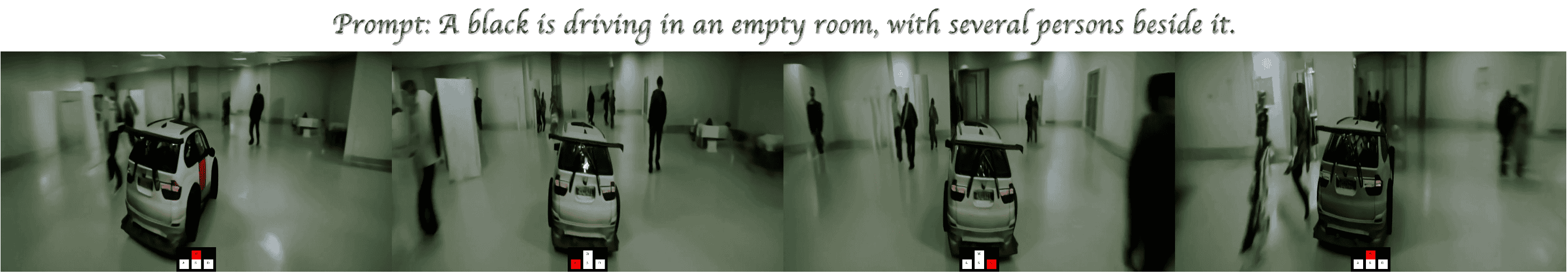}
     
    \end{tabular}

     \caption{More generalization results of \method on unseen scenes and objects.}
     \label{fig:sup_generalization}
   \end{figure*}

We provide more results on the generalization ability of \method in \cref{fig:sup_generalization}.

\subsection{Long Video Generation}
We provide several long video generation demos in the \emph{Supplementary Video files.} Please check them after \rm{Unzip}. \textbf{All videos are heavily compressed to satisfy the supplementary file size limit.}

\clearpage
{
    \small
    \bibliographystyle{ieeenat_fullname}
    \bibliography{main}
}


\end{document}